\documentclass[titlepage]{book}

\usepackage{graphicx, amssymb, amsmath, enumerate, mathrsfs, dsfont}
\usepackage{kotex}
\usepackage{epsfig, color, tcolorbox}
\usepackage[utf8]{inputenc}
\usepackage{caption}
\usepackage{subcaption}
\usepackage{makeidx, minitoc}
\usepackage{fancyhdr}
\usepackage[colorlinks=false]{hyperref}
\hypersetup{pdfborder=0 0 0}
\usepackage{cancel}
\usepackage{imakeidx}
\usepackage{tikz} 
\usetikzlibrary{bayesnet}

\usepackage{natbib}

\usepackage{paralist}

\graphicspath{ {./images/} }

\usepackage{tikz}
\usetikzlibrary{bayesnet}

\title{Machine Learning: a Lecture Note}

\author{
    Kyunghyun Cho \\
    New York University \& Genentech
}

\date{\today}

\begin{document}


\maketitle


\begin{center}
    \textbf{\Large{PREFACE}}
\end{center}

I prepared this lecture note in order to teach DS-GA 1003 ``Machine Learning'' at the Center for Data Science of New York University. This is the first course on machine learning for master's and PhD students in data science, and my goal was to provide them with a solid foundation on top of which they can continue on to learn more advanced and modern topics in machine learning, data science as well as more broadly artificial intelligence. Because of this goal, this lecture note has quite a bit of mathematical derivations of various concepts in machine learning. This should not deter students from reading through this lecture note, as I have interleaved these derivations with accessible explanations on the intuition and insights behind these derivations. Of course, as I was preparing this note, it only became clear how shallow my own foundation in machine learning was. But, I tried. 

In preparing this lecture note, I tried my best to constantly remind myself of ``Bitter Lesson'' by Richard Sutton~\citep{sutton2019bitter}. I forced myself to present various algorithms, models and theories in ways that support scalable implementations, both for compute and data. All machine learning algorithms in this lecture are thus presented to work with stochastic gradient descent and its variants. Of course, there are other aspects of scalability, such as distributed computing, but I expect and hope that other more advanced follow-up courses would teach students with these advanced topics based on the foundation this course has equipped those students with.

Despite my intention to cover as much foundational topics as possible in this course, it only became apparent that one course is not long enough to dig deeper into all of these topics. I had to make a difficult decision to omit some topics I find foundational, interesting and exciting, such as online learning, kernel methods and how to handle missing values. There were on the other hand some topics I intentionally omitted, although I believe them to be foundational as well, because they are covered extensively in various other courses, such as sequence modeling (or large-scale language modeling). I have furthermore refrained from discussing any particular application, hoping that there are other follow-up courses focused on individual application domains, such as computer vision, computational biology and natural language processing. 

There are a few more modern topics I hoped I could cover but could not due to time. To list a few of those, they include ordinary differential equation (ODE) based generative models and contrastive learning for both representation learning and metric learning. Perhaps in the future, I could create a two-course series in machine learning and add these extra materials. Until then, students will have to look for other materials to learn about these topics. 

This lecture note is not intended to be a reference book but was created to be a teaching material. This is my way of apologizing in advance that I have not been careful at all on extensively and exhaustively citing all relevant past literature. I will hopefully add citations more thoroughly the next time I teach this same course, although there is no immediate plan to do so anytime soon.






\tableofcontents
\clearpage


\pagenumbering{arabic}

\chapter{An Energy Function}

A usual way to teaching machine learning is to go through different problem setups. It often starts with binary classification, when perceptron, logistic regression and support vector machines are introduced, and continues with multi-class classification. At this point, it is usual to introduce regression as a continuous version of classification. Often, at this point, one would learn about kernel methods and neural networks, with focus on backpropagation (a more recent development in terms of teaching machine learning.) This is also at a point where one would take a detour by learning probabilistic machine learning, with the eventual goal of introducing a Bayesian approach to machine learning, i.e., marginalization over optimization. The latter half of the course would closely resemble the contents so far however in an unsupervised setting, where we learn that machine learning can be useful even when observations are not associated with outcomes (labels). One would learn about a variety of matrix factorization techniques, clustering as well as probabilistic generative modeling. If the lecturer were ambitious, they would sneak in one or two lectures on reinforcement learning at the very end. 

A main issue of teaching machine learning in such a conventional way is that it is extremely inconvenient for students to see a common foundation underlying all these different techniques and paradigms. It is often challenging for students to see how supervised and unsupervised learning connect with each other. It is even more challenging for students to figure out that classification and clustering are simply two sides of the same coin. In my opinion, it is simply impossible to make a majority of students see the unifying foundation behind all these different techniques and paradigms if we stick to enumerating all these paradigms and techniques. In this course, I thus try to take a new approach to teaching machine learning, largely based on and inspired by an earlier tutorial paper authored by Yann LeCun and his colleagues~\citep{lecun2006tutorial}. Other than this tutorial paper, this approach does not yet exist and will take a shape I continue to write this lecture note as the course continues. 

To begin on this journey, we start by defining an energy function, or a negative compatibility score. This energy function $e$ assigns a real value to a pair of an observed instance and a latent instance $(x,z)$ and is parametrized by a multi-dimensional vector $\theta$.
\begin{align}
    e: \mathcal{X} \times \mathcal{Z} \times \Theta \to \mathbb{R}.
\end{align}
$\mathcal{X}$ is a set of all possible observed instances, 
$\mathcal{Z}$ is a set of all possible latent instances,
and $\Theta$ is a set of all possible parameter configurations. 

When the energy function is low (that is, the compatibility is high,) we say that a given pair $(x,z)$ is highly preferred given $\theta$. When the energy function is high, unsurprisingly we say that the given pair is not as preferred. 

The latent observation $z$ is, as the name suggests, not observed directly. It nevertheless plays an important role in capturing uncertainty. When we only observe $x$, but not $z$, we cannot fully determine how preferable $x$ is. With a certain set of values of $z$, the energy may be low, while it may be high with other values of $z$. This gives us a sense of the uncertainty. For instance, we can compute both the mean and variance of the energy of an observed instance $x$ by
\begin{align}
    e_\mu(x, \theta) = \mathbb{E}[e(x,z, \theta)] = \sum_{z \in \mathcal{Z}} p(z) e(x,z,\theta), \\
    e_v(x, \theta) = \mathbb{E}[(e(x,z, \theta) - e_\mu(x, \theta))^2].
\end{align}

Given an energy function $e$ and the parameter $\theta$, we can derive a variety of paradigms in machine learning by minimizing the energy function with respect to different variables. For instance, let the observation be partitioned into two parts; input and output and assume that there is no latent variable, i.e., $e([x, y], \varnothing, \theta)$. Given a new input $x'$, we can solve the problem of {\it supervised learning} by
\begin{align}
    \hat{y} = \arg\min_{y \in \mathcal{Y}} e([x', y], \varnothing, \theta),
\end{align}
where $\mathcal{Y}$ is the set of all possible outcomes $y$. When $\mathcal{Y}$ consists of discrete items, we call it {\it classification}. If $y$ is a continuous variable, we call it {\it regression}.

When $\mathcal{Z}$ is a finite set of discrete items, a given energy function $e$ defines the cluster assignment of an observation $x$, resulting in {\bf clustering}:
\begin{align}
    \hat{z} = \arg\min_{z \in \mathcal{Z}} e(x, z, \theta).
\end{align}
If $z$ is a continuous variable, we would solve the same problem but call it {\it representation learning}. 

All these different paradigms effectively correspond to solving a minimization problem with respect to some subset of the inputs to the energy function $e$. In other words, given a partially-observed input, we infer the unobserved part that minimizes the energy function. This is often why people refer to using any machine learning model after training as {\bf inference}. 

It is not trivial to solve such a minimization problem. The level of difficulty depends on a variety of factors, including how the energy function is defined, the dimensionalities of the observed as well as latent variables as well as the parameters themselves. Throughout the course, we will consider different setups in which efficient and effective optimization algorithms are known and used for inference. 

As the name `machine learning' suggests, a bulk of machine learning is on estimating $\theta$. Based on what we have seen above, it may be tempting to think that {\bf learning} is nothing but
\begin{align}
    \min_{\theta \in \Theta} 
    \mathbb{E}_{x \sim p_{\mathrm{data}}}
    \left[
    e(x, \varnothing, \theta)
    \right],
\end{align}
when there is no latent variable. It turned out unfortunately that learning is not as easy, since we must ensure that the energy assigned to undesirable observation, i.e. $p_{\mathrm{data}}(x) \downarrow$, must be relatively high. In other words, we must introduce an extra term that regularizes learning:
\begin{align}
    \min_{\theta \in \Theta} 
    \mathbb{E}_{x \sim p_{\mathrm{data}}}
    \left[
    e(x, \varnothing, \theta)
    -
    R(\theta)
    \right].
\end{align}
The choice of $R$ must be made appropriately for each problem we solve, and throughout the course, we will learn how to design appropriate regularizers to ensure proper learning.

Of course it becomes even more involved when there are latent (unobserved) variables $z$, since it require us to solve the problem of inference simultaneously as well. This happens for problems such as clustering where the cluster assignment of each observation is unknown and factor analysis where latent factors are unknown in advance. We will learn how to interpret such latent variables and algorithms that allow us to estimate $\theta$ in the absence of latent variables.

In summary, there are three aspects to every machine learning problem; (1) defining an energy function $e$ (parametrization), (2) estimating the parameters $\theta$ from data (learning), and (3) inferring a missing part given an partial observation (inference). Across these three steps sits one energy function, and once we obtain an energy function $e$, we can easily mix and match these steps from different paradigms of machine learning.

\chapter{Basic Ideas in Machine Learning with Classification}

\section{Classification} 

In the problem of classification, an observation $x$ can be split into the input and output; $[x, y]$. The output $y$ takes one of the finite number of categories in $\mathcal{Y}$. For now, we assume that there is no latent variable, i.e., $\mathcal{Z}=\varnothing$. Inference is quite trivial in this case, since all we need to do is to pick the category that has the lowest energy, after computing the energy for all possible categories one at a time:
\begin{align}
    \hat{y}(x) = \arg\min_{y \in \mathcal{Y}} e([x, y], \varnothing, \theta).
\end{align}

Of course, this can be computationally costly if either $|\mathcal{Y}|$ is large or $x$ is high-dimensional. We can overcome this issue by cleverly parametrizing the energy function for instance as
\begin{align}
    e([x,y], \varnothing, \theta) 
    =
    \mathrm{1}(y)^\top f(x, \theta),
\end{align}
where $\mathrm{1}(y) = [0, \ldots, 0, 1, 0, \ldots, 0]$ is an one-hot vector.
This one-hot vector is all zeroes except for the $y$-th element which is set to $1$.

$f: \mathcal{X} \times \Theta \to \mathbb{R}^{|\mathcal{Y}|}$ is a feature extractor that returns as many real values as there are categories. 
With this parametrization, we can compute the energy values of all categories in parallel. A relatively simple example of $f$ is a linear function, defined as 
\begin{align}
    f(x, \theta) = W x + b,
\end{align}
where $\theta = (W, b)$ with $W \in \mathbb{R}^{|\mathcal{Y}| \times |x|}$ and $b \in \mathbb{R}^{|\mathcal{Y}|}$. When such a linear feature extractor is used, we call it a {\it linear classifier}. 

A natural next question is how we can learn the parameters $\theta$ (e.g. $W$ and $b$). We approach learning from the perspective of optimization. That is, we establish a loss function first and figure out how to minimize the loss function averaged over a training set $D$, where the training set $D$ is assumed to consist of $N$ independently sampled observations from the identical distribution (i.i.d.):
\begin{align}
    D = \left\{ [x^n, y^n] \right\}_{n=1}^N.
\end{align}

Perhaps the most obvious loss function we can imagine is a so-called zero-one ($0$-$1$) loss:
\begin{align}
    L_{0-1}([x; y], \theta) = \mathds{1}(y \neq \hat{y}(x)),
\end{align}
where
\begin{align}
    \hat{y}(x) = \arg\min_{y' \in \mathcal{Y}} e([x, y'], \varnothing, \theta),
\end{align}
as described earlier (reproduced here for emphasis.) 
$\mathds{1}(a)$ is an indicator function defined as
\begin{align}
    \mathds{1}(a) = 
    \begin{cases}
        1,&\text{if } a \text{ is true.} \\
        0,&\text{otherwise.}
    \end{cases}
\end{align}
With this zero-one loss function, the overall objective of learning is then
\begin{align}
    \min_{\theta} \frac{1}{N} \sum_{n=1}^N L_{0-1}([x^n, y^n], \theta).
\end{align}

This optimization problem is unfortunately very difficult,
because there is almost no signal on how we can incrementally change $\theta$ to
gradually decrease the loss function. The zero-one loss is a piece-wise constant function with respect to $\theta$.
It is either $0$ or $1$, and any infinitesimal change to $\theta$ is unlikely to change the loss value. 
In other words, the only way to tackle this problem is to sweep through many (if not all) possible values of $\theta$
and to identify the one that has the lowest overall loss. 
Such an approach is called {\it blackbox optimization}, and is known to be notoriously difficult.

\subsection{Perceptron and margin loss functions}

Instead, we can come up with a proxy to this zero-one loss function, that is easier to optimize. 
We do so by assuming that the energy function is differentiable with respect to $\theta$, that is,
$\nabla_{\theta} e$ exists and is easily computable.\footnote{
    We will shortly see why it is find to assume that it is easily computable. 
}
Then, we just need to ensure that the loss function is not piece-wise constant with respect to
the energy function itself. 

We start by noticing that the zero-one loss is minimized ($=0$) when $y'$ associated with the lowest energy ($=\hat{y}$) 
coincides with $y$ from the training data. 
In other words, the zero-one loss is minimized when the energy associated with the true outcome $y$, i.e.,
$e([x,y], \varnothing, \theta)$, is lower than the energy associated with any other $y' \neq y$. 
This goal can then be written down as satisfying the following inequality:
\begin{align}
    e([x, y], \varnothing, \theta) \leq e([x, \hat{y}'], \varnothing, \theta) - m,
\end{align}
where $m > 0$ and 
\begin{align}
    \hat{y}' = \arg\min_{y' \in \mathcal{Y} \backslash \{y \}} e([x, y'], \varnothing, \theta).
\end{align}

By rearranging terms in this inequality we get
\begin{align}
    m + e([x, y], \varnothing, \theta) - e([x, \hat{y}'], \varnothing, \theta) \leq 0.
\end{align}
In order to satisfy this inequality, we need to minimize the left hand side (l.h.s.) until it hits $0$.
We do not need to further minimize l.h.s. after hitting $0$, since the inequality is already satisfied. 
This translates to the following so-called {\it margin loss} (or a {\it hinge loss}):
\begin{align}
    L_{\mathrm{margin}}([x, y], \theta)
    =
    \max( 0, 
    m + e([x, y], \varnothing, \theta)
    -
    e([x, \hat{y}'], \varnothing, \theta)
    ).
\end{align}
This loss is called a margin loss, because it ensures that there exists at least the margin of $m$ between
the energy values of the correct outcome $y$ and the second best outcome $\hat{y}'$. 
The margin loss is at the heart of {\it support vector machines}~\citep{cortes1995support}. 

Consider the case where $m=0$:
\begin{align}
    L_{\mathrm{perceptron}}([x, y], \theta)
    =
    \max(0, 
    e([x, y], \varnothing, \theta)
    -
    e([x, \hat{y}'], \varnothing, \theta)
    ).
\end{align}
If $y=\hat{y}$ (not $\hat{y}'$), the loss is already minimized at $0$, since 
\begin{align}
    e([x, y], \varnothing, \theta) < e([x, \hat{y}'], \varnothing, \theta).
\end{align}
In other words, if a given example $[x,y]$ is already correctly solved, we do not need to change $\theta$ for this example. 
We only update $\theta$ when $y \neq \hat{y}$. This loss is called a {\it perceptron loss} and dates back to 1950's~\citep{rosenblatt1958perceptron}. 

\subsection{Softmax and cross entropy loss}
\label{sec:softmax}

It is often convenient to rely on the probabilistic framework, since it allows us to use a large set of tools developed for probabilistic inference and statistical techniques. As an example of doing so, we will now derive a probabilistic classifier from the energy function $e([x,y], \varnothing, \theta)$. The first step is to turn this energy function into a Categorical distribution over $\mathcal{Y}$ given the input $x$. 

Let $p_{\theta}(y|x)$ be the Categorical probability of $y$ given $x$. There are two major constraints that must be satisfied:
\begin{enumerate}
    \item Non-negativity: $p_{\theta}(y|x) \geq 0$ for all $y \in \mathcal{Y}$.
    \item Normalization: $\sum_{y' \in \mathcal{Y}} p_{\theta}(y'|x) = 1$.
\end{enumerate}

Of course, there can be many (if not infinitely many) different ways to map $e([x,y], \varnothing, \theta)$ to $p_{\theta}(y|x)$, while satisfying these two conditions~\citep{peters2019sparse}. We thus need to impose a further constraint to narrow down on one particular mapping from the energy function to the Categorical probability. A natural such constraint is the maximum entropy criterion. 

The (Shannon) entropy is defined as 
\begin{align}
    H(y|x; \theta) = -\sum_{y \in \mathcal{Y}} p_{\theta}(y|x) \log p_{\theta}(y|x).
\end{align}
The entropy is large if there is a large degree of uncertainty. In order to cope with the issue of $\log 0$, we assume that
\begin{align}
    H(y|x; \theta) = 0,\text{ if } p_{\theta}(y|x) = \begin{cases} 0\\1 \end{cases}.
\end{align}

Why is this natural? Because, it is our way to explicitly concede that we are not fully aware of the world and that there may be somethings that are not known, resulting in some uncertainty about our potential choice. This is often referred to as the {\it principle of maximum entropy}~\citep{jaynes1957information}.

Then, we can convert the energy values $\left\{ a_1=e([x,y=1], \varnothing, \theta), \ldots, a_d=e([x, y=d], \varnothing, \theta) \right\}$ assigned to different outcome classes $\mathcal{Y}=\left\{ 1, 2, \ldots, d \right\}$ into the Categorical probabilities $\left\{ p_1, \ldots, p_d \right\}$ by solving the following constrained optimization problem:
\begin{align}
    \max_{p_1, \ldots, p_d} -\sum_{i=1}^d a_i p_i - \sum_{i=1}^d p_i \log p_i
\end{align}
subject to
\begin{align}
    &p_i \geq 0, \text{ for all } i=1,\ldots, d \\
    &\sum_{i=1}^d p_i = 1.
\end{align}

We can solve this optimization problem with the method of Lagrangian multipliers. First, we write the unconstrained objective function:
\begin{align}
    J(p_1, \ldots, p_d, \lambda_1, \ldots, \lambda_d, \gamma) 
    =
    -\sum_{i=1}^d a_i p_i - \sum_{i=1}^d p_i \log p_i
    + \sum_{i=1}^d \lambda_i (p_i - s_i^2) 
    + \gamma(\sum_{i=1}^d p_i - 1),
\end{align}
where $\lambda_1, \ldots, \lambda_d$ and $\gamma$ are Lagragian multipliers, and $s_1, \ldots, s_d$ are slack variables.

Let us first compute the partial derivative of $J$ with respect to $p_i$ and set it to $0$:
\begin{align}
\lefteqn{
    \frac{\partial J}{\partial p_i}
    =
    -a_i - \log p_i - 1 + \lambda_i + \gamma = 0} \\
    &\iff
    \log p_i = -a_i + \lambda_i -1 + \gamma \\
    &\iff
    p_i = \exp(-a_i + \lambda_i - 1 + \gamma) > 0.
\end{align}

We notice that $p_i$ is already greater than $0$ at this extreme point, meaning that the first constraint $p_i \geq 0$ is already satisfied. We can just set $\lambda_i$ to any arbitrary value, and we will pick $0$, i.e., $\lambda_i = 0$ for all $i=1, \ldots, d$. This results in
\begin{align}
    p_i = \exp(-a_i) \exp(-1 + \gamma).
\end{align}

Let us now plug it into the second constraint and solve for $\gamma$:
\begin{align}
    &\exp(-1 + \gamma) \sum_{i=1}^d \exp(-a_i) = 1 \\
    \iff&
    -1 + \gamma + \log \sum_{i=1}^d \exp(-a_i) = 0 \\
    \iff&
    \gamma = 1 - \log \sum_{i=1}^d \exp(-a_i).
\end{align}

By plugging it into $p_i$ above, we get
\begin{align}
    p_i =& \exp(-a_i) \exp(-1 + 1 - \log \sum_{j=1}^d \exp(-a_j)) \\
    =& \frac{\exp(-a_i)}{\sum_{j=1}^d \exp(-a_j)}.
\end{align}
This formulation is often referred to as {\it softmax}~\citep{bridle1990probabilistic}. 

Now, we have the Categorical probability $p_i = p_{\theta}(y=i|x)$. 
We can then define an objective function under the probabilistic framework, as
\begin{align}
\label{eq:cross-entropy}
    L_{\mathrm{ce}}([x, y]; \theta)
    =
    -\log p_{\theta}(y | x)
    =
    e([x, y], \varnothing, \theta) 
    +
    \log \sum_{y' \in \mathcal{Y}}
    \exp(-e([x, y'], \varnothing, \theta)).
\end{align}
We often call this a {\it cross-entropy loss}, or equivalently negative log-likelihood. 

Unlike the margin and perceptron losses from above, it is more informative to consider the gradient of the cross-entropy loss:
\begin{align}
    \nabla_\theta L_{\mathrm{ce}} ([x,y], \varnothing, \theta)
    &=
    \nabla_\theta e([x,y], \varnothing, \theta)
    -
    \sum_{y' \in \mathcal{Y}}
    \underbrace{
    \frac{\exp(-e([x, y'], \varnothing, \theta))}
    {\sum_{y'' \in \mathcal{Y}} \exp(-e([x,y''], \varnothing, \theta))}
    }_{
    =p_{\theta}(y' | x)
    }
    \nabla_\theta e([x, y'], \varnothing, \theta)) \\
    &=
    \underbrace{\nabla_\theta e([x,y], \varnothing, \theta)}_{\mathrm{(a)}}
    -
    \underbrace{
    \mathbb{E}_{y |x; \theta}
    \left[ 
    \nabla_\theta e([x, y'], \varnothing, \theta))
    \right]}_{
    \mathrm{(b)}
    }.
\end{align}
This gradient, or an update rule since we update $\theta$ following this direction, is called a Boltzmann machine learning~\citep{ackley1985learning}. 

There are two terms in this update rule; (a) positive and (b) negative terms.
The positive term corresponds to increasing the energy value associated with the true outcome $y$.\footnote{
    Recall that this is a loss which is minimized. 
}
The negative term corresponds to decreasing the energy values associated with all possible outcomes, but they are 
weighted according to how likely they are under the current parameters. 

Let us consider the negative term a bit more carefully:
\begin{align}
    -\sum_{y' \in \mathcal{Y}}
    \frac{\exp(-\beta e([x, y'], \varnothing, \theta))}
    {\sum_{y'' \in \mathcal{Y}} \exp(-\beta e([x,y''], \varnothing, \theta))}
    \nabla_\theta e([x, y'], \varnothing, \theta)).
\end{align}
$\beta$ was added to make our analysis easier. We often call $\beta$ an {\it inverse temperature}. $\beta$ is by default $1$, but by varying $\beta$, we can gain more insights into the negative term.

Consider the case where $\beta = 0$, the negative term reduces to
\begin{align}
    -\frac{1}{|\mathcal{Y}|} \sum_{y' \in \mathcal{Y}}
    \nabla_\theta e([x, y'], \varnothing, \theta)).
\end{align}
This would correspond to increasing the energy associated with each outcome equally. 

How about when $\beta \to \infty$? In that case, the negative term reduces to
\begin{align}
    -\nabla_\theta e([x, \hat{y}], \varnothing, \theta),
\end{align}
where 
\begin{align}
    \hat{y} = \arg\min_{y \in \mathcal{Y}} e([x, y], \varnothing, \theta).
\end{align}

When $\beta\to\infty$, we end up with two cases. First, the classifier makes the correct prediction; $\hat{y}=y$. 
In this case, the positive and negative terms cancel each other, and there is no gradient. Hence, there is no update to the parameters. This reminds us of the perceptron loss from the earlier section. On the other hand, if $\hat{y} \neq y$, it will try to lower the energy value associated with the correct outcome $y$ while increasing the energy value associated with the current prediction $\hat{y}$. This continues until the prediction matches the correct outcome. 

These two extreme cases tell us what happens with the cross entropy loss. It softly adjust the energy values associated with all possible outcomes however based on how likely they are to be the prediction. The cross entropy loss has become more or less {\it de facto} standard when it comes to training a neural network in recent years.

\section{Backpropagation}

Once you decide the loss function, it is time for us to {\it train} a classifier to minimize the average loss. In doing so, one of the most effective approaches has been stochastic gradient descent, or its variant. Stochastic gradient descent, which we will discuss more in-depth later, takes a subset of training instances from $D$, computes and averages the gradients of the loss of each instance in this subset and updates the parameters in the negative direction of this {\it stochastic gradient}. This makes it both interesting and important for us to think of how to compute the gradient of a loss function. 

Let us consider both the margin loss and cross entropy loss, since there is no meaningful gradient of the zero-one loss function and the perceptron loss is a special case of the margin loss:
\begin{align}
    &\nabla_{\theta} L_{\mathrm{margin}}([x, y], \theta) 
    =
    \begin{cases}
        \nabla_{\theta} e([x,y], \varnothing, \theta) 
    - \nabla_{\theta} e([x,\hat{y}'], \varnothing, \theta), &\text{ if } L_{\mathrm{margin}}([x, y], \theta) > 0. \\
    0, &\text{ otherwise.}
    \end{cases}
    \\
    &\nabla_{\theta} L_{\mathrm{ce}}([x, y], \theta)
    =
    \nabla_\theta e([x,y], \varnothing, \theta)
    -
    \mathbb{E}_{y |x; \theta}
    \left[ 
    \nabla_\theta e([x, y'], \varnothing, \theta))
    \right].
\end{align}
In both cases, the gradient of the energy function shows up: $\nabla_\theta e([x, y], \varnothing, \theta))$. 
We thus focus on the gradient of the energy function in this case.

\subsection{A Linear Energy Function}

Let us start with a very simple case we considered earlier. We assume that $x$ is a real-valued vector of $d$ dimensions, i.e., $x \in \mathbb{R}^d$. We will further assume that $y$ takes one of $K$ potential values, i.e., $y \in \left\{ 1, 2, \ldots, K \right\}$. The parameters $\theta$ consist of 
\begin{enumerate}
    \item The weight matrix $W = \left[ 
    \begin{array}{c}
    w_1 \\
    w_2 \\
    \vdots \\
    w_K
    \end{array}
    \right] \in \mathbb{R}^{K \times d}$

    \item The bias vector $b = \left[ \begin{array}{c} 
    b_1 \\ 
    b_2 \\
    \vdots \\
    b_K
    \end{array} \right] \in \mathbb{R}^{K}$
\end{enumerate}

We can now define the energy function as 
\begin{align}
\label{eq:linear-energy}
    e([x, y], \varnothing, \theta) = -w_y^\top x - b_y.
\end{align}

The gradient of the energy function with respect to the associated weight vector $w_y$ is then
\begin{align}
    \nabla_{w_y} e = -x.
\end{align}
Similarly, for the bias:
\begin{align}
    \frac{\partial e}{\partial b_y} = -1.
\end{align}

The first one (the gradient w.r.t. $w_y$) states that for the energy to be lowered for this particular combination $(x,y)$, we should add the input $x$ to the weight vector $w_y$. The second one (the gradient w.r.t. $b_y$) lowers the energy for the outcome $y$ regardless of the input. 

Let us consider the perceptron loss, or the margin loss with zero margin. The first-term gradient, $\nabla_{\theta} e([x,y], \varnothing, \theta)$, updates the weight vector and the bias value associated with the {\it correct} outcome. With a learning rate $\eta > 0$, the updated energy associated with the correct outcome, where we follow the negative gradient,\footnote{
    We will shortly discuss why we do so later in this chapter.
} 
is then smaller than the original energy function:
\begin{align}
    -(w_y + \eta x)^\top x - (b_y + \eta) 
    =&
    -w_y^\top x - b_y - \eta (\| x \|^2 +1)
    \\
    =&
    e([x, y], \varnothing, \theta) - \eta (\| x \|^2 + 1)
    \\
    <&~
    e([x, y], \varnothing, \theta).
\end{align}
This is precisely what we intended, since we want the energy value to be lower with a good combination of the input and outcome. 

This alone is however not enough as a full learning rule. Even if the energy value associated with the right combination is lowered, it may not be lowered enough, so that the correct outcome is selected when the input is presented again. The second-term gradient compliments this by having the opposite sign in front of it. By following the negative gradient of the negative energy associated with the input and the predicted outcome $\hat{y}$, we ensure that this particular energy value is increased:
\begin{align}
    -(w_{\hat{y}} - \eta x)^\top x - (b_{\hat{y}} - \eta) 
    &=
    e([x, \hat{y}], \varnothing, \theta) + \eta (\| x \|^2 + 1)
    \\
    &>
    e([x, \hat{y}], \varnothing, \theta).  
\end{align}

So, this learning rule would lower the energy value associated with the correct outcome and increase that associated with the incorrectly-predicted outcome, until the outcome with the lowest energy coincides with the correct outcome. When that happens, the loss is constant, and no learning happens, because $y = \hat{y}$. 

At this point, we start to see that the derivation and argument above equally apply to $x$, the input. Instead of the gradient of the energy w.r.t. the weight vector $w_y$, but we can compute that w.r.t. the input $x$ as well:
\begin{align*}
    \nabla_{x} e = -w_y,
\end{align*}
assuming that $x$ is continuous and the energy function is differentiable w.r.t. $x$. By following the (opposite of the) gradient in the input space, we can alter the loss function, instead of modifying the weight vectors and biases. 

Of course this is absolutely the opposite of what we are trying to do here, since the main goal is to find a classifier that classifies a {\it given} input $x$ into the correct category $y$. This perspective however leads us naturally to the idea of backpropagation~\citep{rumelhart1986learning}.

\subsection{A Nonlinear Energy Function}
\label{sec:backprop}

Instead of adjusting the weight vector $W$ and the bias vector $b$, we can adjust the input $x$ directly in order to modify the associated energy value. More specifically, with the perceptron loss, that is the margin loss with zero margin, when the prediction is incorrect, i.e. $y \neq \hat{y}$, the gradient of the perceptron loss with respect to the input $x$ is\footnote{
    When it is clear that there is no latent (unobserved) variable $z$, I will skip $\varphi$ for brevity.
}
\begin{align}
\label{eq:obs_grad}
\nabla_x L_{\mathrm{perceptron}}([x, y], \theta) =
    \nabla_x e([x, y], \theta) - \nabla_x e([x, \hat{y}], \theta) 
    =
    -w_y + w_{\hat{y}}.
\end{align}

Similarly to the weight matrix and bias vector above, if we update the input $x$ following the opposite of this direction, we can increase the energy value associated with the correct outcome $y$ while lowering that with the incorrectly-predicted outcome $\hat{y}$. Although this is generally useless with a linear energy function, as we discussed just now, this is an interesting thought experiment, as this tells us that we can solve the problem either by adapting the parameters, i.e. the weight matrix and bias vector, or by adapting the input data points themselves. The latter sounds like an attractive alternative, because it would break us free from being constrained by the linearity of the energy function. 

There is however a major issue with the latter alternative. That is, we do not know how to change the new input in the future (not included in the training set), since such a new input may not come together with the associated correct outcome. We thus need to build a system that predicts what the altered input would be given a new input in the future. 

To overcome this issue, we start by using some transformation $h$ of the input $x$, with its own parameters $\theta'$, instead of the original input $x$. That is, $h=F(x, \theta')$. Analogously, we refer to the newly updated input by $\hat{h}$. We obtain $\hat{h}$ by following the gradient direction from Eq.~\eqref{eq:obs_grad}.  We now define a new energy function $e'$ such that the combination $(h, \hat{h})$ is assigned a lower energy than the other combinations if $h$ and $\hat{h}$ are close to each other.  Under this energy function, the energy is low if this transformation of the input $h=F(x, \theta')$ is similar to the updated input $\hat{h}$. This intuitively makes sense, since $\hat{h}$ is the desirable transformation of the input $x$, as it lowers the overall loss function above. 

A typical example of such an energy function would be
\begin{align}
\label{eq:l2-energy}
    e'([h, \hat{h}], \theta')
    =
    \frac{1}{2} \| \underbrace{\sigma(U^\top x + c)}_{=h} - \hat{h} \|^2,
\end{align}
where $U$ and $c$ are the weight matrix and bias vector, respectively, and $\sigma$ is an arbitrary nonlinear function. $h=\sigma(U^\top x + c)$ would be {\it some transformation} of the input $x$, as described above. 

The loss function in this case can be simply the energy function itself:
\begin{align}
    L_{\ell_2}([h, \hat{h}], \theta') = e'([h, \hat{h}], \theta').
\end{align}
The gradient of the loss function w.r.t. the transformation matrix $U$ is then:
\begin{align}
    \nabla_U = x \left( (h - \hat{h}) \odot h' \right)^\top  
\end{align}
where 
\begin{align}
    h' = \sigma'(U^\top x + c)
\end{align}
with $\sigma'(a) = \frac{\partial \sigma}{\partial a} (a)$, according to the chain rule of derivatives. $\odot$ denotes element-wise multiplication. Similarly, the gradient w.r.t. the bias vector $c$ is
\begin{align}
    \nabla_c = (h - \hat{h}) \odot h'.
\end{align}
Before continuing further, let us examine these gradients. If we look at $\nabla_c$, the first term, or its negation, since we want to minimize the energy, states that we should change $c$ toward $\hat{h}$ away from $h$. If $h$ is further away from $\hat{h}$, we need to change $c$ more. The second term $h'$ is multiplied to $(h-\hat{h})$. This term $h'$ is the slope of the nonlinear activation function $\sigma$ at the current input $U^\top x + c$. If the slope is positive, we should update $c$ following the sign of $\hat{h}-h$ , as usual. But, if the slope is negative, we should flip the direction of $c$'s update, since increasing $c$ would result in decreasing $\hat{h}-h$. 

In order to analyze the gradient w.r.t. $U$, let us consider the gradient w.r.t. one particular element of $U$, i.e., $u_{ij}$. $u_{ij}$ can be thought of as the weight between the $i$-th dimention of the input, $x_i$, and the $j$-th dimension of the transformation, $h_j$. This gradient is written down as
\begin{align}
\label{eq:hebbian-rule}
    \frac{\partial}{\partial u_{ij}} = x_i (h_j - \hat{h}_j) h'_j = (x_i h_j - x_i \hat{h}_j) h'_j.
\end{align}
We already know what $h'_j$ does: it decides whether the slope was positive or negative, and thereby whether the update direction should flip. Because we follow the opposite direction (since we want to lower the energy), the first term $x_i h_j$ is subtracted from $u_{ij}$. This term tells us how strongly the value of $x_i$ is reflected on the value of $h_j$. Since $h_j$ is now being updated away, the effect of $x_i$ on the $j$-th dimension of the transformation via $u_{ij}$ must be reduced. On the other hand, the second term $x_i \hat{h}_j$ does the opposite. It states that the effect of $x_i$ on the $j$-th dimension of the transformation, according to the newly updated value $\hat{h}_j$, must be reflected more on $u_{ij}$. If the new value of the $j$-th dimension has the same sign as $x_i$, $u_{ij}$ should tend toward the positive value. Otherwise, it should tend toward the negative value. 

We can now imagine a procedure where we alternate between computing $\hat{h}$  and updating $U$ and $c$ to match $\hat{h}$. Of course, this procedure may not be optimal, since there is no guarantee (or it is difficult to obtain any guarantee) that repeatedly updating $U$ and $c$ following the gradient of the second energy function leads to improvement in the overall loss when $h=\sigma(U^\top x + c)$ is used in place of the target $\hat{h}$. When the second energy function is truly minimized so that $\sigma(U'^\top x + c')$ coincides with $\hat{h}$, the loss will be smaller than the original $h=\sigma(U^\top x + c)$. It is however unclear whether the loss will be smaller until this minimum is achieved.

Instead, we can think of a procedure in which we update $U$ and $c$ directly without producing $\hat{h}$ as an intermediate quantity. Assume we take just a unit step to update $\hat{h}$:
\begin{align}
    &\hat{h} = h + (w_y - w_{\hat{y}}) \\
    \iff& \hat{h} - h  = - \nabla_h L(h).
\end{align}
That is, we use the learning rate (or step size) of $1$.

Then, 
\begin{align}
    &\nabla_U = x \left( \nabla_h L(h) \odot h' \right)^\top   \\
    &\nabla_c = \nabla_h L(h) \odot h'.
\end{align}
In other words, we can skip computing $\hat{h}$ and directly compute the gradients of the loss w.r.t. $U$ and $c$ using the gradient w.r.t. $h$. 

Just like what we did with $h$ (or originally $x$), we can check how we would change this new $x$ to minimize the second energy function $e'$. This is done by computing the gradient of $e'$ w.r.t. $x$:
\begin{align}
    \nabla_x = U \left( (h-\hat{h}) \odot h' \right),
\end{align}
which is similar to the gradient w.r.t. $U$. If we replace $(h-\hat{h})$ with $\nabla_h L(h)$, we get 
\begin{align}
    \nabla_x = U \left( \nabla_h L(h) \odot h' \right).
\end{align}
It is the third time we are discussing it, but yes, we know what $h'$ does here: it decides the sign of the update. If we ignore $h'$ by simply assuming that $\sigma$ was e.g. an identity map (which would mean that $h'=1$), we realize that $\nabla_x$ is linear transformation of $\nabla_h$ \footnote{
    Whenever it is clear, we will drop some terms for both brevity and clarify. In this case, $\nabla_h$ is $\nabla_h L(h)$.
} , as
\begin{align}
    \nabla_x = U \nabla_h.
\end{align}
Contrast it against the red-coloured term below:
\begin{align}
    h = \sigma({\color{red} U^\top x} + c)
\end{align}
The red-colour term above can be thought of as propagating the input signal $x$ via $U^\top$ to $h$. In contrast $U \nabla_h$ can be thought of as {\it back}-propagating the error signal $\nabla_h$ via $U$ to the input $x$. 

You must see where we are heading toward now. Let us replace $x$ once more, this time, with $z$. In other words, 
\begin{align*}
    h = \sigma(U^\top z + c)
\end{align*}
and
\begin{align*}
    z = \sigma(V^\top x + s).
\end{align*}
We can analogously introduce yet another energy function $e''$ defined as 
\begin{align}
    e''([z, \hat{z}], \theta'') = \frac{1}{2} \| z - \hat{z} \|^2,
\end{align}
where 
\begin{align}
    \hat{z} &= z - \nabla_z \\
    &= z - U \nabla_h.
\end{align}
Following the exactly same steps of derivation from above, we end up with 
\begin{align}
    &\nabla_V = x \left( \nabla_z \odot z' \right)^\top   \\
    &\nabla_s = \nabla_z \odot z',
\end{align}
where
\begin{align}
    \nabla_z = U \nabla_h.
\end{align}
In one single sweep, we could backpropagate the error signal from the loss function all the way back to $x$ and compute the gradient of the loss function w.r.t. all the parameters, $W, b, U, c, V$ and $s$. Of course, in doing so, we had to store the so-called forward activation vectors, $x, z$ and $h$, which is often referred to book-keeping.

This process of computing the gradient of the loss fucntion w.r.t. all the parameters from multiple stages of nonlinear transformation of the input is called backpropgation. This can be generalized to any computation graph without any loops (though, loops can be unrolled for a finite number of cycles in practice) and is a special case of automatic differentiation~\citep{baydin2018}, called reverse-mode automatic differentiation. 

Because reverse-mode automatic differentiation is efficient both in terms of computation and memory (both linear), it is universally used for computing the gradient and is well-implemented in many widely used deep learning tools, such as PyTorch and Jax. This universality implies that once we decide on a loss function and an energy function such that the loss function is differentiable w.r.t. the parameters of the energy function, we can simply assume the gradient would be readily available. 

\section{Stochastic Gradient Descent}

Once we have defined an energy function and an associated loss function, we can compute the gradient of this loss function w.r.t. the parameters. With the gradient, we can update the parameters repeatedly so that we can minimize the loss function. It is important to observe that we have defined the loss function for each individual training example, and eventually our goal becomes minimizing the average of the loss of all training examples. For a random reason, we will use $f_i(\theta)$ to denote the loss function of the $i$-th example at $\theta$, and thereby the overall loss is
\begin{align}
    f(\theta) = \frac{1}{N} \sum_{i=1}^N f_i(\theta).
\end{align}
When the overall loss is the average (or sum) of the individual loss functions, we say that the loss is {\it decomposable}. 

We can view such an overall loss function as computing the expected individual loss function:
\begin{align}
\label{eq:expected_loss}
    f(\theta) = \mathbb{E}_i \left[ f_i (\theta) \right],
\end{align}
where $i \sim \mathcal{U}(1, \ldots, N)$. Of course, we can replace this uniform distribution with an arbitrary data distribution and write this as
\begin{align}
    f(\theta) = \mathbb{E}_{x \sim p_\mathrm{data}}\left[ f(x; \theta) \right],
\end{align}
although we will for now stick to the uniform indexing over the training set.

With Eq.~\eqref{eq:expected_loss}, we also get
\begin{align}
\label{eq:expected_grad}
\nabla f = \mathbb{E}_i \left[ \nabla f_i \right],
\end{align}
because the expectation over a finite, discrete random variable can be written down using a finite sum.

There are two constants we should consider when deciding how we are going to minimize $f$ w.r.t. $\theta$. They are the number of training examples $N$ and the number of parameters $\mathrm{dim}(\theta)$ (if not confusing, we would use $\mathrm{dim}(\theta)$ and $|\theta|$ interchangeably.)  Let us start with the latter $|\theta|$. If the number of parameters is large, we cannot expect to compute any high-order derivative information of the function $f$ beyond the first-order derivative, that is its gradient. Without access to higher-order derivative, we cannot benefit from advanced optimization algorithms, such as Newton's algorithm. Unfortunately, in modern machine learning, $|\theta|$ can be as larger as tens of billions, and we are often stuck with first-order optimization algorithms. 

If $N$ is large, it becomes increasingly burdensome to compute $f$ not to mention $\nabla f$ directly each update. In other words, we can only expect to use the true gradient of $f$ only when there are few training examples only, i.e., small $N$. In modern machine learning, we are often faced with hundreds of thousands, if not millions or billions, of training examples, and it is often impossible for us to exactly compute the overall loss. In short, we are in a situation where we cannot even use the full, true gradient information to update the parameters.

In order to cope with large $N$ and large $|\theta|$, we often resort to a stochastic gradient estimate rather than the full gradient, where the stochastic gradient is defined as
\begin{align}
    g_{i_t} = \nabla f_{i_t}(\theta_t), 
\end{align}
where $i_t$ was drawn from the uniform distribution over $\left\{1, \ldots, N \right\}$. We then update the parameters using this stochastic gradient estimate by
\begin{align}
\label{eq:sgd}
    \theta_{t+1} = \theta_t - \alpha_t g_{i_t}.
\end{align}
In doing so, it is a usual practice to maintain a set of so-called checkpoints and pick the best one within this checkpoint set. We will discuss how we pick the best checkpoint according to which criteria in the next section in more detail, as this is where optimization and learning deviate from each other. 

For now, let us stick to optimization and in particular iterative optimization. When thinking about optimization, there are two distinct concepts that are equally important. The first one is convergence. With convergence we mean whether iterative optimization gradually moves the iterate $\theta_t$ toward a desirable solution. A desirable solution could the global minimum (if it exists), any local minimum or any extremum (where the gradient is zero.) It is important to know whether the iterate converges to such a desirable solution and if so, at which rate. The second important concept is the descent property. An iterative optimization algorithm is descent if it always makes progress, that is, $f(\theta_{t+1}) \leq f(\theta_t)$ for all $t$. 

As we will learn about it shortly in the next section, the desirable solution is not defined with the overall loss function $f$. Rather, the desirable solution for us is defined using another function $f^*$. This function $f^*$ is similar to $f$ almost everywhere over $\theta$ but these two functions differ. It is thus more desirable for us to enumerate a series of $\theta_t$'s with small $f(\theta_t)$ values and eventually pick one using $f^*(\theta_t)$. In other words, it is not convergence but the descent property. 

\subsection{Descent Lemma}

We start by stating and proving one of the most fundamental results in optimization, called the {\it descent lemma}. According to the descent lemma, the following inequality holds when $\nabla f$ is an $L$-Lipschitz continuous function, i.e., $\| \nabla f(x) - \nabla f(y) \| \leq L \| x - y \|$:
\begin{align}
    \label{eq:descent-lemma}
    f(y) \leq f(x) + \nabla f(x)^\top (y - x) + \frac{L}{2} \| y - x \|^2.
\end{align}
This inequality allow us to upper-bound the value of a function at a point $y$ given the value as well as the gradient at another point $x$. 

Let $g(t) = f(x + t(y - x))$ so that $g(0) = f(x)$ and $g(1) = f(y)$. Then, 
\begin{align}
    f(y) - f(x) = g(1) - g(0) = \int_{0}^1 g'(u) \mathrm{d} u = \int_0^1 \nabla f(x+t(y-x))^\top (y-x) \mathrm{d}t.
\end{align}
By subtracting $\nabla f(x)^\top (y-x)$ from both sides, we get
\begin{align}
    f(y) - f(x) - \nabla f(x)^\top (y-x) =
    \int_0^1 (\nabla f(x + t(y-x)) - \nabla f(x))^\top (y-x) \mathrm{d} t.
\end{align}
We can upperbound it using the Cauchy-Schwarz inequality, i.e. $a^\top b \leq \|a\| \|b \|$:
\begin{align}
    f(y)-f(x)-\nabla f(x)^\top (y-x) \leq \int_0^1 \| \nabla f(x+t(y-x)) - \nabla f(x) \| \| y - x \| \mathrm{d}t.
\end{align}
We can use the assumption above that $\nabla f$ is an $L$-Lipschitz function to simplify it into
\begin{align}
    f(y) - f(x) - \nabla f(x)^\top (y-x) \leq \int_0^1 Lt \| y-x \|^2 \mathrm{d}t = \frac{L}{2} \| y-x \|^2,
\end{align}
which is in turn
\begin{align}
    f(y) \leq f(x) + \nabla f(x)^\top (y-x) + \frac{L}{2} \| y-x \|^2.
\end{align}
If we assume that $N$ is not too large, we can compute the gradient exactly and update the parameters following the negative gradient direction:
\begin{align}
\label{eq:gd}
&\theta_{t+1} = \theta_t - \alpha_t \nabla f(\theta_t) \\
\iff& \theta_{t+1} - \theta_t = -\alpha_t \nabla f(\theta_t)
\end{align}
Let us plug $(\theta_t, \theta_{t+1})$ into $(x,y)$ in the descent lemma:
\begin{align}
f(\theta_{t+1}) &\leq f(\theta_t) - \alpha_t \| \nabla f(\theta_t)\|^2 
+ \alpha_t^2 \frac{L}{2} \| \nabla f(\theta_t) \|^2 \\
&=f(\theta_t) -(\alpha_t - \frac{L}{2} \alpha_t^2) \| \nabla f(\theta_t) \|^2.
\end{align}
Since $\| \nabla f(\theta_t) \|^2 \geq 0$, we want to find $\alpha_t$ that maximizes $-\frac{L}{2} \alpha_t^2 + \alpha_t$. We simply compute the derivative of this expression w.r.t. $\alpha_t$ and set it to zero:
\begin{align}
    -L \alpha_t + 1 = 0 \iff \alpha_t = \frac{1}{L}.
\end{align}
In other words, if we set the learning rate to $1/L$ (that is, inverse proportionally to how rapidly the function changes), we can make the most progress each time. Of course, this does not directly apply to the stochastic case, since the descent lemma does not apply to the stochastic gradient estimate as it is. 

\subsection{Stochastic Gradient Descent}
\label{sec:sgd}

Resuming from the descent lemma above, we will use the stochastic gradient update rule from Eq.~\eqref{eq:sgd}. Let's restate the stochastic gradient rule:
\begin{align}
    \theta_{t+1} = \theta_t - \alpha_t g_{i_t} \iff \theta_{t+1} - \theta_t = - \alpha_t g_{i_t}.
\end{align}
Plugging in $(\theta_t, \theta_{t+1})$ into the descent lemma, we get
\begin{align}
    f(\theta_{t+1}) \leq f(\theta_t) - \alpha_t \nabla f(\theta_t)^\top g_{i_t} + \alpha_t^2 \frac{L}{2} \| g_{i_t} \|^2.
\end{align}
We are interested in the expected progress here over $i_t \sim \mathcal{U}(1, \ldots, N)$:
\begin{align}
\label{eq:decent-lamma}
    \mathbb{E}\left[f(\theta_{t+1})\right] 
    &\leq f(\theta_t) - \alpha_t \nabla f(\theta_t)^\top \mathbb{E}\left[g_{i_t}\right] + \alpha_t^2 \frac{L}{2} \mathbb{E} \| g_{i_t} \|^2 \\
    &=f(\theta_t) 
    - \alpha_t {\color{blue} \underbrace{\| \nabla f(\theta_t) \|^2}_{=\text{(a)}}}
    + \alpha_t^2 {\color{red} \underbrace{\frac{L}{2} \mathbb{E} \| g_{i_t} \|^2}_{=\text{(b)}}},
\end{align}
because $\nabla f(\theta_t) = \mathbb{E}_{i_t} \left[ g_{i_t} \right]$.

There are two terms that are both positive but have opposing signs. The first term (a) is good news. It states that on expectation we would make a positive progress, that is, to lower the expected value after a stochastic gradient step. Since this term is multiplied with $\alpha_t$, we may be tempted to simply set $\alpha_t$ to a large value to make a big improvement on expectation. Unfortunately, this is not the case because of the second term (b). 

Although the stochastic gradient is an unbiased estimate of the full gradient, it is still a noisy estimate. The second term (b) reflect this noise. Imagine we are close to the/a minimum of $f$ such that $\nabla f (\theta_t)= 0$.  The second term (b) is then the trace of the covariance of the stochastic gradient. Because it is not zero (i.e. noisy), stochastic gradient {\it descent} will not decrease the objective function on expectation but may increase it. 

In order to control away the second term (b), we must ensure that $\alpha_t$ is small enough so that $\alpha_t \gg \alpha_t^2$, or must assume further constraints on $f$.  If we decrease $\alpha_t$ over $t$ , stochastic gradient descent will on expectation make progress (i.e., descent) and eventually passes by the/a minimum of $f$. More details on the convergence rate(s) of stochastic gradient descent are out of the scope of this course.

In summary, we use stochastic gradient descent in modern machine learning, and with a small learning rate stochastic gradient descent exhibits the descent property on expectation. We will therefore worry less and rely on stochastic gradient descent throughout the course. 

\subsection{Adaptive Learning Rate Methods}

Although we have approached the problem of stochastic optimization by stating that we follow the (negative) stochastic gradient estimate at each update, it is not necessarily the only way to view this problem. We can instead view the problem of learning as online optimization. In online optimization, or online learning, we play a game in which at each turn $t$ we receive the stochastic gradient estimate $g_t = \nabla_{\theta} f_{i_t} (\theta_{t-1})$ and use it to update our estimate of the parameters, $\theta_{t-1}, g_t \to \theta_t$. We receive the penalty as the difference between the stochastic estimate of the loss at the updated parameter and that at the optimal parameter configuration,\footnote{
    The optimality in this context of online adaptation is defined as the final solution reached by the online optimization procedure. If we follow the direction that is correlated with the gradient, we know that we are making progress on average toward the local extreme configuration due to the decent lemma above. We thus know that asymptotically the optimal solution here $\theta^*$ would have a lower loss than any other intermediate points. This makes the online learning perspective different from the optimization perspective from above.
} 
i.e., $f_{i_t}(\theta_t) - f_{i_t}(\theta^*)$. We call this penalty a {\it regret}, since this quantifies how much better we could've done in hindsight (that is, regret.) The goal is to minimize the regret over time:
\begin{align}
    R(T) = \sum_{t=1}^T \underbrace{f_{i_t}(\theta_{t}) - f_{i_t}(\theta^*)}_{\geq 0}.
\end{align}
The regret must grow sub-linearly, i.e, $R(T) = o(T)$, since linear growth, i.e., $R(T) = O(T)$, implies that the learning algorithm is not converging toward the optimal solution (or its associated minimum value.) 

We (try to) achieve this goal by finding an appropriate update rule that maps $\theta_{t-1}$ and $g_t$ to $\theta_t$. In doing so, it is relatively straightforward to think of the following simplified framework, that generalizes stochastic gradient descent:
\begin{align}
    \theta_t \leftarrow \theta_{t-1} + \eta_t \odot g_t,
\end{align}
where $\eta_t$ is a collection of learning rates for all parameters.\footnote{
    It is possible to use a matrix $\eta_t$ instead of a vector $\eta_t$, and there could be a good chance that we would achieve a better regret bound. Unfortunately, this could increase the computationally complexity dramatically for each update, from $O(|\theta|)$ to $O(|\theta|^2)$, which can be prohibitive in many modern applications.
} 
By adapting $\eta_t$ appropriately, we can achieve the sublinear regret. In SGD above, $\eta_t$ was often a scalar, i.e. $\eta_t^i = \eta_t^j$ for all $i \neq j$. SGD in fact achieves the sublinear regret, $O(\sqrt{T})$ with $\eta_t = \frac{1}{\sqrt{T}}$, but it turned out that we can do better either asymptotically or practically by taking into account the loss function landscape, that is, how the loss changes w.r.t. the parameters, more carefully.

\paragraph{Adagrad~\citep{duchi2011adaptive}.}

For each parameter $\theta^i$, the magnitude of the partial derivative of the loss, $(g_t^i)^2$, tells us how sensitive the loss value was to the change in $\theta^i$. Or, another way is to view it as the impact of the change in $\theta^i$ on the loss. By accumulating this over time, $\sqrt{\sum_{t'=1}^t (g_{t'}^i)^2}$, we can measure the overall impact of $\theta^i$ on the loss. We can then normalize each update inverse-proportionally in order to ensure each and every parameter has more or less the equal impact on the loss. That is,
\begin{align}
    \label{eq:adagrad}
    \theta_t \leftarrow \theta_{t-1} + 
    \begin{bmatrix}
        \frac{1}{\sqrt{\sum_{t'=1}^t (g_{t'}^1)^2}} \\
        \vdots \\
        \frac{1}{\sqrt{\sum_{t'=1}^t (g_{t'}^{|\theta|})^2}}
    \end{bmatrix}
    \odot
    g_t
\end{align}

The regret of Adagrad is $O(\sqrt{T})$, just like that of SGD, assuming $\| g_t \| \ll \infty$. It however often decreases faster especially when many parameters are inconsequential (\textit{sparse}) and/or quickly learned (because the accumulated magnitude rapidly grows and its inverse converges to zero quickly.) 

\paragraph{Adam~\citep{kingma2014adam}.}

A major disadvantage of Adagrad above is that the per-parameter learning rate decays monotonically, often resulting in a premature termination. This is especially problematic with a non-convex optimization problem, such as the ones in training deep neural networks, as it may require many updates for the optimizer to get close enough to a good solution in the parameter space. We can address it by not accumulating the magnitude of the gradient over the full duration but using exponential smoothing:
\begin{align}
    v_t \leftarrow \beta_v v_{t-1} + (1-\beta_v) g_t^2,
\end{align}
where $\beta_v \in [0, 1]$. Then, we use $\sqrt{v_t}$ as the learning rate instead, leading to
\begin{align}
    \theta_t^i \leftarrow \theta_{t-1}^i + \frac{g_t^i}{\sqrt{v_t^i + \epsilon}},
\end{align}
where $\epsilon > 0$ is a small scalar to prevent the degenerate case.

Adam furthermore uses exponential smoothing to reduce the variance of the gradient estimate as well:
\begin{align}
    m_t \leftarrow \beta_m m_{t-1} + (1-\beta_m) g_t.
\end{align}
This results in the following final update rule:
\begin{align}
    \theta_t^i \leftarrow \theta_{t-1}^i + \alpha \frac{m_t^i}{\sqrt{v_t^i + \epsilon}},
\end{align}
where $\alpha \in (0, 1]$ is a default step size. 

Adam also has $O(\sqrt{T})$ regret and exhibits an overall similar asymptotic behaviour to Adagrad. Adam is however often favoured over Adagrad, because the per-parameter learning rate is not monotonically decreasing anymore. Since it was proposed earlier, there have been a number of improvements to Adam, although they are out of the scope of this course.

Overall, whenever we refer to stochastic gradient descent in the rest of the course, we are generally referring to Adam or its variants that adaptively update the learning rate of each parameter on the fly. Although it is just a folk wisdom, quite a few researchers, including myself, attribute the recently-observed surprising successes of many conventional machine learning algorithms with gradient-descent optimization to these adaptive learning rate algorithms.

\section{Generalization and Model Selection}

\subsection{Expected risk vs. empirical risk: a generalization bound}
\label{sec:gen-bound}

A risk is another word we use to refer to the loss. In this section, we will use {\it risk} instead of {\it loss}, as the former is more often used in this particular context. If you are confused by the term ``risk'', simply read it out loud as ``loss'' whenever you run into it.

For each example $(x,y)$, we now know how to construct an energy function and also an associate loss function $L([x,y], \theta)$. Let $p_\mathrm{data}(x,y)$  be some unknown distribution from which we draw an example $(x,y)$. We do not know what this distribution is, but we assume that this is the distribution from which the training examples were drawn and any future instance would be drawn as well.\footnote{
    This is certainly not true in reality but is a reasonable starting point. We will discuss later in the course what we can do if this assumption does not hold, hopefully if time permits.
} Then, our goal must be to minimize
\begin{align}
    \label{eq:expected-risk}
    R(\theta) = \mathbb{E}_{\mathrm{data}} \left[ L([x,y], \theta) \right].
\end{align}

Unfortunately, this expected risk is not computable, and we only have access to a sample-based proxy to the expected risk, called the empirical risk:
\begin{align}
\label{eq:empirical-risk}
    \hat{R}(\theta) = \frac{1}{N} \sum_{n=1}^N L([x^n, y^n], \theta).
\end{align}
For brevity and clarity, let $S_n = \sum_{k=1}^n L([x^k, y^k], \theta)$. Then, we can express these risks as
\begin{align}
R(\theta) = \mathbb{E}_{\mathrm{data} \times \cdots \times \mathrm{data}} 
\left[ \frac{1}{N} S_N \right], \text{ and } \hat{R}(\theta) = \frac{1}{N} S_N.
\end{align}
The former holds because each instance $(x,y)$ is drawn independently from the same data distribution.

Let's assume that an individual loss is bounded between $0$ and $1$ (which would be the case for the 0-1 loss.) Then, we can use the Hoeffding's inequality to get
\begin{align}
    p( | R(\theta) - \hat{R}(\theta) | \geq \epsilon) \leq 
    2\exp \left( - \frac{2 (N \epsilon)^2}{\sum_{n=1}^N (1 - 0)^2} \right)
    = 2 \exp \left( - 2 N \epsilon^2 \right).
\end{align}
This inequality tells us that the gap between the expected and empirical risks shrinks exponentially with $N$, the number of training examples we use to compute the empirical risk. This inequality applies to any $\theta$, implying that this convergence of the empirical risk toward the expected risk is uniform over the parameter space (or the corresponding classifier space.) Such uniform convergence is nice in that we do not have to worry about how well learning works (that is, what kind of solution we end up with after optimization), in order to determine how much deviation we would anticipate between the empirical risk (the one we can compute) and the expected risk at any $\theta$. On the other hand, there is a big question of whether we actually care about most of the parameter space; it is likely that we do not and we only care about a small subset of the parameter space over which iterative optimization, such as stochastic gradient descent, explores. We will discuss this a bit more later, but for now, let's assume we are happy with this uniform convergence.\footnote{In practice, we are not.} 

Let's imagine that someone (or some learning algorithm) gave me $\theta$ that is supposed to be good with a particular empirical risk $\hat{R}(\theta)$. Is there any way for me to check how much worse the expected risk $R(\theta)$ would be, based on the Hoeffding's inequality above? Of course, such a statement would have to be probabilistic, since we are working with random variables, $R(\theta)$ and $\hat{R}(\theta)$. 

The inequality above allows us to express that
\begin{align}
| R(\theta) - \hat{R}(\theta) | < \epsilon
\end{align}
with some probability at least $1-\delta$. Be aware that the direction of the inequality has flipped.

If $| R(\theta) - \hat{R}(\theta) | < \epsilon$, we know that $R(\theta) < \hat{R}(\theta) + \epsilon$. We are interested in this latter inequality, because we want to upper-bound the expected (true) risk. If the true risk was lower than the empirical risk, we are happy and do not care about it. We want to know if we were to be unhappy (that is, the expected risk was greater than the empirical risk), how unhappy we would be in the worst case. 

Because we want to make such a statement with the probability of at least $1-\delta$, we equate the right-hand side above with $\delta$:
\begin{align}
    &2\exp(-2N\epsilon^2) = \delta \\
    \iff& -2N \epsilon^2 = \log \frac{\delta}{2} \\
    \iff& \epsilon^2 = \frac{1}{2N} \log \frac{2}{\delta} \\
    \iff& \epsilon = \sqrt{\frac{1}{2N} \log \frac{2}{\delta}}.
\end{align}

Combining these two together, we can now state that with probability at least $1-\delta$, we have
\begin{align}
\label{eq:basic-gen-bound}
R(\theta) < \hat{R}(\theta) + \sqrt{\frac{1}{2N} \log \frac{2}{\delta}}.
\end{align}
given the model parameter $\theta$. 

This generalization bound makes sense. If we want to get a strong guarantee, i.e., $(1-\delta) \to 1$ (equivalently $\delta \to 0$), we end up with a much loser bound, since the bound is $O(\sqrt{\log \frac{1}{\delta}})$. We can counter this by collecting more training examples, i.e., $N \to \infty$, since the bound shrinks rapidly as $N$ grows: $O(N^{-\frac{1}{2}})$.

This bound looks reasonable, but there is a catch. The catch is that this is based on a single, given model $\theta$. In other words, this bound is too optimistic, as in reality, we often need to choose $\theta$ ourselves among many alternatives by the process of learning. In doing so, we need to consider the possibility that we somehow picked one that has the worst generalization gap $|R(\theta) - \hat{R}(\theta)|$. In other words, we need to consider the generalization bounds of all possible model parameters. 

For simplicity, we assume that $\theta \in \Theta$ where $\Theta$ is a finite set of size $K$. Learning is then a process of selecting one of $K$ possible parameter configurations based on data. We use the idea of so-called \textit{union bound} from the basic probability theory, which states that
\begin{align}
    p(e_1 \cup e_2 \cup \cdots \cup e_N) \leq \sum_{i=1}^N p(e_i). 
\end{align}
This is somewhat obvious, because a pair $(e_i, e_j)$ may not be mutually exclusive. Think of a Venn diagram. With this, we want to compute
\begin{align}
    p(\cup_{\theta \in \Theta} |R(\theta) - \hat{R}(\theta) | \geq \epsilon) 
    \leq 
    \sum_{\theta \in \Theta} p(|R(\theta) - \hat{R}(\theta) | \geq \epsilon) 
    \leq 
    \underbrace{2 |\Theta| \exp \left( - 2 N \epsilon^2 \right)}_{=2 \exp \left( \log |\Theta| - 2 N \epsilon^2 \right)}. 
\end{align}
We can follow the exactly same logic above:
\begin{align}
    &2\exp(\log | \Theta|-2N\epsilon^2) = \delta \\
    \iff &\epsilon = \sqrt{
    \frac{\log |\Theta| -\log 2\delta }
    {2N}
    }.
\end{align}

This makes sense, as the generalization bound now depends on the size of $\Theta$, our hypothesis space. If the hypothesis set is large, there is a greater chance of us finding a solution that is good empirically $\hat{R}(\theta) \downarrow$ but is on expectation very bad $R(\theta) \uparrow$. This also implies that we need $N$ (the number of training examples) to grow exponentially w.r.t. the size of the hypothesis space $\Theta$. 

This bound only works with a finite-size hypothesis set $\Theta$ without favouring any particular parameter configuration. In order to work with an infinitely large hypothesis set, we must come up with different approaches. For instance, the Vapnik–Chervonenkis (VC) dimension can be used to bound the complexity of the infinitely large hypothesis set~\citep{VapnikChervonenkis1971}. Or, we can use the PAC-Bayes bound, where a prior distribution over the (potentially infinitely large) hypothesis set is introduced~\citep{McAllester1999}. These are all out of the scope of this course, but we briefly touch upon the idea of PAC-Bayes bound here before ending this section.

\paragraph{PAC-Bayesian bound.}

The original PAC-Bayes result states that
\begin{align}
\label{eq:mccallester}
    D_{\mathrm{KL}}(\mathcal{B}(\hat{R}(Q)) \| \mathcal{B}(R(Q))) \leq
    \frac{1}{N}
    \left(
    D_{\mathrm{KL}}(Q \| P) + \log \frac{N+1}{\delta}
    \right)
\end{align}
with probability at least $1-\delta$. Although this inequality looks quite dense, these terms are extremely descriptive, once we define and learn how to read them.

First, $R(Q)$ and $\hat{R}(Q)$ are defined analogously to $R(\theta)$ and $\hat{R}(\theta)$, except that we marginalize out $\theta$ using the so-called posterior distribution $Q(\theta)$. That is,
\begin{align}
    &R(Q) = \mathbb{E}_{Q}\left[ R(\theta) \right] \\
    &\hat{R}(Q) = \mathbb{E}_Q \left[ \hat{R}(\theta) \right].
\end{align}
$Q$ can be any distribution and can depend on data $D$ consisting of $N$ examples. 

Because we continue to assume we work with a bounded loss, we can assume that $R(Q) \in [0, 1]$ and $\hat{R}(Q) \in [0,1]$. 
Then, we can define Bernoulli distributions using these two values as the means. We denote these distributions as 
$\mathcal{B}(R(Q))$ and $\mathcal{B}(\hat{R}(Q))$, respectively. You can think of these distributions as how expected and empirical risks vary as $\theta$ follows the distribution $Q$. We can then measure the discrepancy between these two quantities, which is by definition the generalization gap, by using KL divergence. This is the left-hand side of the inequality above. 

The right-hand side is then the bound on how much discrepancy between the empirical and expected risks there could be on average given $Q$. 
There are two terms here. The first term is the KL divergence between the posterior $Q$ and the so-called prior $P$, where $P$ is constrained to be independent of data $D$. You can think of $P$ as our prior belief about which parameter $\theta$ would be good. On the other hand $Q$ is our belief after observing the data $D$. The first term therefore states that the discrepancy will be greater if our prior belief was incorrect, that is, our belief after observing data changed dramatically from the prior belief. This effect will however vanish rapidly as the number of training examples increases due to $\frac{1}{N}$.

We can read two things from the second term $\frac{1}{N} \log \frac{N + 1}{\delta}$. Because $\delta$ is in the denominator, we know that we would potentially get a greater discrepancy if we want to get a stronger guarantee, that is, $\delta \to 0$. $\frac{\log(N+1)}{N}$ vanishes toward $0$ as the data size increases, i.e. $N\to \infty$. The rate of this convergence is however quite slow, i.e. sublinear. 

Similarly to what we did earlier, we can turn this inequality in Eq.~\eqref{eq:mccallester} into a generalization bound. In particular, we use the Pinsker's inequality. In our case with Bernoulli random variables, we get
\begin{align}
    \label{eq:pinsker}
    \left( \hat{R}(Q) - R(Q) \right)^2 \leq \frac{1}{2} D_{\mathrm{KL}}(\mathcal{B}(\hat{R}(Q)) \| \mathcal{B}(R(Q))).
\end{align}
Then,
\begin{align}
    | \hat{R}(Q) - R(Q)| \leq
    \sqrt{\frac{1}{2N}
    \left(
    D_{\mathrm{KL}}(Q \| P) + \log \frac{N+1}{\delta}
    \right)
    }.
\end{align}
We end up with the following generalization bound:
\begin{align}
    R(Q) \leq \hat{R}(Q) + \sqrt{\frac{1}{2N}
    \left(
    D_{\mathrm{KL}}(Q \| P) + \log \frac{N+1}{\delta}
    \right)
    }.
\end{align}

Unlike the earlier generalization bound, and its variants, this PAC-Bayesian bound provides us with more actionable insights. First, we want the posterior distribution $Q$ to be {\it good} in that it results in a lower empirical risk on average. It sounds obvious, but the earlier generalization bound was designed to work with any parameter configuration (uniform convergence) and did not tell us what it means to choose a good parameter configuration. With the PAC-Bayesian bound, we already know that we want to {\it choose} the parameter configuration so that the empirical risk is low on average. In other words, we should use a good learning algorithm.

The posterior distribution $Q$ however cannot be too far away from where we start from. As the bound is a function of the discrepancy between $Q$ and our prior belief $P$. Flipping the coin around, it also states that we must choose our prior $P$ so that it puts high probabilities on parameter configurations that are likely to be probable under the posterior distribution $Q$. In other words, we want to ensure that we need a minimum amount of work to go from $P$ to $Q$, in order to minimize the generalization bound. 

In summary, the PAC-Bayesian bound tells us that we should have some good prior knowledge of the problem and that we should not train a predictive model too much, thereby ensuring that the posterior distribution $Q$ stays close to the prior distribution $P$. This will ensure that the expected risk does not deviate too much from the empirical risk.

\subsection{Bias, Variance and Uncertainty}
\label{sec:bias-variance}

An alternative way to write the 0-1 loss is to rely on the squared difference between the true label and predicted label, in the case of binary classification, where there are only two categories to which the input may belong. Let us use $y \in \{-1, 1\}$ to indicate two classes. Then,
\begin{align}
    \label{eq:0-1-loss-binary}
    L([x, y], \theta) = 
    \frac{1}{4} (y - \hat{y}(x, \theta))^2, 
\end{align}
where $\hat{y}(x, \theta) = \arg\min_{c \in \{-1, 1\}} e([x, c], \theta)$. If $y$ and $\hat{y}$ are the same, this loss is zero. Otherwise, it is 

As we discussed earlier, an instance $(x,y)$ is drawn from an underlying data distribution $p_{\mathrm{data}}(x,y)$ which can be written down as 
\begin{align}
p_{\mathrm{data}}(x, y) = p_{\mathrm{data}}(x)p_{\mathrm{data}}(y|x) 
\end{align}
following the definition of conditional probability.

We can furthermore imagine a distribution over $\theta$ as well: $q(\theta)$. This distribution may be to have come out of nowhere. It is however only natural to have a distribution over $\theta$ rather than a single value of $\theta$ if we realize that learning always depends on some randomness, either due to arbitrary symmetry breaking in optimization, random sampling of training examples or sometimes the lack of technical capabilities in reducing noise in our systems. We will discuss this uncertainty in the model parameters in depth later, and for now, we assume that this $q(\theta)$ is given to us.

We can then write down the expected 0-1 loss for binary classification under this (unknown) data distribution over the model parameters as:
\begin{align}
\label{eq:bias-variance-decomposition}
    \frac{1}{4} \mathbb{E}_{x, y, \theta}
    (y - \hat{y}(x, \theta))^2 &\propto 
    \mathbb{E}_{x}
    \left[
    \mathbb{E}_{y|x} \left[ (y - \mu_y )^2 \right]
    +\mu_y^2 
    \right.
    \\
    &\quad\quad\quad
    +
    \mathbb{E}_{\theta} \left[ (\hat{y}(x, \theta) - \hat{\mu}_y)^2 \right]
    + \hat{\mu}_y^2 
    \\
    &\quad\quad\quad
    \left.
    - 
    2 
    \mathbb{E}_{y|x} \mathbb{E}_{\theta}
    \left[ (y - \mu_y) (\hat{y}(x, \theta) - \hat{\mu}_y) \right]
    -2 \mu_y \hat{\mu}_y
    \right] 
    \\
    &=\mathbb{E}_{x}
    \left[ 
    \underbrace{\mathbb{E}_{y|x} \left[ (y - \mu_y )^2 \right]}_{=\text{(a)}}
    +
    \underbrace{\mathbb{E}_{\theta} \left[ (\hat{y}(x, \theta) - \hat{\mu}_y)^2 \right]}_{=\text{(b)}}
    \right.
    \\
    &\quad\quad\quad
    \left.
    -
    2 \underbrace{\cancel{\mathbb{E}_{y|x} \mathbb{E}_{\theta}
    \left[ (y - \mu_y) (\hat{y}(x, \theta) - \hat{\mu}_y) \right]}}_{=0}
    + \underbrace{(\mu_y - \hat{\mu}_y)^2}_{=\text{(c)}}
    \right],
\end{align}
where 
\begin{align}
    &\mu_y = \mathbb{E}_{y|x} \left[y\right],
    \\
    &\hat{\mu}_y = \mathbb{E}_\theta \left[\hat{y}(x, \theta)\right].
\end{align}

There are so many terms we need to consider in this equation, but we will consider them one at a time, from the back. First, let us start with $(\mu_y - \hat{\mu}_y)^2$. This term (c) tells us about how well our learner captures the mean of the true output $y$. This term does not care about how much variance there is either under the data distribution $p_{\mathrm{data}}(y|x)$ nor under the model distribution $q(\theta)$. It only talks about getting the outcome correct on average. This term is referred to as a \textit{bias}. When this term (c) is zero, we call our predictor unbiased. 

The second term from the back, which is zero, is the (negative) covariance between the true outcome $y$ and the predicted one $\hat{y}(x, \theta)$, both of which are random variables. Because we did not assume anything about $q(\theta)$, in general we cannot assume $\theta$ is in anyway correlated with $y|x$, implying that there should not be any covariance. We can ignore this term.

Let us continue with the two remaining terms, (a) and (b). The first term (a) is the variance of the true outcome $y$. This reflect inherent uncertainty present in the true outcome given an input $x$. This inherent uncertainty cannot be reduced, since it is not what we control but is given to us by the nature of the problem we are tackling. If this quantity is large, there is only so much we can do. We often refer to this as aleatoric uncertainty or irreducible uncertainty. 

The second term (b) is also uncertainty, as it measures the variance arising from the uncertainty in the model parameters. This uncertainty is however controllable and thereby reducible with efforts, since it arises from our uncertainty $q(\theta)$ in choosing the parameters $\theta$. When the model is simpler, we tend to have a better grasp at learning and can reduce this reducible (or epistemic) uncertainty greatly. When the model is complex and thereby exhibits many symmetries that must be broken arbitrarily, it is difficult (if not impossible) to reduce this epistemic uncertainty much. This term is often referred to as {\it variance}.

It should be quite clear at this point that there must be some inherent trade off between the bias (c) and the variance (b). The more complex a classifier is the higher variance we end up with, but due to its complexity, it would be able to fit data well, resulting in a lower bias. When a classifier is simple, the variance will be lower, but the bias will be higher. Learning can thus be thought of as finding a good balance between these two competing quantities.\footnote{
    I must emphasize here that the complexity of a classifier is not easy to quantify. When I speak of `complex' or `simple' here, I am referring to this mythical measure of the classifier's complexity and do not mean that we can compute it easily. 
}

The explanation above is slightly different from a usual way in which \textit{bias-variance tradeoff} is described~\citep{wiki:BiasVarianceTradeoff}. In particular, we are considering a generic distribution $q(\theta)$ that may or may not be directly related to any particular training dataset, when the conventional approach often sticks to the strong dependence on the training dataset and the distribution over the training set. This is a minor difference, but this can come in handy when we start thinking about more exotic ways by which we come with $q(\theta)$. If time and space permits later in the course, we may learn one or two techniques that involve such exotic techniques, such as transfer learning and multi-task learning.

\subsection{Uncertainty in the error rate}
\label{sec:error-rate}

We first need to talk about \textit{random variables}. In probability courses you probably have taken earlier, you must have learned about the strict distinction between random variables and non-random variables. In fact, a random variable does not take any particular value but carries with it a probability distribution over all possible values it can take. Once we draw a sample from this distribution, this value is \textit{not} random anymore but is deterministic. 

It unfortunately becomes easily cumbersome to explicitly distinguish between random variables and the samples drawn from their distributions. That is one of the reasons why we have not explicitly stated whether any particular variable is random or not so far. Another reason, perhaps more important, is that almost every variable in machine learning is random, because almost every variable depends on a set of samples drawn from an unknown underlying distribution. For instance, the parameters $\theta$ are random, because either they were initialized by drawing a sample from a so-called prior distribution, or because they were updated using a stochastic gradient estimate that is a function of a set of samples drawn from the data distribution. From this perspective, in fact, prediction $
\hat{y}$ we make using a model parametrized with $\theta$ is a random variable as well. The loss, or the risk, is thereby a random variable, as we have seen in \S\ref{sec:gen-bound}. 

\paragraph{Confidence interval: capturing test set variation.}

Let us stick to the zero-one loss (although this is not strictly necessary, it makes the following argument easier to follow.) The loss $l$ is a function of (1) a particular observation $[x,y]$ drawn from the data distribution $p_{\mathrm{data}}$ and (2) the parameters $\theta$. Both of these are sources of randomness, but for now, let's assume that $\theta$ is given to us as a fixed value, rather than as a random variable with a distribution attached to it. If we assume to have access to $N$ test examples, that were independently drawn from the identical distribution $p_{\mathrm{data}}$, we have 
\begin{align}
    (l_1, l_2, \ldots, l_N),
\end{align}
where each $l_n$ is itself a random variable. Each and every one of these $N$ random variables follows the same distribution. Because these all follow the same distribution, they also share the mean and variance:
\begin{align}
    \mu = \mathbb{E}[l]~\text{and}~\sigma^2 = \mathbb{V}[l] < \infty,
\end{align}
where we safely assume that the variance is finite. 

According to the central limit theorem, we then know that 
\begin{align}
\sqrt{N}( \bar{l}_N - \mu ) \to^{d} \mathcal{N}(0, \sigma^2),
\end{align}
where $\to^d$ refers to the convergence in distribution, and 
\begin{align}
\label{eq:test_set_accuracy}
\bar{l}_N = \frac{1}{N} \sum_{n=1}^N l_n.
\end{align}
$\bar{l}_N$ is a random variable that refers to the average loss computed over the $N$ examples. In other words, with larger $N$, we expect that the average accuracy we get from considering $N$ examples is centered at the true average $\mu$ with the variance $\frac{\sigma^2}{N}$. So, the more $N$ , the more confidence we have in trusting that the sample average does not deviate too much from the true average. With small $N$, however, we cannot be confident that our sample average accuracy is close enough to the true average, and this lack of confidence is proportional to the true variance underlying the accuracy. Unfortunately, we do not have access to the true variance of the accuracy but often can get a rough sense of it by considering the sample variance. 

If $N$ is large, we can compute the confidence interval\footnote{
    The confidence interval for a quantity with the confidence level $\gamma$ means that if we repeat the process of inferring the target quantity and measure the confidence interval, the true target quantity would be included in the confidence interval proportional to $\gamma$. 
} 
and use it to compare against another classifier or your prior expectation on the accuracy. For instance, because the accuracy estimate converges to the normal distribution, we can use so-called t-test, since the difference between the true mean and the mean of the estimate converges toward the Student's t distribution. In that case, the confidence interval for the binary accuracy (simply $1-l^\star$, where $l^\star$ is the true loss of the classifier) is given by
\begin{align}
    \mathrm{CI} \approx 
    \left[ 
        (1-\bar{l}_N) 
        - Z 
        \sqrt{\frac{\bar{l}_N (1-\bar{l}_N)}{N}},
                (1-\bar{l}_N) 
        + Z 
        \sqrt{\frac{\bar{l}_N (1-\bar{l}_N)}{N}}
    \right],
\end{align}
where $Z$ is determined based on the target confidence level $\gamma$. If $\gamma=0.99$, $Z$ would be approximately $2.576$.

Let $l_0$ be the accuracy by the existing classifier. We will assume this is the exact quantity because we have been running this classifier for a very long time. We can use this confidence interval to get some sense of whether we want to replace the existing classifier with this new one. If $l_0$ lies comfortably outside this confidence interval, we would feel more comfortable considering this option.

This approach focuses on estimating the error rate, and associated confidence, given a classifier $\theta$. In other words, the randomness we are considering stems from the choice of the test set $D$. If we repeatedly obtain new test sets and compute the associated confidence intervals, we anticipate the true accuracy to be included in the confidence interval approximately $\gamma$ times. This however tells us only one side of the story. Let us consider two additional aspects.

\paragraph{Credible interval: capturing model variations.}

There are quite a few factors that make our learning algorithm stochastic. First, our objective function tends to have many local minima, arising from reasons such as co-linear features and scaling invariance. For instance, if we use the zero-one loss, the following classifiers are all equivalent:
\begin{align}
    \hat{y} = \arg\max_{y=1, \ldots, |y|} \left(\alpha \left(W^\top x + b\right) \right)_y,\quad \alpha > 0,
\end{align}
where $\left(\cdot\right)_j$ refers to the $j$-th element of the vector, because the zero-one loss is invariant to the multiplicative scaling of the energy value. A pair of co-linear features are defined to have linear relationship given the target outcome. Imagine that
\begin{align}
    x_j = \alpha x_i,
\end{align}
when $y=c$. We then say that $(x_i, x_j)$ are co-linear given $y=c$. In this case, the following two energy functions are equivalent:
\begin{align}
    e([x, c], \theta) = -\left[ w_{c,1}, \ldots, w_{c,i}, \ldots, \underbrace{0}_{=w_{c,j}}, \ldots, w_{c,|x|} \right] x - b_c, \\
    e'([x, c], \theta') = -\left[ w_{c,1}, \ldots, \underbrace{0}_{=w_{c,i}}, \ldots, \frac{1}{\alpha} w_{c,i}, \ldots, w_{c,|x|} \right] x - b_c,
\end{align}
for any $\alpha \neq 0$. We cannot really distinguish these two energy functions. 

There are more of these, which we will touch upon over the rest of the course, and they all lead to the issue that our learner will pick one of these equivalent (or nearly equivalent) solutions at random. Such randomness arises from many factors, including stochastic initialization, stochastic construction of minibatches in stochastic gradient descent and even non-determinism in the implementation of underlying compute architectures. That is, learning is not really a deterministic process but a random process, resulting in a random $\hat{\theta}$. In other words, every time we train a model, we are effectively sampling $\hat{\theta}$ from a conditional distribution over a random variable $\theta$ given the training set $D$, i.e., $\hat{\theta} \sim p(\theta | D)$. This distribution is often referred as a posterior distribution, and if time permits, we will learn about this distribution more carefully in the context of Bayesian machine learning later.

We considered $\bar{l}_N$, the test set accuracy, in Eq.~\eqref{eq:test_set_accuracy} as a random variable whose stochasticity arose from the choice of the test set. Here we however consider it as a random variable whose randomness is induced by the choice of the parameters $\theta$ rather than the test set $D'$. This is understandable now that $\theta$ is a random variable rather than a given deterministic variable as before. We can then write the probability of $\bar{l}_N$ as 
\begin{align}
\label{eq:posterior_risk}
    p(\bar{l}_N | D, D') = \int p(\bar{l}_N | \theta, D') p(\theta | D) \mathrm{d}\theta,
\end{align}
where we safely assume $\theta$ is independent of the test set $D'$. 

It may be confusing to see $p(\bar{l}_N | \theta, D')$, since we often get one test accuracy (loss) once we have a model and a fixed test set. This is however not true in general, as running a model, that is performing $\arg\max$ on the energy function, is often either noisy on its own or computational intractable so that we must resort to some kind of randomization. 

We can then derive a so-called \textit{credible interval} of the test-set accuracy, such that the true test-set accuracy would be contained within this interval with the probability $\gamma$. Let $\gamma = 1 - \alpha$ for convenience. Then, we are looking for an interval $\left[l, u\right]$:
\begin{align}
    p(\bar{l}_N \leq l | D, D') = \frac{\alpha}{2}
    \quad\text{and}\quad
    p(\bar{l}_N \geq u | D, D') = \frac{\alpha}{2}.
\end{align}
This credible interval is reasonable when $p(\bar{l}_N|D, D')$ is unimodal, but this may not be the case. The probability density may be concentrated in two well-separated sub-regions, in which case this credible interval would be unnecessarily wide and uninformative. 

In that case, we can try to define a credible region $C$, which may not be contiguous. The credible region is define to satisfy
\begin{align}
    &\int_{\bar{l}_N \in C} p(\bar{l}_N | D, D') \mathrm{d}\bar{l}_N = \gamma, \\
    &p(\bar{l}_N | D, D') \geq p(\bar{l}'_N | D, D')\text{ for all } \bar{l}_N \in C \wedge \bar{l}_N' \notin C.
\end{align}
The second condition is often referred as \textit{density dominance}. Effectively, the credible region consists of one or more contiguous sub-regions such that no point within these sub-regions have lower densities than any other points outside these regions. By inspecting this credible region, we can get a good sense of how the true accuracy (or error) rate would be with the probability of $\gamma$.

In practice, we often cannot compute any of those quantities exactly, because the posterior distribution $\theta | D$ is tractable nor not even known. Instead, we use Monte Carlo approximation by training models many times, benefitting from the stochasticity in learning. Let $\left\{ \theta_1, \ldots, \theta_M \right\}$ be a set of resulting models. For each $\theta_m$, we draw a sample of the test loss $\bar{l}_N^m$, resulting in $\left\{ \bar{l}_N^1, \ldots, \bar{l}_N^M \right\}$. We can then use these samples to characterize, understand and analyze how the true test accuracy would be with the learning algorithm given the training and test sets, $D$ and $D'$.

\paragraph{Capturing training set variations.}

In addition to the randomness arising from the construction of the test set as well as the learning process itself, there is yet another source of randomness we want to take into account. This source of randomness arises from the construction of the training set $D$. If we continue from the credible region above, we do not want $p(\bar{l}_N | D, D')$ but rather 
\begin{align}
    p(\bar{l}) = \sum_{D} \sum_{D'} p(\bar{l}_{|D'|} | D, D') p(D) p(D') \mathrm{d}D \mathrm{d}D'.
\end{align}
In words, we want to check the variability of the test-set accuracy $\bar{l}$ after marginalizing out both training and test sets. Unfortunately, we often do not have access to the distribution over the dataset. Rather, we are only given a single dataset which is split into two data sets; one for training and the other for evaluation.

In this case, we can resort to the idea of so-called \textit{bootstrap resampling}. The idea is simple: (1) we resample $N$ examples from the original set of $N$ training examples with replacement, (2) compute the sample statistics of interest and (3) repeat (1-2) $M$ times. In step (2), we can split the resampled set into the resampled training set and the resampled test set. We use the resampled training set to train a model and then the resampled test set to evaluate the trained model to obtain $\bar{l}_{|D'|}^m$. After $M$ such iterations, we end up with  $\left\{ \bar{l}_N^{(m)} \right\}_{m=1}^M$. These sampled statistics then serve as a set of samples drawn from $p(\bar{l})$, allowing us to get a good sense of how the proposed learning algorithm works on this particular problem (\textit{not} a particular dataset).

There are many ways to characterize the uncertainty in evaluating how well any learning algorithm works. Although we have considered a few aspects of uncertainty we should consider in this section, there are many more ways to think of this problem. For instance, if we want to compare two learning algorithms, how should we take into account the uncertainty? If there is uncertainty in my learning algorithm, is there a better way to benefit from this uncertainty? We will touch upon some of these questions in the rest of the course.

\section{Hyperparameter Tuning: Model Selection}
\label{sec:hyperopt}

We often use the term `hyperparameter' to refer to anything that we can control in order to affect learning. For instance, in the case of stochastic gradient descent, a learning rate $\alpha$ (or any knobs in a learning rate scheduler) is a major hyperparameter. There are so many hyperparameters in machine learning. For instance, the parameters of the disdtribution one uses to initialize the model parameters are hyperparameters. The choice/parametrization of an energy function is yet another hyperparameter which is highly complex. We will use $\lambda$ to refer to the collection of all hyperparameters. 

We have learned so far that the model parameters $\theta$ should be estimated from data $D$. How should we then estimate the hyperparameters $\lambda$ ? We start by realizing that learning corresponds to
\begin{align}
    \mathrm{Learn}(D; \lambda, \epsilon) = \arg\min_{\theta} \hat{R}(\theta; D).
\end{align}
In other words, learning is the process of minimizing the empirical risk. This learning process is however not only a function of data $D$ but also of the hyperparameters $\lambda$ and noise $\epsilon$. 

We now need to find the right set of hyperparameters. What should be the objective function here? We can use a separate dataset $D_{\mathrm{val}}\cap D = \emptyset$, called a \textit{validation set}, to measure how good each hyperparamer set is:
\begin{align}
\label{eq:hyperopt}
    \mathrm{Tune}(D_{\mathrm{val}}, D;\epsilon')
    = \arg\min_{\lambda} \mathbb{E}_{\epsilon} \left[ \hat{R}(\mathrm{Learn}(D; \lambda, \epsilon); D_{\mathrm{val}})\right].
\end{align}
This hyperparameter tuning process is a function of both the training and validation sets as well as some source of noise $\epsilon'$. 

We can then obtain the final model by 
\begin{align}
    \hat{\theta} = \mathrm{Learn}(D; \mathrm{Tune}(D_{\mathrm{val}}, D; \epsilon'), \epsilon),
\end{align}
or
\begin{align}
    \hat{\theta} = \mathrm{Learn}(D \cup D_{\mathrm{val}}; \mathrm{Tune}(D_{\mathrm{val}}, D; \epsilon'), \epsilon).
\end{align}
We can furthermore obtain several such models by repeated sampling $\epsilon$.\footnote{
    We can also repeatedly sample $\epsilon'$ to obtain more than one set of good hyperparameters as well, but this process tends to be too expensive computationally to be practical, since we need to repeatedly train many new models for the purpose of optimization.
} We will learn about what we can do with such a case of having multiple models and what it means to have them later when we talk about Bayesian machine learning (if time permits) in \S\ref{sec:bayes}. 

The question is then how to implement and execute hyperparameter optimization in Eq.~\eqref{eq:hyperopt}. One could be tempted to use gradient-based optimization here as well, which is perfectly the right first reaction. There is however a major issue. We already saw this issue earlier and had to come up with stochastic gradient descent, and this issue is the computational cost of computing the gradient, since the gradient requires us to compute 
\begin{align}
    \mathrm{Jac}_\lambda \mathrm{Learn}(D; \lambda, \epsilon).
\end{align}
There are many different ways to approximate this quantity, such as forward-mode automatic differentiation as well as implicit function theorem. Nevertheless, this quantity is ultimately a fairly expensive quantity to compute due to many factors including the ever-increasing dataset size $|D|$ and thereby the ever-increasing optimization cost of learning. 

It is thus more usual to treat hyperparameter optimization as a black-box optimization problem, where we can evaluate the outcome (that is, the loss computed on the validation set) of a particular hyperparameter combination but cannot access anything else of this learning process. 

\textit{Random search} is one of the most widely used black-box optimization based approaches to hyperparameter optimization. In random search, we start by defining a prior distribution $p(\lambda)$ over the hyperparameters $\lambda$. We draw $K$ samples from this prior distribution, $\left\{ \lambda_1, \ldots, \lambda_K \right\}$, and in parallel evaluate them by training a model using each of these sampled hyperparameters. We then pick the best hyperparameter based on the validation risk, $r_k = \hat{R}(\mathrm{Learn}(D; \lambda_k, \epsilon); D_{\mathrm{val}})$. 

Instead of simply picking the best one, one can update the prior over the hyperparameters based on
\begin{align}
    \left\{ (\lambda_1, r_1) ,\ldots, (\lambda_K, r_K) \right\},
\end{align}
such that the probability is concentrated in the neighbourhood of low-risk hyperparameter configurations. The whole process can then be repeated using this updated distribution as the prior over the hyperparameters. This iterative approach is akin to the widely used method called the cross-entropy method~\citep{rubinstein2004cross}.

\subsection{Sequential model-based optimization for hyperparameter tuning}

Instead of drawing independent hyperparameter configurations, we can think of drawing a series of correlated hyperparameter configurations. Let $D_{n-1}=((\lambda_1, r_1), \ldots, (\lambda_{n-1}, r_{n-1}))$ be a series of hyperparameter configurations and their associated validation risks, selected and tested so far. At time $n$, we need to decide which hyperparameter to test next. This decision requires us to ask which criteria we want the next hyperparameter configuration to satisfy. There are many possible criteria, but one particular easy-to-understand criterion is \textit{expected improvement}. 

The expected improvement literally computes how much improvement we would see in the risk on expectation. This expectation is computed over the posterior distribution, similarly to Eq.~\eqref{eq:posterior_risk}:
\begin{align}
    p(r | \lambda, D_{n-1}) = \int p(r | \lambda, \theta) p(\theta | D_{n-1}) \mathrm{d}\theta.
\end{align}
$p(r|\lambda, \theta)$ is a model that predicts the output $r$ given the hyperparameter configuration $\lambda$, using the parameters $\theta$. See Eq.~\eqref{eq:l2-energy} and surrounding discussion on how to create such a model. The expected improvement of a hyperparameter configuration $\lambda$ is then defined as 
\begin{align}
    \mathrm{EI}(\lambda) = 
    \mathbb{E}_{r | \lambda, D_{n-1}} 
    \left[
    \max\left(
    0, \hat{r}_{n-1} - r
    \right)
    \right],
\end{align}
where 
\begin{align}
    \hat{r}_{n-1} = \min_{i=1, \ldots, n-1} r_{i}.
\end{align}
This can often be approximated using samples:
\begin{align}
    \mathrm{EI}(\lambda) \approx \frac{1}{M} \sum_{m=1}^M \max(0, \hat{r}_{n-1} - r_m),
\end{align}
where $r_m \sim r| \lambda, D_{n-1}$. 

We then want to draw the next hyperparameter configuration from the following distribution:
\begin{align}
    q(\lambda | D_{n-1}) \propto 
    \exp\left( \beta \mathrm{EI}(\lambda) \right),
\end{align}
where $\beta \geq 0$. When $\beta =0$, we recover the random search, and when $\beta \to \infty$, we always choose the hyperparameter configuration with the best expected improvement. It is however intractable often to search for the best hyperparameter configuration each time to maximize the expected improvement, and we only sample the next hyperparameter configuration proportionally to the expected improvement. 

When the number of hyperparameter is large, i.e. $|\lambda| \gg 1$, it can be challenging to sample exactly from this distribution. In that case, it makes sense to narrow down the space by make the density concentrated locally around the best hyperparameter so far:
\begin{align}
    q(\lambda | D_{n-1}) \propto \exp(\beta \mathrm{EI}(\lambda) - \alpha D(\lambda, \hat{\lambda}_{n-1}),
\end{align}
where  $\hat{\lambda}_{n-1}$ is the best hyperparameter configuration so far, and $D$ is a problem-specific distance metric. We can then readily sample from this distribution by first drawing a random set of samples in the neighbourhood of the best hyperparameter configuration so far and picking one of them proportionally to the expected improvement. This variation resembles iterative optimization, such as stochastic gradient descent. 

Overall, this approach, often called sequential model based optimization~\citep{jones1998efficient}, consists of repeating three steps; (1) fit an uncertainty-aware predictor of the risk given a hyperparameter configuration, (2) draw the next hyperparameter configuration that maximizes the expected improvement according to the trained predictor, and (3) test the newly selected hyperparameter configuration. Of course, it is easy to see that we do not have to test only one hyperparameter configuration at a time. Instead, we can draw many samples from the proposal distribution $q$, test all of them (by training multiple models and evaluating them on the validation set) and update the uncertainty-aware predictor on all accumulated pairs of hyperparameter configuration and associated validation risk. This approach has become \textit{de facto} standard when training a new deep neural network with many hyperparameters~\citep{bergstra2011algorithms}.

\subsection{We still need to report the test set accuracy separately}

The hyperparameter optimization algorithm above can be thought of as the implementation of $\mathrm{Tune}$ in 
\begin{align}
    \hat{\theta} = \mathrm{Learn}(D; \mathrm{Tune}(D_{\mathrm{val}}, D; \epsilon'), \epsilon),
\end{align}
Once we found the best hyperparameter configuration, we train the final model on the training set $D$ to obtain our final model parameter $\hat{\theta}$. How well would it work? 

Unfortunately, we cannot use the validation risk, as that was the objective by which $\hat{\theta}$ was selected. Meanwhile, when this model is deployed in the wild, the world will not be so kind and a set of examples thrown at this model will not be so perfect for the model. We thus need another set, called the test set, $D_{\mathrm{test}}$ in order to check the test accuracy. This set must be separate from both the training and validation sets, and we can report the risk on this set as is, or we can report more statistics, as we discussed earlier in \S\ref{sec:error-rate}.

\chapter{Building blocks of neural networks}
\label{sec:building-blocks}

Earlier in \S\Ref{sec:backprop}, we talked about how general transformation $F(x; \theta)$ can be. As an example back then, we considered
\begin{align}
    \label{eq:linear-block}
    F^\sigma_{\mathrm{linear}}(x; \theta) = \sigma(U^\top x + c),
\end{align}
where $\sigma$ is a point-wise nonlinearity such as a rectified linear unit:
\begin{align}
    \sigma(a) = \max(0, a).
\end{align}
By stacking this block repeatedly, we can create an increasingly more nonlinear transformation, which is the basic idea behind multi-layer perceptrons~\citep{rumelhart1986learning}. We often call such a nonlinear transformation function that consists of a stack of such nonlinear layers a \textit{deep neural network}.

This linear layer\footnote{
    Although this block is far from being linear, we often refer to this block as a linear block or a linear layer.
} is not the only option, although this is widely used due to its lack of inductive biases. That is, if we do not possess any particular knowledge about the input $x$, it is safe to treat it as a flat finite-dimensional vector and feed it through a stack of these linear layers. It is however often the case that we know about underlying structures of an observation. For instance, if we are dealing with a set of items as an observation, we want our transformation to be permutation equivariant or invariant, as there is no inherent order among the items within a set. 

In this (short) chapter, we will introduce a few more of these basic building blocks to build a deep neural network. In addition to these blocks, we can be as creative as possible as long as your newly designed blocks are differentiable w.r.t. both their own parameters and inputs. Some blocks may lack any parameters, and that is perfectly fine. For instance, I can have a block that simply reverses the order of items within an input in a deterministic manner.

\section{Normalization}

Let us consider the simple squared energy function from Eq.~\eqref{eq:l2-energy}, with an identity nonlinearlity:
\begin{align}
    e'([x, y], (u, c))
    =
    \frac{1}{2} ( u^\top x + c - y )^2.
\end{align}
We will further assume that $y$ is a scalar and thereby $u$ is a vector rather than a matrix.

The overall loss is then
\begin{align}
    J(\theta) = \frac{1}{N} \sum_{n=1}^N e'([x_n, y_n], (u, c)) = \frac{1}{2N} \sum_{n=1}^N (u^\top x_n + c - y_n)^2.
\end{align}

The gradient of the loss w.r.t. $u$ is then
\begin{align}
    \nabla_u = \frac{1}{N} \sum_{n=1}^N (u^\top x_n + c - y_n) x_n^\top, \\
    \nabla_c = \frac{1}{N} \sum_{n=1}^N (u^\top x_n + c - y_n).
\end{align}

So far, there is nothing different from our earlier exercises. We now consider the Hessian of the loss:
\begin{align}
    H = 
    \begin{bmatrix}
        \frac{1}{N} \sum_{n=1}^N x_n x_n^\top & \frac{1}{N}\sum_{n=1}^N x_n \\
        \frac{1}{N} \sum_{n=1}^N x_n & 1 
    \end{bmatrix}.
\end{align}

The Hessian matrix tells us about the curvature of the objective function and directly relates to the difficulty of optimization by a gradient-based approach. In particular, gradient-based optimization is more challenging when the condition number is larger, where the condition number is defined as
\begin{align}
    \kappa = \frac{|\max_i \lambda_i(H)|}{|\min_i \lambda_i(H)|}  \geq 1,
\end{align}
where $\lambda_i(H)$ is the $i$-th eigenvalue of $H$. 

It is out of scope of this course to discuss in depth why the condition number matters for optimization. At a high level, you can think of the eigenvalues of the Hessian of the objective function as quantifying how stretched out this function is along the associated eigenvector directions. That is, if the eigenvalue of an eigenvector is large, it means that the function value changes more dramatically along this eigenvector direction. When the objective value changes very differently across all these directions (orthogonal directions, as they are eigenvector directions of a symmetric matrix), stochastic gradient descent suffers, as it will easily oscillate along the directions with steep changes while it will not make much progress along the directions with only little changes. We thus want the eigenvalues of the Hessian to be similar to each other, for such an iterative optimization algorithm to work well. For more rigorous discussion, refer to your favourite convex optimization book \citep{nocedal2006numerical}. 

Based on this definition, an identity matrix has the minimal condition number. In other words, we can transform the Hessian matrix into the identity matrix, in order to facilitiate gradient-based optimization~\citep{lecun1998gradient}. In this particular case, because the Hessian matrix does not depend on $\theta$ but only on the observations $x_n$'s, we can simply transform the input in advance by
\begin{align}
    &\text{(1)}~x_n \leftarrow x_n - \frac{1}{N} \sum_{n'=1}^N x_{n'}&\text{(centering)}\\
    &\text{(2)}~x_n \leftarrow \left(\frac{1}{N} \sum_{n'=1}^N x_{n'} x_{n'}^\top \right)^{-\frac{1}{2}} x_n&\text{(whitening)}
\end{align}
This will result in the identity Hessian matrix, improving the convergence of gradient-based optimization.

Such normalization is a key to the success in optimization, but it is challenging to apply it in practice exactly, as the Hessian matrix is often non-stationary when we train a deep neural network. The Hessian matrix changes as we update the model parameters, and there is no tractable way to turn the Hessian matrix into the identity matrix. Furthermore, it is even more challenging to invert this Hessian matrix. It however turned out that normalizing (as a weaker version of whitening) of the input to each block helps in learning. Such normalization could also be considered as a building block, and let us look at a few widely used ones here.

\paragraph{Batch normalization~\citep{ioffe2015batch}.}

This is one of the building blocks that sparked the revolution in deep learning, greatly facilitating learning:
\begin{align}
    F_{\mathrm{batch-norm}}(x; \theta=(m, s))
    =
    m + \exp(s) \cdot \left( (x - \mu) \oslash \sigma \right),
\end{align}
where $\mu$ and $\sigma^2$ are the mean and diagonal covariance of the input to this block. Because the inverse of a full covariance matrix, which is often similar to the Hessian matrix up to an additive term, is costly, we are only consider the diagonal of the covariance matrix, which is readily invertible. 

Instead of using the full training set to estimate $\mu$ and $\sigma^2$ , which will be prohibitively expensive, we use the minibatch at each update during training to get stochastic estimates of these two quantities. This practice is perfectly fine during training but it becomes problematic when the model is deployed, as the model will receive one example at a time. With a single example, we cannot estimate either $\mu$ nor $\sigma^2$, or if we do, it will simply subtract out the input in its entirety. It is a usual practice instead to either fully re-estimate $\mu$ and $\sigma^2$ using the full training set once training is over or keep the running estimates of $\mu$ and $\sigma^2$ during training and use them after training is over.

\paragraph{Layer normalization~\citep{ba2016layer}.}

Instead of normalizing values across examples, it is possible to normalize values within each example across dimensions. When we do so, we call it layer normalization:
\begin{align}
    F_{\mathrm{layer-norm}}(x; \theta=(m, s)) 
    =
    m
    +
    \frac{\exp(s)}{\underbrace{\sqrt{\frac{1}{|x|} \sum_{i=1}^{|x|} (x_i - \mu)^2}}_{=\sigma}}
    \left(x - \underbrace{\frac{1}{|x|}\sum_{i=1}^{|x|}x_i}_{=\mu}\right),
\end{align}
where we assume $x$ is a finite-dimensional vector. We can certainly modify it to cope better with other types of the input, but that is out of the scope of this course. It is rather unclear why this should help with optimization, but it has been found to greatly facilitate learning in many large-scale experiments. 

Unlike batch normalization, one must be careful when using layer normalization, as it can easily break the relationships among different examples. For instance, imagine a simple binary classification problem, where the positive class consists of all input vectors whose Euclidean norms are less than 1 and the negative class of all input vectors whose Euclidean norms are greater than or equal to 1. Let $x_+ = [0.9,0]$ and $x_-=[2, 0]$. After layer normalization, they are transformed into $\hat{x}_+ = [0.5,-0.5]$  and $\hat{x}_- = [0.5, -0.5]$, respectively. Suddenly, these two inputs, which belong to two separate classes, are not distinguishable from each other. This happens, unlike with batch normalization, because normalization is applied differently to different instances, while batch normalization applies the same normalization to all instances simultaneously.

\section{Convolutional blocks}

Many problems in machine learning boil down to detecting patterns within an input that repeatedly appear within the training data set. Consider for instance an object detection algorithm. Initially we do not know what kind of patterns are considered representative of each object. Learning thus must figure out which patterns repeatedly appear whenever the input was associated with a particular object label. These patterns are however not global but localized, since the object may appear anywhere within an input image. It may appear at the center but also appear at any corner of the image, without any impact on the object identity. In other words, an object detector should be translation invariant.\footnote{
We say $F$ is \textit{equivariant} to a particular transformation $\mathcal{T}$ when
\begin{align}
    F(\mathcal{T}(X)) = \mathcal{T}(F(X)).
\end{align}
We say $F$ is \textit{invairant} to a particular transformation $\mathcal{T}$ when 
\begin{align}
    F(\mathcal{T}(X)) = F(X).
\end{align}
}

Any invariance could be implemented as a stack of equivariant blocks followed by a reduction operator, such as summation. We thus need to implement a translation equivariant block. In this section, we consider a so-called convolution block, or more precisely correlation block. 

We start by considering an infinitely long discretized time series, $x=\left[ \ldots, x_{t-1}, x_t, x_{t+1}, \ldots \right]$ with $|x| \to \infty$,  as an input to this block. Each item $x_t$ is a finite-dimensional real vector. The parameter of this block is a set of finite-length filter sequences, $f^k=[f_1^k, f_2^k, \ldots, f_{2M+1}^k]$ with $M \ll \infty$ and $k=1, \ldots, K$. Similarly to $x_t$, each $f_t^k$ is also a finite-dimensional real vector with $|f_t^k| = |x_t|$. The convolution block then returns an infinitely-long time series, $h=\left[\ldots, h_{t-1}, h_t, h_{t+1}, \ldots \right]$, where $|h_t| = K$. 

Let $h_t^k$ be the $k$-th element of $h_t$. We then compute it as
\begin{align}
    h_t^k = \sum_{m'=-M}^{m'=M} x_{t+m'}^\top f^k_{m'+M+1}.
\end{align}
In other words, we apply the $k$-th filter $f^k$ at each position $t$ to check how similar (in the sense of dot product) the signal centered at $t$ is to the filter $f^k$. 

Another way to write it down is 
\begin{align}
    h_t = \sum_{m'=-M}^{m'=M} F_{m'+M+1} x_{t+m'},
\end{align}
where 
\begin{align}
    F_{m} = 
    \begin{bmatrix}
        f_{m}^1 \\
        f_{m}^2 \\
        \vdots \\
        v_{m}^K
    \end{bmatrix} \in \mathbb{R}^{d \times K}
\end{align}
with $d = |x_t|$. The full parameters of this 1-D convolution block can be summarized as a 3-D tensor of size $d \times K \times (2M+1)$. 

It is pretty straightforward to see that this operation is translation equivariant. If we shift every $x_t$ by $\delta$, the resulting $h_t$ will shift by $\delta$ without any impact on its computed value. Unfortunately, in practice, this does not hold perfectly, as we do not work with an infinitely long sequence. We must decide how to handle the boundaries of the sequence with a finite-length sequence, and this choice will impact the degree of translation equivariance near the boundaries. Detailed discussion on how we handle boundaries is out of the scope of this course, though.

We can now readily extend this 1-D convolution to $N$-D convolution. For instance, 2-D convolution would work on an infinitely large image, and 3-D convolution on an infinitely large-and-long video. Furthermore, we can extend it by introducing various features, such as a stride. These are also out of the scope of this course, but I recommend the first half of the classic by \citet{lecun1998gradient}.

\section{Recurrent blocks}

Often, strong equivariance or invariance tends to be too strict. Perhaps we want equivariance only in a particular context and not in another context. It is however difficult to implement it in a strict sense. We can take one step up in the level of abstraction and work on applying the same operator repeatedly over the input. This is the core idea behind a recurrent block. 

A recurrent block works on a \textit{sequence} of input items $\left( x_1, x_2, \ldots, x_T \right)$, just like the 1-D convolution block above. This block consists of a neural network that is applied repeated to $x_t$ sequentially (that is, one at a time.) This neural net takes as input the concatenation of $x_t$ and the memory (or hidden state) $h_{t-1}$ and returns an updated memory $h_t$:
\begin{align}
    h_t = F([x_t, h_{t-1}]; \theta_r),
\end{align}
where $\theta_r$ is the parameters of this recurrent function $F$. $\theta_r$ includes the intial hidden state $h_0$. Once we sweep the input sequence with $F$, the recurrent block returns the same-length sequence by concatenating all $h_t$'s: $\left(h_1, h_2, \ldots, h_T\right)$.

The advantage of such a recurrent block over e.g. the 1-D convolution above is that it effectively has an unlimited context size. In the 1-D convolution, any output at time $t$ depends only on $2M+1$ input vectors centered at $t$. On the other hand, the recurrent block takes into account all inputs up to $t$ to compute the hidden state $h_t$ at time $t$. Furthermore, by simply stacking two recurrent blocks with the sequence reversal inbetween, we can make each output vector to depend on the entire sequence readily. 

A representative (and simple) example of widely-used (and easy-to-use) recurrent blocks is a gated recurrent unit~\citep[GRU;][]{cho2014learning} which is defined as 
\begin{align}
    F_{\mathrm{GRU}} = u_t \odot h_{t-1}
    + (1- u_t) \odot \tilde{h}_t,
\end{align}
where
\begin{align}
    &r_t = \sigma\left( W_r x_t + U_r h_{t-1} + b_r \right) &\text{(Reset Gate)}\\
    &u_t = \sigma\left(W_u x_t + U_u h_{t-1} + b_u \right) &\text{(Update Gate)} \\
    &\tilde{h}_t = \tanh\left(W_h x_t + U_h (r_t \odot h_{t-1}) + b_h \right) &\text{(Candidate State)}
\end{align}
This (weighted) linear combination has shown to effectively address the issue of vanishing gradient~\citep{bengio1994learning}, and has become a standard practice in machine learning over the past decade or so~\citep{he2016deep}.

\section{Permutation equivariance: attention}

We are often faced with a situation where the input to a block is a set of vectors $X = \left\{ x_i \right\}_{i=1}^{N}$. We want to transform each item $x_k$ in the context of all the other items in this set, resulting in another set of vector $H = \left\{ h_i \right\}_{i=1}^N$. We want this layer to be equivariant to permutation such that $F((x_{\sigma(i)} )_{i=1}^N) = (h_{\sigma(i)})_{i=1}^N$, where $\sigma: \left\{1, \ldots, N\right\} \to \left\{1, \ldots, N \right\}$ is a permutation operator. Let's consider one canonical way to build such a permutation equivariant block.

This block begins with three linear blocks:
\begin{align}
    &k_i = F_{\mathrm{linear}}(x_i; \theta_k), \\
    &q_i = F_{\mathrm{linear}}(x_i; \theta_q), \\
    &v_i = F_{\mathrm{linear}}(x_i; \theta_v).
\end{align}
We are referred to as the key, query and value vectors of the $i$-th item $x_i$. 

For each $j$-th item $x_j$, we check how compatible it is to the current $i$-th item $x_i$:
\begin{align}
\label{eq:attention-weight}
    \alpha_i^j = \frac{\exp(q_i^\top k_j)}{\sum_{j'=1}^N \exp(q_i^\top k_{j'})}.
\end{align}
We normalize it to sum to one using softmax.

Now, we use these importance weights to compute the weighted average of the values:
\begin{align}
    \hat{v}_i = \sum_{j=1}^N \alpha_i^j v_i.
\end{align}

It is a usual practice to repeat this process $K$ times to produce 
\begin{align}
    \hat{v}_i \leftarrow 
    \begin{bmatrix}
        \hat{v}_i^1 \\
        \vdots \\
        \hat{v}_i^K
    \end{bmatrix}.
\end{align}
When we do this, each of $K$ such processes is called an attention head. 

At this point, $\hat{v}_i$ is a linear function of the input $X=\left\{x_1, \ldots, x_N\right\}$. We want to introduces some nonlinearity here by introducing the final linear layer together with a residual connection:
\begin{align}
    h_i = 
    F_{\mathrm{layer-norm}}(F_{\mathrm{linear}}^\sigma(\hat{v}_i; \theta_h); \theta_l)
    + F_{\mathrm{linear}}(x_i; \theta_r), 
\end{align}
where the lack of the superscript in the second term means that there is no nonlinearity. If $|h_i| = |x_i|$, it is customary to fix $\theta_r=(I, 0)$. It is usual to add a layer normalization block after $\hat{v}_i$ or at $h_i$, to facilitate optimization.

When implemented in a single block, this block is often referred to as the (multi-headed) attention block~\citep{bahdanau2014neural,vaswani2017attention}.

\paragraph{Positional Encoding.}

Another way to think of the attention block above is to view it as a way to handle a variable-sized input. Regardless of the size of the input set, this attention block can work with the input. It is thus tempting to use the attention block for a variable-length sequence, which was the original motivation behind the attention block. There is one hurdle that must be overcome in that case. That is, we must ensure that each item in a sequence is marked with its position. 

There are two major approaches to this. The first approach is based on additive marking. For each position $i$, let $e_i$ be a vector of size $|x|$ and represent the $i$-th position. There are many ways to construct this vector, and sometimes it is even possible to learn this vector from data, although we can only handle the length seen during training in the latter case. One particular approach is to use sinusoidal functions so that each dimension of $e_i$ captures different rates at which the position changes. For instance,
\begin{align}
    e_i^d = \begin{cases}
        \sin\left(
        \frac{i}{L^{i/|x|}}
        \right),&\text{if } i \mod 2 = 0 \\
        \cos\left(
        \frac{i}{L^{(i-1)/|x|}}
        \right),&\text{if } i \mod 2 = 1
    \end{cases}
\end{align}
where $L$ is a hyperparameter and is often set to $10000$. This vector is then added to each input item, i.e., $x_i + e_i$ before being fed to the attention block.

The first approach, the additive approach, makes it easy for the attention block to capture the locality of each vector, because neary vectors tend to have similar positional embeddings, and to capture the absolute position based pattern, as each absolute position is represented by its unique positional embedding vector. It is however challenging for the attention block to capture the patterns based on relative positions beyond simple locality. 

In particular, consider how the so-called attention weight on the $j$-th item for the $i$-th item was computed in Eq.~\eqref{eq:attention-weight}. The weight is proportional to the dot product between the $i$-th query vector and the $j$-th key vector:
\begin{align}
    q_i^\top k_j &= (W_q (x_i+e_i))^\top (W_k (x_j+e_j)) 
    \\
    &= 
    x_i^\top W_q^\top W_k x_j + e_i^\top W_q^\top W_k x_j 
    + x_i^\top W_q^\top W_k e_k 
    + e_i^\top W_q^\top W_k e_k,
\end{align}
where we assumed zero bias vectors for both query and key vectors. From from the first term in the expanded expression, we notice that the content-based relationship between the $i$-th input and $j$-th input is largely independent of their positions. In other words, the semantic relationship between these two inputs is stationary across their relative positions, which may be restrictive in many downstream applications.

Focusing on the first term above, we can think of a way to ensuring that this pairwise semantic relationship is position-dependent. More specifically, we want it to depend on the relative position between $x_i$ and $x_j$:
\begin{align}
    \left<q_i, k_j\right>_{j-i} = q_i^\top R_i^\top R_j x_j,
\end{align}
where $R_m$ is an orthogonal matrix that is parameterized by a scalar position $m$ and changes smoothly w.r.t. $m$. One way to construct such an orthogonal matrix is to build a block diagonal matrix where a 2-D rotation matrix is repeated along the diagonal:
\begin{align}
    R_m =
    \begin{bmatrix}
        R^2_1(m) & 0 & \cdots & 0 \\
        0 & R^2_2(m) & \cdots & 0 \\
        0 & 0 & \cdots & 0 \\
        0 & \cdots & 0 & R^2_{|x|/2}(m)
    \end{bmatrix}, 
\end{align}
where $R^2_k(m)$ is a 2-dimensional rotation matrix that rotates a 2-dimensional real vector and defined as
\begin{align}
    R_k^2(m) = 
    \begin{bmatrix}
        \cos(m L^{k/|x|}) & -\sin(m L^{k/|x|}) \\
        \sin(m L^{k/|x|}) & \cos(m L^{k/|x|})
    \end{bmatrix}.
\end{align}
In other words, we rotate every pair of elements of the query/key vector based on its position before computing the dot product between these two vectors. Since this rotation depends on the relative position between the query and key vectors, this approach can capture position-dependent semantic relationship between the $i$-th input and the $j$-th input. This idea has become one of the standard approaches to incorporating positional information in the attention block in recent years~\citep{su2021roformer}.

\chapter{Probabilistic Machine Learning and Unsupervised Learning}
\label{sec:probabilistic-ml}

\section{Probabilistic interpretation of the energy function}

Although we already learned about how to turn an energy function into a probability function in \S\ref{sec:softmax}, we will go slightly deeper in this section, as it will help us derive a series of machine learning algorithms in this chapter. 

The energy function is defined w.r.t. the observation $x$, the unobserved (latent) variable $z$ and the model parameters $\theta$: $e(x, z, \theta)$. For now, we will assume that $\theta$ is \textit{not} a random variable, unlike $x$ and $z$. We can then compute the joint distribution over $x$ and $z$ as
\begin{align}
    p(x, z; \theta) = \frac{\exp(-e(x,z,\theta))}{\iint \exp(-e(x',z',\theta)) \mathrm{d}x' \mathrm{d}z'}.
\end{align}
Of course, it is often (if not almost always) challenging to compute the normalization constant (or the partition function) in the denominator. Such a challenge hints at a different approach to the same problem. Instead of defining an energy function first and then deriving the probability  function, why not directly define the probability function? After all, we can recover the underlying energy function given a probability function up to a constant:
\begin{align}
    e(x,z,\theta) = -\log p(x,z; \theta) + \log Z(\theta).
\end{align}
In fact, it may be even easier to decompose the joint probability function $p(x,z)$ further,\footnote{
    As usual, we will omit $\theta$ if its existence, or lack thereof, is clear from the context.
} using the chain rule of probability:
\begin{align}
    p(x,z) = p(z) p(x|z).
\end{align}

Such a decomposition gives us an interesting way to interpret this {\it probabilistic model}. $z$ is a latent variable that determines the intrinsic properties of the observation $x$. We therefore first draw an intrinsic property configuration $z$ from the {\it prior} distribution $p(z)$. Given this intrinsic property configuration $z$, we draw the actual observation $x$. 

For instance, you can imagine that $z$ refers to an object category (a dog, a cat, a car, etc.) We first draw a category of an object we want to paint by selecting $z$ according to the prior distribution $p(z)$. This prior distribution reflects the frequencies of these object categories in the world. Given the object category $z$, we can now paint the object by drawing $x$ from $p(x|z)$. This conditional distribution encapsulates all variations of the object $z$ in its visual form, such as lightning condition, background, texture, etc. 

Another distribution, or the probability function, of our interest is the posterior distribution over $z$ given the observation $x$. Continuing from the example above, we can think of trying to infer which object $z$ a given painting $x$ depicts. Such inference is often imperfect and results in a distribution over the object categories rather than picking one correct category. We can derive this distribution using the Bayes' rule:
\begin{align}
    p(z|x) = \frac{p(x|z)p(z)}{\int p(x|z')p(z') \mathrm{d}z'} = \frac{p(x|z)p(z)}{p(x)}.
\end{align}

Just like earlier when we tried to turn the energy function into a probability function, posterior inference is often computationally intractable due to the normalization constant in the denominator: $\int p(x|z')p(z') \mathrm{d}z'$.

With these probability functions in our hands, we can now define a generic loss function:
\begin{align}
    \label{eq:log-likelihood}
    L_\mathrm{ll}(x, \theta) = -\log \int p(x|z; \theta) p(z; \theta) \mathrm{d}z.
\end{align}
We often refer to this loss function as the negative log-likelihood or log-probability. 

If you are not comfortable with having a single-variable observation $x$, we can write this down in terms of the input-outcome pair $(x,y)$:
\begin{align}
    L_{\mathrm{ll}}([x, y], \theta) &= -\log \int p(y|x, z; \theta) p(x) p(z;\theta) \mathrm{d}z \\
    &=-\log \int p(y|x, z; \theta)p(z; \theta) \mathrm{d}z + \mathrm{const.},
\end{align}
where we assume that $p(x)$ is simply given and is not optimized with its own parameters. When $z$ does not exist, it reduces to the cross-entropy loss from Eq.~\eqref{eq:cross-entropy}:
\begin{align}
    L_{\mathrm{ll}}([x, y], \theta) = -\log p(y|x; \theta).
\end{align}

In the rest of this chapter, we focus on the case where we have input-only observations. We often call such a setup \textit{unsupervised learning}.

\section{Variational inference and Gaussian mixture models}
\label{sec:variational-inference}

We will derive something magical in this section, although it will not look magical at all in hindsight by the end of this section. To do so, let us first re-state that it is challenging to derive the posterior distribution $p(z|x)$ over the latent variable given an observation, unless we explicitly put severe constraints on the forms of $p(x|z)$ and $p(z)$.\footnote{
    In short, $p(z)$ should be a so-called conjugate prior to the likelihood $p(x|z)$, so that the posterior $p(z|x)$ follows the same distributional family as $p(x|z)$. 
} Instead of computing $p(z|x)$ directly, we can perhaps find a proxy $q(z; \phi(x))$ to this exact posterior distribution, called an approximate posterior. This approximate posterior probability function is parametrized by $\phi(x)$, where we use $(x)$ to denote that these parameters are specific to $x$.  When it is not confusing, we would drop $(x)$ here and there for both brevity and clarity. 

We are now faced with a task to make the proxy $q(z; \phi(x))$ a good approximation to the true posterior $p(z|x)$. We will do so by minimizing the Kullback-Leibler (KL) divergence which is defined as 
\begin{align}
    \label{eq:kl-div}
    D_{\mathrm{KL}}(q \| p) &=
    -\int q(z; \phi(x)) \log \frac{p(z|x)}{q(z; \phi(x))} \mathrm{d}z \\
    &= 
    -\mathbb{E}_{z \sim q} \left[ \log p(z|x) \right]
    - \mathcal{H}(q) \geq 0
\end{align}
where $\mathcal{H}(q)$ is the entropy of $q$ defined as
\begin{align}
    \label{eq:entropy}
    \mathcal{H}(q) = -\int q(z) \log q(z) \mathrm{d}z.
\end{align}
It is important to notice the inequality above, that is, the KL divergence is by definition non-negative. 

Let us continue from the KL divergence:
\begin{align}
    D_{\mathrm{KL}}(q \| p) &=
    -\int q(z; \phi(x)) \log \frac{p(z|x)}{q(z; \phi(x))} \mathrm{d}z \\    
    &= -\int q(z; \phi(x)) \log \frac{p(x|z)p(z)}{p(x) q(z;\phi(x))} \mathrm{d}z \\
    &= \log p(x) - \int q(z; \phi(x)) \log p(x|z) \mathrm{d}z - \int q(z;\phi(x)) \log \frac{p(z)}{q(z;\phi(x))} \mathrm{d} z \\
    &= \log p(x) - \mathbb{E}_{z\sim q} \left[ \log p(x | z) \right] + D_{\mathrm{KL}}(q(z; \phi(x)) \| p(z)).
\end{align}

To find $q$ (or its parameters $\phi(x)$), we minimize the second and third terms above, since the first term $\log p(x)$ is not a function of $q$ . In other words, 
\begin{align}
\label{eq:approx-posterior-inf}
    \hat{\phi}(x) &= \arg\min_{\phi(x)} - \mathbb{E}_{z\sim q} \left[ \log p(x | z) \right] + D_{\mathrm{KL}}(q(z; \phi(x)) \| p(z)) \\
    &= \arg\max_{\phi(x)} \underbrace{\mathbb{E}_{z\sim q} \left[ \log p(x | z) \right] - D_{\mathrm{KL}}(q(z; \phi(x)) \| p(z))}_{=J(\phi(x))}.
\end{align}
If we design $q(z; \phi(x))$, it is often possible to compute the (stochastic) gradient of this objective function $J$ w.r.t. $\phi(x)$ and use stochastic gradient descent to update $\phi(x)$ iteratively to find $q$ that is a better proxy to the true distribution than at the beginning.

In an interesting twist, this objective function $J$ is a lower bound to $\log p(x)$, because the KL divergence is greater than or equal to $0$:
\begin{align}
    \label{eq:variational-lowerbound}
    \log p(x; \theta) \geq \underbrace{\mathbb{E}_{z\sim q} \left[ \log p(x | z; \theta) \right] - D_{\mathrm{KL}}(q(z; \phi(x)) \| p(z))}_{=J(\theta)}.
\end{align}

This means that we can indirectly maximize the log-probability assigned to an observation $x$ by the model by maximizing its lowerbound. Maximizing the lowerbound does not guarantee that the actual quantity increases, but it ensures that the actual quantity is higher than the achieved maximum lowerbound. The quality of doing so is determined by the gap between the lowerbound and the actual quantity, and this gap turned out to be exactly the KL divergence between the approximate posterior and the true posterior, $D_{\mathrm{KL}}(q \| p)$. In other words, if our approximation to the posterior is good, we get a tighter lowerbound and consequently can maximize the true target quantity better.

Since the same objective function $J$ is used for both minimizing the KL divergence in order to find a better approximate posterior \eqref{eq:approx-posterior-inf} and maximizing the lowerbound to the true quantity \eqref{eq:variational-lowerbound}, we can perform both optimization simultaneously~\citep{neal1998view}:
\begin{align}
\label{eq:variation-inference-objective}
    \max_{\phi(x_1), \ldots, \phi(x_N), \theta}
    \frac{1}{N}
    \sum_{n=1}^N
    \mathbb{E}_{z\sim q(z;\phi(x_n))} \left[ \log p(x_n | z; \theta) \right] - D_{\mathrm{KL}}(q(z; \phi(x_n)) \| p(z)),
\end{align}
where $x_1, \ldots, x_N$ are the training examples. This formulation furthermore allows us to use stochastic gradient descent from \S\ref{sec:sgd}. This procedure is often referred to as stochastic variational inference and learning. Inference refers to estimating $\phi(x_n)$, and learning refers to estimating $\theta$.

\subsection{Variational Gaussian mixture models}
\label{sec:mog}

Let us consider a practical use case of stochastic variational inference and learning above. We start by defining a mixture of Gaussians. A generative story behind a mixture of Gaussians (or equivalently a Gaussian mixture model) is that there exist a finite number of Gaussian distributions, which are referred to as ``components'', and a latent variable $z$ selects one of these components. Once the component is selected, an observation $x$ is drawn from the corresponding Gaussian distribution. 

To map this story onto the probability functions, we begin with a prior distribution over the components:
\begin{align}
    p(z ) = \frac{1}{M},
\end{align}
where $M$ is the number of Gaussian components. This prior states that each and every component is equally likely to be selected. This can be relaxed, but we will stick to this for now. $z$ can take any one of $\left\{1, \ldots, M \right\}$.

Once the component is selected, we draw an observation $x$ from
\begin{align}
    p(x|z) = \mathcal{N}(x | \mu_z, \Sigma_z), 
\end{align}
where $\mu_z$ and $\Sigma_z$ are the mean and covariance of the $z$-th Gaussian component. For simplicity, let us assume that $\Sigma_z = I$, that is, the covariance is an identity matrix. In such a case, we say that the component is spherical Gaussian. 

We introduce an approximate posterior for each training example $x_n$. This approximate posterior is
\begin{align}
    q(z=k; \phi_n) = \alpha_k^n,
\end{align}
where $\alpha_z^n \geq 0$ and $\sum_{k=1}^M \alpha_k^n = 1$ for all $n=1,\ldots, N$.

We can now write down the objective $J$:
\begin{align}
    J(\alpha^1, \ldots, \alpha^N, \mu_1, \ldots, \mu_M)
    =&
    \frac{1}{N}
    \sum_{n=1}^N
    \left(
    \sum_{m=1}^M 
    \alpha_m^n 
    \left( 
    -\frac{1}{2} 
    \| x^n - \mu_m \|^2
    - 
    \frac{d}{2} \log 2\pi
    \right)
    \right.
    \\
    &
    \left.
    \qquad\quad\quad+
    \sum_{m=1}^M
    \alpha_m^n  \log M
    -
    \sum_{m=1}^M 
    \alpha_m^n \log \alpha_m^n
    \right),
\end{align}
where $d = \mathrm{dim}(x^n)$. 

Let's compute the gradient of $J$ w.r.t. $\mu_k$:
\begin{align}
    &\nabla_{\mu_k} = \frac{1}{N} \sum_n
    \left(
    \alpha_k^n (x^n - \mu_k) 
    \right)
    =
    \frac{1}{N}
    \left( 
    \sum_n\alpha_k^n x^n 
    -\mu_k \sum_{n} \alpha_k^n
    \right) = 0 \\
    \label{eq:gmm-mean}
    \iff&
    \mu_k = \sum_{n=1}^N \frac{\alpha_k^n}{\sum_{n'=1}^N \alpha_k^{n'}} x^n.
\end{align}
We can compute the exact solution to $\mu_k$ analytically, that maximizes $J$.

Let's do the same for $\alpha_k^n$:
\begin{align}
    &\nabla_{\alpha_k^n} 
    =
    -\frac{1}{2} \| x^n - \mu_k \|^2 
    -
    \frac{d}{2} \log 2\pi
    - \log M 
    - \log \alpha_k^n 
    - 1 = 0 \\
    \iff& 
    \log \alpha_k^n = -\frac{1}{2} \| x^n - \mu_k \|^2 -
    \frac{d}{2} \log 2\pi - \log M - 1 \\
    \iff&
    \alpha_k^n = \frac{\exp\left(-\frac{1}{2} \| x^n - \mu_k \|^2 -
    \frac{d}{2} \log 2\pi - \log M\right)}{\sum_{k'=1}^K \exp\left(-\frac{1}{2} \| x^n - \mu_{k'} \|^2 -
    \frac{d}{2} \log 2\pi  - \log M\right)}, \\
    \label{eq:gmm-posterior}
    \iff&
    \alpha_k^n = \frac{\exp\left(-\frac{1}{2} \| x^n - \mu_k \|^2\right)}{\sum_{k'=1}^K \exp\left(-\frac{1}{2} \| x^n - \mu_{k'} \|^2\right)},
\end{align}
because $\sum_{k=1}^K \alpha_k^n = 1$.

For the approximate posterior, we can solve it analytically and exactly. In fact, if we analyze the solution above more carefully, we realize that it is identical to the true posterior:
\begin{align}
    \log \underbrace{\alpha_k^n}_{p(z=k|x^n)} = \log \underbrace{\mathcal{N}(x^n | \mu_k, I)}_{=p(x^n|z=k; \theta)} 
    + \log \underbrace{\frac{1}{M}}_{=p(z=k)} - \log Z.
\end{align}
where $Z$ is the normalization constant. In other words, the KL divergence between $q(z; \phi(x))$ and $p(z|x)$ is zero. It also implies that there is no gap between the variational lowerbound and the true log-evidence $\log p(x)$.

Gaussian mixture models are special in that the variational lowerbound is tight, i.e., there is no gap. They are also special in that we can find the analytical solution to posterior inference and likelihood maximization in a relatively simple manner. Even in this case, one should notice that the solutions to setting these gradients to zero are co-dependent. We thus need to iteratively update these quantities multiple times until some kind of convergence is achieved. This process is called expectation-maximization (E-M), or more generally a coordinate-ascent algorithm. Each of these two steps (updating the posterior and updating the parameters) is guaranteed to improve the variational lowerbound, and this alternating procedure will ultimately find a local maximum. 

Although there is an analytical solution to the parameters at each E-M iteration, it may not be desirable to use this analytical solution, because it requires us to use the posterior means of all $N$ training examples. When $N$ is large, this step, which needs to be repeated, can be prohibitively expensive. We can instead use stochastic gradient descent by computing the posterior means of only a small subset of training set (which can be done exactly as we have derived earlier) and only slightly updating the parameters following the stochastic gradient computed using this minibatch alone. Each E-M iteration is not guaranteed to improve the overall variational lowerbound, but on average with small enough step sizes, stochastic gradient descent makes progress. This would be a good approach to implement the Gaussian mixture model on a very large dataset.

Once learning is over, we can use the fitted Gaussian mixture model to (a) draw more samples and (b) infer the posterior distribution over the components given a new observation. 

\subsection{K-means clustering}

Let us introduce a \textit{temperature} $\beta \geq 0$ as a hyperparameter in Eq.~\eqref{eq:gmm-posterior}:
\begin{align}
    \alpha_k^n = \frac{\exp\left(-\frac{1}{2\beta} \| x^n - \mu_k \|^2\right)}{\sum_{k'=1}^K \exp\left(-\frac{1}{2\beta} \| x^n - \mu_{k'} \|^2\right)}.
\end{align}

When the temperature is high, i.e. $\beta \to \infty$, the posterior distribution is closer to the uniform distribution. This is understandable if you think of statistical thermodynamics. When the temperature is high, there is no particular configuration that is more likely than others, since all molecules are bouncing around non-stop with high energy. On the other hand, when the temperature approaches $0$, the posterior converges toward one of the corners of the $(K-1)$-dimensional simplex, meaning that only one of the components is probable and all the others are not at all. This would be an extreme case that interests us here.

When $\beta \to 0$, we can rewrite the solution to posterior inference as
\begin{align}
    \alpha_k^n = 
    \begin{cases}
    1,&\text{if } \| x^n - \mu_k \|^2 = \min_{k'=1,\ldots,K}  \| x^n - \mu_{k'} \|^2 \\
    0,&\text{otherwise}.
    \end{cases}
\end{align}
In this case, we can be more economical by storing 
\begin{align}
\label{eq:hard-em}
    \hat{z}^n = \arg\max_{k=1, \ldots K} \alpha_k^n,
\end{align}
instead of $K$ values for each $n$-th training example. In other words, we need only $\lceil \log_2 K \rceil$ bits as opposed to $K \times B$ bits where $B$ is the number of bits for representing a real value in one's system. 

In this case, the update rule for the mean of each component in Eq.~\eqref{eq:gmm-mean} can be simplified as well:
\begin{align}
    \mu_k &= \sum_{n=1}^N \frac{\alpha_k^n}{\sum_{n'=1}^N \alpha_k^{n'}} x^n \\
    &= \sum_{n=1}^N \mathds{1}(\hat{z}^n = k) x^n.
\end{align}
That is, we collect all training examples that belong to the $k$-th component and compute the average of these training examples. This further saves a significant amount of compute, as we only go through on average $N/K$ training examples for each component to compute its mean vector. 

Because we are effectively making the hard choice of to which component each training example belongs to \eqref{eq:hard-em}, we often refer to this special case as \textit{hard} expectation-maximization (EM). Furthermore, because we are effectively grouping the training examples into $K$ clusters, and each cluster is represented by its mean, this algorithm is called {\it $K$-means clustering} as well. This is one of the most widely used algorithms in unsupervised learning and data analysis, and the variational inference based approach we started with allows us to more flexibly extend this algorithm to work with more non-trivial distributions.

\section{Continuous latent variable models}

Let's restate the objective function derived from the variational inference principle earlier in Eq.~\eqref{eq:variation-inference-objective}:
\begin{align}
    \max_{\phi(x_1), \ldots, \phi(x_N), \theta}
    \frac{1}{N}
    \sum_{n=1}^N
    \mathbb{E}_{z\sim q(z;\phi(x_n))} \left[ \log p(x_n | z; \theta) \right] - D_{\mathrm{KL}}(q(z; \phi(x_n)) \| p(z)).
\end{align}
Looking at this formula, there is absolutely no reason for us to assume that $z$ is a discrete variable, as we did with the mixture of Gaussians above. $z$ can very well be a continuous real-valued vector. 

Let us try a simple case here by assuming that
\begin{align}
\label{eq:probabilistic-pca}
    &p(z) = \mathcal{N}(z; 0, \sigma^2 I^{|z|}) \\
    &p(x|z; \theta) = \mathcal{N}(x; W z + b, I^{|x|}),
\end{align}
where $\theta = (W, b)$ with $W \in \mathbb{R}^{|x| \times |z|}$ and $b \in \mathbb{R}^{|z|}$. $\sigma^2 $ is a hyperparameter and controls the strength of regularization. We will discuss more what we mean by this. We further use a simple approximate posterior for each example $x^n$:
\begin{align}
    q(z; \phi(x_n)) = \mathcal{N}(z; \mu_n, I^{|z|}),
\end{align}
where $\phi(x_n) = (\mu_n)$.

Then, the objective for each training example $x^n$ becomes
\begin{align}
    J_n &= \mathbb{E}_{z \sim q_n} \left[ -\frac{1}{2} \| x_n - Wz - b \|^2 - \frac{|x|}{2} \log 2\pi\right]
    -
    \frac{1}{2} * \left[ \frac{K + ||\mu_n||^2}{\sigma^2} - K + 2K \ln(\sigma) \right] \\
    &= -\mathbb{E}_z \left[ 
    \frac{1}{2} \| x_n \|^2 + \frac{1}{2}\| Wz + b \|^2 - x_n^\top (W z + b) \right] - \frac{1}{2\sigma^2} \| \mu_n \|^2 \\
    &= 
    -\mathbb{E}_z \left[ \frac{1}{2} z^\top W^\top W z +\frac{1}{2} \|b\|^2
    +b^\top W z-x_n^\top Wz - x_n^\top b\right]
    -\frac{1}{2\sigma^2} \| \mu_n \|^2 + \mathrm{const.} \\
    &= 
    -\frac{1}{2} \mathrm{tr~} W \mathbb{E}_z[(z-\mu_n) (z-\mu_n)^\top] W^\top 
    -\frac{1}{2} \mathrm{tr~} W\mu_n \mathbb{E}_z[z]^\top W^\top 
    -\frac{1}{2} \mathrm{tr~} W \mathbb{E}_z[z] \mu_n^\top W^\top 
    +\frac{1}{2} \mathrm{tr~} W \mu_n \mu_n^\top W^\top  \\
    &\quad
    - \frac{1}{2} \|b\|^2 - b^\top W \mu_n + x_n^\top W \mu_n + x_n^\top b - \frac{1}{2\sigma^2} \| \mu_n\|^2 + \mathrm{const.} \\
    &=
    -\frac{1}{2} \mathrm{tr~}  W W^\top
    - \frac{1}{2} \mu_n^\top W^\top  W \mu_n 
    - \frac{1}{2} \| b\|^2
    - b^\top W \mu_n
    + x_n^\top W \mu_n
    + x_n^\top b
    - \frac{1}{2\sigma^2} \| \mu_n \|^2+ \mathrm{const.}
\end{align}
where $\mathrm{const.}$ refers to the terms that do not depend on either $\phi(x_n)$ nor $\theta$. 

Let's perform posterior inference first by computing the gradient of $J=\frac{1}{N} \sum_{n} J_n$ w.r.t. $\phi(x_n)$:
\begin{align}
&\nabla_{\mu_n} = -W^\top W \mu_n + W^\top (x_n - b) - \frac{1}{\sigma^2} \mu_n = 0 \\
&-(W^\top W + \sigma^{-2} I) \mu_n + W^\top (x_n - b) = 0 \\
&\mu_n = (W^\top W +  \sigma^{-2}  I)^{-1} W^\top (x_n - b).
\end{align}
Just like with the MoG above, we get a clean, analytical solution to each $\mu_n$. Because we need to compute the inverse of $W^\top W + I \in \mathbb{R}^{K \times K}$, this may be somewhat expensive, but we need to compute it once and use it for all $N$ $\mu_n$'s. 

Let's look at the role of $\sigma^2$ from the prior $p(z)$ earlier in this context. When $\sigma^2 \to \infty$, the expression above simplifies to 
\begin{align}
    \mu_n = \underbrace{(W^\top W)^{-1} W^\top}_{=\text{(a)}} (x_n - b),
\end{align}
where (a) is the \textit{pseudoinverse} of $W$. When $W$ is a square and invertible matrix, this corresponds to $W^{-1}$. In that case, we can think of $\mu_n$ as the solution to
\begin{align}
    \mu_n = W^{-1} (x_n - b),
\end{align}
which is equivalent to
\begin{align}
    x_n = W \mu_n + b.
\end{align}
This expression is the mean of the $p(x|z; \theta)$ from above. 

If there is no prior information available for $z$, i.e. $\sigma^2 \to \infty$, our best guess at which latent configuration led to $x_n$ (that is, posterior inference) is to multiply $x_n$ (after subtracting the bias $b$) with the inverse of the \textit{forward} matrix $W$. In other words, the prior knowledge we have (in this case, that the latent configurations are more probably if they are closer to the origin) would affect posterior inference. This is what we meant earlier by regularization and that $\sigma^2$ controls the strength of regularization.

We now compute the gradient of $J$ w.r.t. $W$ and $b$ \textit{given} $\mu_n$'s. Let's begin with $b$:
\begin{align}
&\nabla_b = \frac{1}{N} \sum_{n=1}^N \left( -b - W\mu_n + x_n \right) = 0 \\
\iff &
b = \frac{1}{N} \sum_{n=1}^N \left( W \mu_n - x_n\right).
\end{align}
This expression makes an intuitive sense. $b$, the bias, is the average offset between what we get given the latent configuration and what we actually observe. 

Let us continue with $W$:
\begin{align}
\nabla_W &= \frac{1}{N} \sum_{n=1}^N 
\left(
- W - W \mu_n \mu_n^\top - b \mu_n^\top + x_n \mu_n^\top 
\right)\\
&=-W\left(I + \frac{1}{N}\sum_{n=1}^N \mu_n \mu_n^\top\right)+ 
\frac{1}{N}\sum_{n=1}^N (x_n - b) \mu_n^\top.
\end{align}
Then,
\begin{align}
    W = 
    \left(\frac{1}{N} \sum_{n=1}^N (x_n - b)\mu_n^\top \right)
    \left(I + \frac{1}{N}\sum_{n=1}^N \mu_n \mu_n^\top\right)^{-1}.
\end{align}

The first term in the product in the right-hand side can be thought of implementing so-called a Hebbian learning rule: ``\textit{neurons that fire together, wire together}''~\citep{hebb1949organization}. If the $i$-th dimension of the observation $x_i$ fires (that is, beyond the bias $b_i$) and the $j$-th dimension of the latent variable $\mu_j$ fires together (where `fire' is defined as any deviate away from the bias or zero), the strength of the weight value $w_{ij}$ between them must be large.  This already shows up as the second term in the gradient w.r.t. $W$ above. 

The second term in the right-hand side (which corresponds to the first term in the gradient) is more complicated. This works as whitening $\mu_n$ inside the first term. That is, it makes $\mu_n$'s to be distributed so that the covariance is closer to the identity. This works as making $W$ capture the covariance between the mean-subtracted observations $(x_n-b)$'s and the whitened latent configurations $\left(\mu_n \left(I + \frac{1}{N}\sum_{n=1}^N \mu_n \mu_n^\top\right)^{-1}\right)$'s. By doing so, in the next iteration of this EM procedure, $\mu_n$'s will be distributed such that their collectively covariance will be closer to the identity, which is what we imposed by saying that the prior over the latent variable should be a spherical Gaussian distribution. In other words, this is also the effect of regularization due to the prior distribution.

By cycling through these steps of updating $\mu_n$'s, $W$ and $b$, the variational lowerbound will improve gradually until convergence. In the limit of $\sigma^2 \to \infty$ and with the constraints that $W$ is orthogonal, i.e., $W W^\top = I$ and that $b = \frac{1}{N} \sum_{n=1}^N x_n$, we recover \textit{principal component analysis}~\citep{hotelling1933analysis}. Our derivation here is a special case of a more general version called probabilistic principal component analysis~\citep{tipping1999probabilistic}.  In particular, we follow the variational inference approach~\citep{ilin2010practical}.

This variational lowerbound based approach again enables us to use stochastic gradient descent which is much more scalable than the exact EM procedure. At each iteration, we pick a minibatch of training examples, infer the (approximate) posterior means and use them to compute the gradient of the variational lowerbound w.r.t. $W$ and $b$ . Instead of computing the optimal values given this minibatch, we simply update them slightly following the stochastic gradient direction. 

\subsection{Variational autoencoders}
\label{sec:vae}

A natural question, based on what we have already seen in \S\ref{sec:backprop}, is whether we can use a nonlinear transformation for $p(x|z)$ instead of the linear one in Eq.~\eqref{eq:probabilistic-pca}. This is totally possible with
\begin{align}
p(x|z; \theta) = \mathcal{N}(x | F(z; \theta), I^{|x|}),
\end{align}
where $F$ is an arbitrary nonlinear function, parametrized by $\theta$, that maps $z$ to $x$ and is differentiable w.r.t. $\theta$. In other words, we are okay with any kind of parametrization as long as we can compute\footnote{
    I will use $\frac{\partial}{\partial x}$ notation to refer to the Jacobian matrix unless it is confusing.
}
\begin{align}
    \mathrm{Jac}_{\theta} F(z; \theta) = \frac{\partial F}{\partial \theta}(z; \theta).
\end{align}

This small change has a big consequence in terms of the modeling power of the continuous latent variable model. This is due to the peculiar (and amazing) properties of normal distributions. Let us revisit the linear case above \eqref{eq:probabilistic-pca}:
\begin{align}
    &p(z) = \mathcal{N}(z; 0, \sigma^2 I^{|z|}) \\
    &p(x|z; \theta) = \mathcal{N}(x; W z + b, I^{|x|}),
\end{align}

Then, the joint probability can be written down as
\begin{align}
    \log p(x,z; \theta) &= \log p(z) + \log p(x|z; \theta) = 
    -\frac{1}{2\sigma^2}\|z\|^2    -\frac{1}{2}\|x - Wz - b \|^2 + \mathrm{const.} \\
    &= 
    -\frac{1}{2} \left(
    z^\top (\sigma^{-2} I) z + (x-b)^\top I (x-b) + z^\top W^\top W z - (x-b)^\top W z - z W^\top (x-b) 
    \right) + \mathrm{const.} \\
    &=
    -\frac{1}{2} \left(
    (x-b)^\top I (x-b) +
    z^\top (W^\top W + \sigma^{-2}I) z 
    - (x-b)^\top W z
    - z^\top W^\top z
    \right) + \mathrm{const.}
\end{align}

Let $v = \left[ x, z\right]^\top$ and $\mu = [b, 0^{|z|}]^\top$. Then,
\begin{align}
    \log p(x, z; \theta) - \log Z(\theta) = 
    -\frac{1}{2} \left(
        (v - \mu)^\top 
        \underbrace{
        \left[
        \begin{array}{c c}
            I &  -W \\
            -W^\top & W^\top W + \sigma^{-2} I
        \end{array}
        \right]
        }_{=\Sigma^{-1}}
        (v-\mu)
    \right)
    + \mathrm{const.}
\end{align}
This shows that the joint distribution over $[x,z]$ is also Gaussian with the mean $\mu$. Although we just needed to show this for our further argument, let us also check the covariance matrix of the joint distribution.

There is a magical formula called the block matrix inversion lemma:
\begin{align}
\label{eq:block-matrix-inversion}
    \begin{bmatrix}
        A & B \\
        C & D
    \end{bmatrix}^{-1} = 
    \begin{bmatrix}
        A^{-1} + A^{-1}B(D-CA^{-1}B)^{-1}CA^{-1} & -A^{-1}B(D-CA^{-1}B)^{-1} \\
        -(D-CA^{-1}B)^{-1}CA^{-1} & (D-CA^{-1}B)^{-1}
    \end{bmatrix}.
\end{align}
We can use this to write down the covariance of the joint distribution $p(x,z; \theta)$:
\begin{align}
    \Sigma &= \begin{bmatrix}
        I + W(W^\top W + \sigma^{-2}I - W^\top W)^{-1} W^\top &
        W(W^\top W + \sigma^{-2} I - W^\top W)^{-1} \\
        (W^\top W + \sigma^{-2} I - W^\top W)^{-1} W^\top &
        (W^\top W + \sigma^{-2} I - W^\top W)^{-1} 
    \end{bmatrix}
    \\
    &= 
    \begin{bmatrix}
        I + \sigma^{2} W W^\top & \sigma^2 W \\
        \sigma^2 W^\top & \sigma^2 I 
    \end{bmatrix}.
\end{align}

Because the marginal distribution over any subset of dimensions of a normal random variable is also normal, we know that the marginal distribution over $x$, i.e., $p(x) = \int p(x|z)p(z) \mathrm{d}z$, is also normal. In other words, the linear relationship between $x$ and $z$ in the probabilistic principal component analysis (PCA) above can only represent a Gaussian distribution over $x$. This is a critical limitation.

Such a limitation does not hold anymore if we use a nonlinear function to model the relationship between $x$ and $z$. In that case, the joint distribution $p(x,z)$ would not be in general Gaussian, because the covariance structure would not be stationary but change dynamically depending on $x$ and $z$. We would rather get a mixture of Gaussians with an infinitely many components:
\begin{align}
\label{eq:marginal-probability-vae}
    p(x) = \int p(x|z) p(z) \mathrm{d}z = \int p(z) \mathcal{N}(x| F(z; \theta), I) \mathrm{d}z.
\end{align}

Let's consider the variational lowerbound with this nonlinear formulation for one particular example $x_n$:
\begin{align}
\label{eq:vae-obj}
    J_n &= \mathbb{E}_{z \sim q_n} 
    \left[ -\frac{1}{2} \| x_n - F(z;\theta) \|^2 - \frac{|x|}{2} \log 2\pi\right]
    -
    \frac{1}{2} \left[ \frac{K + \|\mu_n\|^2}{\sigma^2} - K + 2K \ln(\sigma) \right].
\end{align}

The gradient of $J_n$ w.r.t. $\mu_n$ is then
\begin{align}
    \nabla_{\mu_n} &= 
    -\int
    \frac{\exp\left(-\frac{1}{2} \| z - \mu_n \|^2 \right)}
    {(2\pi)^{|z|/2}}
    \frac{1}{2} (z - \mu_n) \| x_n - F(z; \theta) \|^2 - \frac{\mu_n}{\sigma^2} \\
    &=-\frac{1}{2} \mathbb{E}_z \left[ (z-\mu_n) \| x_n - F(z; \theta)\|^2 \right] - \frac{\mu_n}{\sigma^2} \\
    &=-\frac{1}{2} \left( 
    2x_n \mathbb{E}_z \left[ z F(z;\theta) \right] - 2x_n \mu_n \mathbb{E}_z \left[F(z;\theta)\right] + \mathbb{E}_z \left[z \| F(z; \theta) \|^2 \right] - \mu_n \mathbb{E}_z \left[\| F(z; \theta) \|^2 \right]
    \right) - \frac{\mu_n}{\sigma^2}.
\end{align}
It is clear that without knowing the form of $F$, it is not possible to come up with an analytical solution to $\mu_n$ in general. Even worse, it is unclear how to compute the gradient analytically either, due to the challenging expectations that must be computed. We can however use sampled-based Monte Carlo approximation, since we can choose the approximate posterior $q$ to be readily samplable:
\begin{align}
    \nabla_{\mu_n}
    \approx
    \tilde{\nabla}_{\mu_n}
    =
    -\frac{1}{2} (\tilde{z} - \mu_n) \| x_n - F(\tilde{z}; \theta)\|^2 - \frac{\mu_n}{\sigma^2}.
\end{align}

In this particular case of Gaussian posterior, we can draw a sample using a reparametrization trick:\footnote{
    A reparametrization trick refers to formulating the process of sampling from a particular distribution as nonlinearly and deterministically transforming noise drawn from another distribution:
    \begin{align}
        z = g(\epsilon; \phi),\quad \epsilon \sim p(\epsilon).
    \end{align}
    This allows us to compute the derivative of the sample $z$ w.r.t. the parameters of this deterministic function, i.e., $\frac{\partial g}{\partial \phi}(z)$. This is a handy trick, since sampling is often considered a non-differentiable operator. Despite its usefulness, it is not always possible to come up with such reparametrization.
}
\begin{align}
    \tilde{z} = \mu_n + \sigma \epsilon, 
\end{align}
where $\epsilon \sim \mathcal{N}(0, I_{|z|})$. Plugging this in, we get
\begin{align}
    \tilde{\nabla}_{\mu_n} = -\frac{\sigma}{2} \epsilon \| x_n - F(\mu_n + \sigma \epsilon; \theta) \|^2 - \frac{\mu_n}{\sigma^2}.
\end{align}
We can then use this stochastic gradient estimate to find the solution to $\mu_n$. 

Looking at the gradient above, we can however see what this gradient direction points at. Particularly, the first term considers directions that are similar to the mean $\mu_n$ but with some noise. We then weigh each such direction (or the difference between this direction and the current estimate of $\mu_n$) by their quality, where the quality is defined as how similar the decoded observation, $F(z; \theta)$, is to the actual observation $x_n$ (notice the negative sign that turns this distance into the quality.) In other words, we look for the change to $\mu_n$ that makes the decoded observation closer to the actual observation. This makes perfect sense, from the perspective of $p(y|z)$. The second term simply brings $\mu_n$ toward the origin at the rate inversely proportional to the prior variance which is inversely proportional to the regularization strength. 

This lack of an analytical solution to $\mu_n$ is problematic, because we must keep $\mu_n$ for all $N$ training examples across E-M iterations, even with stochastic gradient descent. At each iteration, we would select a small number $M$ of training examples, retrieve the associated observations $\left\{x_m\right\}_{m=1}^M$ \textit{as well as} the associated current estimate of the posterior means $\left\{ \mu_m \right\}_{m=1}^M$, update the posterior means slightly following the gradient direction, update the parameters slightly following the stochastic gradient direction, and finally store back the updated estimate of the posterior means. This does not put too much pressure on computation but it puts a huge pressure on the storage and I/O, as we need $O(b \times |z| \times N)$ bits to store the posterior means. 

\paragraph{Amortized inference.}

Instead of storing the approximate posterior means for all the training examples, we can compress them into a powerful deep neural network. Let $G: \mathcal{X} \to \mathbb{R}^{|z|}$ be an inference network, as we will ask $G$ to approximately infer the posterior over the latent variable given an input $x$. This $G$ is also parametrized by its own set of parameters $\theta_G$. This inference network effectively work as a compressed version of the table containing $\left\{ \mu_m \right\}_{m=1}^M$, since we can retrieve $\mu_m$ by 
\begin{align}
    \mu_m = G(x_m; \theta_G).
\end{align}
In fact, this inference network even allows us to retrieve an approximate posterior distribution given a novel input $x' \notin D$ thanks to its generalization capability.

Let us now plug this inference network $G$ into the per-instance objective function from Eq.~\eqref{eq:vae-obj}:
\begin{align}
    J_n &= \mathbb{E}_{z \sim q_n(z; G(x_n; \theta_G), \sigma^2)} 
    \left[ -\frac{1}{2} \| x_n - F(z;\theta) \|^2 - \frac{|x|}{2} \log 2\pi\right]
    \\
    &\qquad-
    \frac{1}{2} \left[ \frac{K + \|G(x_n; \theta_G)\|^2}{\sigma^2} - K + 2K \ln(\sigma) \right].
\end{align}
Because of the expectation is difficult to evaluate, we will consider a single-sample estimate of $J_n$:
\begin{align}
    \tilde{J}_n 
    = 
    -\frac{1}{2}
    \| x_n - F(G(x_n; \theta_G)+\sigma \epsilon; \theta) \|^2 
    - \frac{1}{2 \sigma^2} \| G(x_n; \theta_G)\|^2 + \mathrm{const.},
\end{align}
where $\epsilon \sim \mathcal{N}(0, I)$. 

There are two non-constant terms in this approximate objective. The first term is the reconstruction error. The input $x_n$ is processed by the inference network $G$ first, and then the noisy version of the output of $G$ is then processed by $F$ to reconstruct the input. The objective is maximized when the difference between the original input and the reconstructed input is minimized (see the negation in front of the L2 norm.) This process is often referred as \textit{autoencoding}, and this is why this whole framework is called a \textit{variational autoencoder}~\citep{kingma2013auto}.

The second term is a regularizer that pushes the L2 norm of the output from the inference network to be small. This ensures that all the inputs $\left\{ x_1, \ldots, x_M \right\}$ are mapped to the latent space, i.e. the space of the latent variable $z$, as tightly as possible. Without this term, the norm of the output of $G$ can grow indefinitely, pushing the inferred posteriors of all the inputs to be as far away as possible, since this would ensure that $F$ can reconstruct the original input perfectly even with the injected noise. This would however make it impossible for $F$ to cope with any $z$ sampled from the prior or located between any pair of inputs' inferred posterior distributions, resulting in a lousy generative model. 

Thanks to the reparametrization trick, we can compute the gradient of $\tilde{J}_n$ w.r.t. all parameters, including those of $F$ and those of $G$. In other words, we can use backpropagation to train both the inference and generation networks, $G$ and $F$, respectively. This allows us to train the inference network extremely efficiently, without having to maintain a whole database of instance-specific approximate posterior parameters. Furthermore, as discussed earlier, this inference network can be used with a novel input, making it useful for analyzing a set of inputs that are not present during training. The approximate posterior computed by the inference network can be further finetuned using gradient descent to match the true posterior better~\citep{hjelm2016iterative}.

Perhaps more importantly, this implies that such end-to-end learning of inference and generation networks is possible with backpropagation and stochastic gradient descent as long as we can use a reparametrization trick to sample from an approximate posterior without breaking the differentiability. This opens wide a door to a whole new set of opportunities to scale up various probabilistic models that were cumbersome to derive and use before, although these are out of the scope of this course.

\subsection{Importance sampling and its variance.}
\label{sec:importance-sampling}

Before ending this section, let us think of how we should compute the log-marginal probability of an observation $x$ from Eq.~\eqref{eq:marginal-probability-vae}:
\begin{align}
    p(x) = \mathbb{E}_{z \sim p(z)}\left[ p(x|z) \right].
\end{align}
Unlike in the training time, we are less under time pressure, and therefore a natural approach would be a naive Monte-Carlo approximation:
\begin{align}
    p(x ) \approx \frac{1}{M} \sum_{m=1}^M p(x | z_m), 
\end{align}
where $z_m \sim p_z(z)$. 

Unfortunately this naive approach can have a large variance.  For brevity, let $f(z) = p(x | z)$ and $p(z) = p_z(z)$. Because we already know that it is unbiased, we can then write the variance as
\begin{align}
    \mathbb{V}\left[\frac{1}{M} \sum_{m=1}^M f(z_m)\right] 
    = \frac{1}{M^2}\mathbb{V}\left[ \sum_{m=1}^M f(z_m) \right]
    = \frac{1}{M^2}\sum_{m=1}^M \mathbb{V}\left[  f(z_m) \right]
    = \frac{1}{M^2} M \mathbb{V}\left[f(z)\right]
    = \frac{\mathbb{V}\left[f(z)\right]}{M},
\end{align}
because $z_m$'s are identically distributed according to $p(x)$. 

It turned out that we can reduce this variance by avoiding sampling from $p_z$ directly but from another distribution $q_z$. This technique is called importance sampling:
\begin{align}
    \mathbb{E}_{z\sim p_z}\left[ f(z)\right] = \mathbb{E}_{z\sim q_z}\left[ \frac{p_z(z)}{q_z(z) } f(z)\right] \approx \frac{1}{M} \sum_{m=1}^M \frac{p_z(z_m)}{q_z(z_m)} f(z_m).
\end{align}
We can then control the variance of this estimator by choosing $q_z$ carefully. To understand how we can choose $q_z$ carefully, consider the variance of this estimator:
\begin{align}
    \mathbb{V}\left[ 
    \frac{1}{M} \sum_{m=1}^M \frac{p_z(z_m)}{q_z(z_m)} f(z_m)
    \right]
    = \frac{1}{M} \mathbb{V}\left[ \frac{p_z(z_m)}{q_z(z_m)} f(z_m) \right].
\end{align}
Let's plug $p(x|z)$ and $p_z(z)$ back in:
\begin{align}
    \frac{1}{M} \mathbb{V}\left[ \frac{p_z(z_m)}{q_z(z_m)} f(z_m) \right]
    =
    \frac{1}{M} \mathbb{V} \left[ 
    \frac{p(x|z) p(z)}
    {q(z)}
    \right].
\end{align}
By using $\mathbb{V}[X] = \mathbb{E}[X^2] - \mathbb{E}[X]^2$, we get 
\begin{align}
    \mathbb{V} \left[ 
    \frac{p(x|z) p(z)}
    {q(z)}
    \right] = 
    \int 
    \frac{p(x|z)^2 p_z(z|)^2}{q(z)} 
    \mathrm{d} z
    - 
    \mathrm{const.}
\end{align}
The second term is constant w.r.t. $q$, because that is nothing but the original quantity we are trying to approximate. 

Recall the following definition of Cauchy-Schwarz inequality:
\begin{align}
    \left|\left<u, v\right>\right|^2 \leq \left<u, u\right> \left<v,v\right>,
\end{align}
where $\left<\cdot, \cdot\right>$ is an inner product that generalizes a standard dot product. We can define an inner product on the square-integrable functions\footnote{
    A function $f$ is square-integrable when
    \begin{align}
        \int f(x) \mathrm{d}x < \infty.
    \end{align}
} as
\begin{align}
    \left<f,g\right> = \int f(x) g(x) \mathrm{d}x
\end{align}
over the domain of $x$. Then, we can write the Cauchy-Schwarz inequality as
\begin{align}
    \int f(x)g(x) \mathrm{d}x \leq \int f^2(x) \mathrm{d}x \int g^2(x) \mathrm{d}x.
\end{align}

Because $\int q(z) \mathrm{d}z = 1$ by definition, we observe that
\begin{align}
    \left(\int \left(\frac{p(x|z)p_z(z)}{\sqrt{q(z)}}\right)^2 \mathrm{d}z\right)
    \underbrace{\left(\int q(z) \mathrm{d}z\right)}_{=1}
    \geq 
    \left( \int \frac{p(x|z)p_z(z)}{\sqrt{q(z)}} \sqrt{q(z)} \mathrm{d}z\right)^2
    =
    \left(\int p(x | z) p_z(z) \mathrm{d} z\right)^2.
\end{align}
Considering both sides carefully, we see that they are equal when 
\begin{align}
    C q(z) = p(x|z)p_z(z).
\end{align}
This is easy to check by plugging it into the left hand side of the inequality above:
\begin{align}
    \left(\int \left(\frac{C q(z)}{\sqrt{q(z)}}\right)^2 \mathrm{d}z \right)
    \left(\int q(z) \mathrm{d}z\right)
    = C^2,
\end{align}
and then into the right hand side of the inequality:
\begin{align}
    \left( \int C q(z) \mathrm{d}z \right)^2 = C^2.
\end{align}

Because $\int q(z) \mathrm{d}z = 1$, 
\begin{align}
    C = \int p(x|z) p_z (z) \mathrm{d}z.
\end{align}
Putting them all together, we get the following optimal $q$:
\begin{align}
    q^*(z) = \frac{p(x|z)p_z(z)}{\int p(x|z')p_z(z')\mathrm{d}z'},
\end{align}
which turned out to be exactly the posterior distribution over $z$ given $x$. In other words, if we sample from the posterior distribution instead of the prior distribution and reweigh $p(x|z)$ according to their ratio $\frac{p(z)}{p(z|x)}$, our approximation is both unbiased and has the minimal variance.

This is however not the right way forward, since the posterior probability has in its own denominator the intractable integral. Rather, this says that the so-called \textit{proposal distribution} $q$ must be close to the true posterior distribution $p(z|x)$, which is in fact exactly the criterion we used to derive the variational lowerbound earlier in \S\ref{sec:variational-inference}.  When the variational lowerbound, which serves as the objective function for latent-variable models, is maximized, the KL divergence between $q$ (the approximate posterior) and $p$ (the true posterior) shrinks. In other words, we can simply use the trained $q$ inference network as the proposal distribution to approximate the log-marginal probability of an observation $x$ after training to obtain an unbiased, low-variance estimator of the quantity.\footnote{
    The variational lowerbound can be used as a proxy to the log-marginal probability as well. This is indeed a standard practice during training, to monitor the progress of learning. This quantity is however a biased estimate of the log-marginal probability, and it is important to use importance sampling to check the true log-marginal probability.
} It turned out maximizing variational inference had yet another advantage.

\chapter{Undirected Generative Models}

We have studied a few different approaches to generative modeling in the previous chapter. These approaches can be thought of as fitting a \textit{directed} graphical model where there are two variables, the observation $x$ and the latent $z$. These two variables are connected by a directed edge going from $z$ to $x$. In this model, we defined two relatively simple, or perhaps more correctly relatively easily-described, distributions, $p(z)$ and $p(x|z)$ but were able to model a complicated distribution over the observation by the process of marginalization, $p(x) = \int p(z) p(x|z) \mathrm{d}z$. Now, we must ask whether there are other ways to do the same. 

\section{Restricted Boltzmann machines: the Product of Experts}
\label{sec:rbm}

We begin with a pretty old idea called restricted Boltzmann machines~\citep[RBM;][]{smolensky1986harmony}. An RBM defines a bipartite graph with undirected edges between two groups; $x$ and $z$. Each partition consists of the dimensions of the observation $x$ or the latent $z$ . These partitions are fully connected with each other, but there is no edges within each partition. Each edge has a weight value, resulting in a matrix $W \in \mathbb{R}^{|x| \times |z|}$ . Each node also has its own scalar bias, resulting in two vectors $b \in \mathbb{R}^{|x|}$ and $c \in \mathbb{R}^{|z|}$. We then define an energy function as
\begin{align}
    \label{eq:rbm-energy}
    e(x, z, \theta=(W, b,c)) =& -x^\top W c - x^\top b - z^\top c 
    \\
    =& -\sum_{i=1}^{|x|} \sum_{j=1}^{|z|} w_{ij} x_i z_j 
    - \sum_{i=1}^{|x|} x_i b_i
    - \sum_{j=1}^{|z|} z_j c_j.
\end{align}
Although it is not necessary for $x$, we restrict $z$ to be a binary vector: $z \in \left\{0,1\right\}^{|z|}$. 

As we have done over and over so far, we can turn this energy function into the joint probability function:
\begin{align}
    \log p(x,z; \theta) = -e(x, z, \theta) 
    - \log \int_{x' \in \mathcal{X}}\sum_{z' \in \left\{0, 1\right\}^{|z|}} \exp \left( -e(x', z', \theta) \right) \mathrm{d}x',
\end{align}
where $\mathcal{X}$ is a set of all possible values $x$ can take. If $\mathcal{X}$ is a finite set, we replace $\int$ with $\sum$. 

Let us focus on the normalization constant, $\sum_{z' \in \left\{0, 1\right\}^{|z|}} \exp(-e(x,z',\theta))$. Since
\begin{align}
\exp(a + b) = \exp(a) \exp(b),
\end{align}
we can rewrite it into
\begin{align}
    \exp(-e(x, z, \theta)) =& \left(\prod_{i=1}^{|x|} \prod_{j=1}^{|z|} \exp(w_{ij}x_i z_j)\right)
    \left( \prod_{i=1}^{|x|} \exp(x_i b_i) \right)
    \left( \prod_{j=1}^{|z|} \exp(z_j c_j) \right) \\
    =& \prod_{i=1}^{|x|} \exp(x_i b_i) \prod_{j=1}^{|z|} \exp(w_{ij}x_i z_j + z_j c_j).
\end{align}

Now, I want to marginalize out $z$ from this expression. In most cases, this would be intractable, because there are $2^{|z|}$ possible values $z$ can take. This bipartite structure however turned out to be a blessing we can rely on. 

Let's consider the following simple case:
\begin{align}
    \sum_{z \in \left\{0, 1\right\}^2} \prod_{j=1}^2 f_j(z_j) 
    &=
    f_1(0)f_2(0) + f_1(0)f_2(1) + f_1(1)f_2(0) + f_1(1)f_2(1) \\
    &=
    f_1(0)(f_2(0)+f_2(1)) + f_1(1)(f_2(0)+f_2(1)) \\
    &=
    (f_1(0) + f_1(1))(f_2(0)+f_2(1)) \\
    &=
    \prod_{j=1}^2 (f_j(0) + f_j(1)).
\end{align}
Instead of summing exponentially many terms, we can multiply $|z|$ terms only:
\begin{align}
    \sum_{z \in \left\{0, 1\right\}^{|z|}} \exp(-e(x,z,\theta)) &= 
    \sum_{z \in \left\{0, 1\right\}^{|z|}}
    \prod_{i=1}^{|x|} \exp(x_i b_i) \prod_{j=1}^{|z|} \exp(w_{ij}x_i z_j + z_j c_j) \\
    &=
    \prod_{i=1}^{|x|} \exp(x_i b_i) 
    \sum_{z \in \left\{0, 1\right\}^{|z|}}
    \prod_{j=1}^{|z|} \exp(w_{ij}x_i z_j + z_j c_j) \\
    &=
    \exp\left(\sum_{i=1}^{|x|} x_i b_i\right) 
    \prod_{j=1}^{|z|}
    \left( 1 + \exp(w_{ij} x_i + c_j)\right) \\
\end{align}

You can think of the left-hand side of this derivation as the unnormalized probability function $\tilde{p}(x; \theta)$ of $x$, since the normalization constant of $p(x,z; \theta)$ is neither a function of $x$ nor $z$. In that case, we can write it down as
\begin{align}
    \tilde{p}(x; \theta) \propto \phi_0(x) \prod_{j=1}^{|z|} \phi_j(x),
\end{align}
where
\begin{align}
    &\log \phi_0(x) = x^\top b, \\
    &\log\phi_j(x) = \log(1 + \exp(w_{\cdot j}^\top x + c_j)).
\end{align}
We call each $\phi_k$ an expert, and this is a typical formulation of a product of experts~\citep[PoE;][]{hinton2002training}.

PoE's are unlike a mixture of experts (MoE), such as a mixture of Gaussians from \S\ref{sec:mog}.  MoE's have a significant advantage over PoE's in that they are readily normalized as long as each and every expert is well-normalized. PoE's however can model a much sharper distribution, unlike MoE's. The entropy of a MoE is always lowerbounded by the entropy of an individual component. This is not the case with a PoE, because the scores from the experts are multiplied rather than averaged. It is possible for any one expert to simply \textit{veto} by outputing a value close to 0, while this wouldn't affect the overall outcome in the case of an MoE. 

We use the log-likelihood objective, averaged over the whole training set, for training this RBM:
\begin{align}
L_{\mathrm{ll}}(x, \theta) = e(x, \theta) 
    + \log \int \exp \left( -e(x', \theta) \right) \mathrm{d}x'.
\end{align}
Just like earlier, we use stochastic gradient descent, and to do so, we need to be able to compute the gradient of this per-example loss w.r.t. the energy $e$. Once we can compute it, we can use the chain rule of derivatives to compute the gradient w.r.t. each parameter. So,
\begin{align}
    \label{eq:rbm-grad}
    \nabla_\theta L_{\mathrm{ll}} 
    =& \nabla_\theta e(x, \theta) 
    -
    \int 
    \underbrace{
    \frac{\exp(-e(x', \theta))}
    {\int \exp(-e(x'', \theta)) \mathrm{d}x''}
    }_{=p(x'; \theta)}
    \nabla_\theta e(x', \theta)
    \mathrm{d} x' \\
    =&
    \underbrace{\nabla_\theta e(x, \theta)}_{=\text{(a)}}
    -
    \underbrace{\mathbb{E}_{x';\theta} \nabla e(x', \theta)}_{=\text{(b)}}.
\end{align}

There are two terms in this gradient. The first term (a) is called a \textit{positive} phase, since it proactively decreases (recall that we are taking the negative gradient direction) the energy of the positive example, where the positive example refers to one of the training examples $x$ from the training set. The second term (b) is called a \textit{negative} phase, where it proactively increases the energy of a configuration $x'$ that is highly probable under the current model, i.e. $p(x';\theta)\uparrow$. This is exactly what we saw earlier when we learned about the cross-entropy loss for classification in \S\ref{sec:softmax}. 

Unlike the cross entropy with softmax earlier, we are in a worse situation here, because the number of possible values $x$ can take is much greater. In fact, it is exponentially larger, since we often use RBMs or any of these generative models to model a distribution over a high-dimensional space. In other words, we cannot compute the negative phase (b) exactly in a tractable time, or sometimes we just do not know how to compute it at all. 

In the remainder of this section, we study how we can efficiently draw these negative samples and use them for learning.

\subsection{Markov Chain Monte Carlo (MCMC) Sampling}
\label{sec:mcmc}

Let's imagine that we want to draw a set of samples from a complicated target distribution $p^*(x)$. It would be great if we could draw samples independently in parallel, but this is often impossible. Rather, we need to come up with a way to draw a series of samples such that collectively they form a set of independent samples from the target distribution. How would we do this?

We do so by defining a Markov chain $(\mathcal{X}, p^0, \mathcal{T})$, where $\mathcal{X}$ is the set of all possible observations (i.e. the state space), $p^0$ is the initial distribution over $\mathcal{X}$, and $\mathcal{T}$ is a transition operator.  The transition operator is really nothing but a conditional distribution over $\mathcal{X}$ given a sample from $\mathcal{X}$, i.e., $\mathcal{T}(x|x')$. We can draw a series of observations $(x_1, x_2, \ldots)$ by repeatedly sampling $x_t \sim \mathcal{T}(x | x_{t-1})$ with $x_0 \sim p_0(x)$.  Eventually, that is, the latter part of this series of repeated sampling, we want those samples to be drawn from the target distribution $p^*(x)$. In other words, we want a stationary distribution $p^\infty$, which is the normalized cumulative visit counts for all states and satisfies
\begin{align}
    p^{\infty} = \mathcal{T} p^{\infty},
\end{align}
to match $p^*$. Once we converge to the stationary distribution, which matches the target distribution, we can simply apply the transition operator repeatedly and be convinced that the collected series of samples form collectively a set of samples from the target distribution.

In addition to this condition ($p^\infty = p^*$), we need to meet an extra condition. That is, this stationary distribution has to be unique. If there are other stationary distributions, we may not be able to tell that even after running this transition operator indefinitely that we are collecting samples from the true distribution. To do so, we further put a constraint that this Markov chain is \textit{ergodic}. In an ergodic Markov chain, any state (or a region of the state space, in the case of an infinitely large $\mathcal{X}$) is reachable from any other state within a finite number of transition steps. This ergodicity guarantees that there is only one stationary distribution, and that repeated applications of the transition operator will eventually converge toward this unique stationary distribution.

Sampling from a complicated target distribution $p^*$ then boils down to designing a transition operator $\mathcal{T}$ such that the resulting Markov chain has a unique stationary distribution. The next question is how we can guarantee that there exists a stationary distribution, since the ergodicity tells us that there is a unique stationary distribution \textit{if} there is a stationary distribution under this Markov chain. There are more than one way to do so, and one relatively well-known way is the principle of detailed balance. Detailed balance in a Markov chain is defined as having the transition operator $\mathcal{T}$ satisfy
\begin{align}
    \mathcal{T}(x' |x) p^{\infty}(x)  = \mathcal{T}(x | x') p^{\infty}(x').
\end{align}

As pretty clear from the equation, it says that whatever flows from one state to another must flow back. This is stronger than having a stationary distribution, as a stationary distribution $p^{\infty}$ may not satisfy this. When detailed balance is satisfied, we often refer to such a Markov chain as a reversible Markov chain, since we will not be able to tell the direction of time once it converged. 

Our goal is then to design a transition operator $\mathcal{T}$ such that the resulting Markov chain is ergodic and satisfies detailed balance.\footnote{
    This statement does not exclude the possibility of designing a Markov chain that allows us to sample from a target distribution even when it does not satisfy detailed balance. Furthermore, this statement does not exclude the possibility of expanding the state space by augmenting $x$ with an extra variable. It has been shown that this can be beneficial with so-called Hamiltonian Monte Carlos methods~\citep{neal1993hybrid}.
} We refer to the procedure of sampling by collecting a series of visited states from such a Markov chain by \textit{Markov Chain Monte Carlo} (MCMC) methods.

One of the most popular and widely-used MCMC algorithm is Metropolis-Hastings (M-H) algorithm~\citep{Hastings1970}. The M-H algorithm assumes that we have access to the unnormalized probability $\tilde{p}^*(x)$ of the target distribution:
\begin{align}
    p^*(x) = \frac{\tilde{p}^*(x)}{\int \tilde{p}^*(x)\mathrm{d}x}.
\end{align}
This assumption makes the M-H algorithm particularly suitable for many energy-based models, such as restricted Boltzmann machines (RBM), since we can easily often the unnormalized probability but cannot tractably compute the normalization constant.

We first assume we are given (or can create) a proposal distribution $q(x|x')$ that is often centered at $x'$ and whose probability mass is largely concentrated in the neighbourhood of $x'$. $q$ must be ergodic, that is, if we repeatedly sample from $q(x|x')$, we should be able to reach any state (or a region of the state space) within a finite number of steps. We then define an acceptance probability $\alpha(x | x')$ such that
\begin{align}
    \label{eq:mh-acceptance}
    \alpha(x|x') = \min\left( 1, 
    \frac
    {\tilde{p}^*(x) q(x'|x)}
    {\tilde{p}^*(x') q(x |x')}
    \right)
\end{align}
Then, the transition operator is
\begin{align}
    \mathcal{T}(x|x') = \alpha(x|x') q(x | x') + (1-\alpha(x|x')) \delta_{x'}(x),
\end{align}
where 
\begin{align}
    \delta_{x'}(x) = 
    \begin{cases}
        \infty,& \text{if } x = x' \\
        0,&\text{otherwise}
    \end{cases}
\end{align}
and 
\begin{align}
    \int \delta_{x'}(x) \mathrm{d}x = 1.
\end{align}
We can sample from this transition operator given the past sample $x'$ by
\begin{align}
    &\text{(1) } \tilde{x} \sim q(x|x')&\text{(Candidate generation)} \\
    &\text{(2) } \tilde{u} \sim \mathcal{U}[0, 1]&\text{(Random draw)} \\
    &\text{(3) } x = \begin{cases}
        \tilde{x}, &\text{if } \tilde{u} \leq \alpha(\tilde{x} |x') \\
        x',&\text{otherwise}
    \end{cases}
    &\text{(Acceptance)}
\end{align}

This transition operator satisfies both ergodicity and detailed balance, and a lot of MCMC algorithms can be viewed as variants of the M-H algorithm with particular choices of the proposal distribution $q$. 

\paragraph{Gibbs Sampling.}

Let's assume that $x$ is a finite-dimensional vector. We can then define a conditional probability over one particular dimension $d$ given all the other dimensions $\neq d$ as
\begin{align}
    p_d(x_d | x'_{1, \ldots, d-1, d+1, \ldots, |x|}) 
    =
    \frac{p([x'_1, \ldots, x'_{d-1}, x, x'_{d+1}, \ldots, x_{|x|}])}
    {\int p([x'_1, \ldots, x'_{d-1}, \tilde{x}, x'_{d+1}, \ldots, x_{|x|}]) \mathrm{d} \tilde{x}}.
\end{align}
Assume $d$ follows a uniform distribution, i.e. $d \sim \mathcal{U}\left\{1, 2, \ldots, |x|\right\}$ and we start from $x'=[x'_1, \ldots, x'_{|x|}]$. We now replace the $d$-th dimension of $x$ by sampling from the conditional distribution $p_d$, resulting in $x = [x'_1, \ldots, x'_{d-1}, \tilde{x}_d, x'_{d+1}, \ldots, x'_{|x|}]$. In order to compute the acceptance probability, we must compute
\begin{align}
\frac
    {\tilde{p}^*(x) p_d(x'|x)}
    {\tilde{p}^*(x') p_d(x |x')}
    &=
    \frac
    {\tilde{p}^*([x'_1, \ldots, x'_{d-1}, x_d, x'_{d+1}, \ldots, x'_{|x|}]) p_d(x'_d | [x'_1, \ldots, x'_{d-1}, x'_{d+1}, \ldots , x'_{|x|}])}
    {\tilde{p}^*([x'_1, \ldots, x'_{|x|}]) p_d(x_d | [x'_1, \ldots, x'_{d-1}, x'_{d+1}, \ldots , x'_{|x|}])} 
    \\
    &=
    \frac
    {{\color{red} \cancel{\tilde{p}^*([x'_1, \ldots, x'_{d-1}, x_d, x'_{d+1}, \ldots, x'_{|x|}])}} 
    {\color{blue} \cancel{\tilde{p}^*([x'_1, \ldots, x'_{d-1}, x'_d , x'_{d+1}, \ldots , x'_{|x|}])}}
    {\color{green} \cancel{C([x'_1, \ldots, x'_{d-1}, x'_{d+1}, \ldots , x'_{|x|}])}}}
    {{\color{blue} \cancel{\tilde{p}^*([x'_1, \ldots, x'_{|x|}])}}
    {\color{red} \cancel{\tilde{p}^*([x'_1, \ldots, x'_{d-1}, x_d, x'_{d+1}, \ldots , x'_{|x|}])}}
    {\color{green} \cancel{C([x'_1, \ldots, x'_{d-1}, x'_{d+1}, \ldots , x'_{|x|}])}}} 
    \\
    &=1
\end{align}
In other words, the acceptance probability is $1$, and we always accept this new sample which differs from the previous sample in just one dimension $d$. 

This procedure is called Gibbs sampling. We pick one coordinate, sample from the conditional distribution of that particular coordinate, replace it with the newly sampled coordinate value and repeat it. This procedure is often applicable even when we have access only to the unnormalized probability, since the conditional probability is often tractable in that case. Furthermore, because every sample is automatically accepted, there is almost no extra overhead in implementation, which makes it an attractive algorithm choice. 

\paragraph{Variational inference would not work.}

Based on what we have learned in \S\ref{sec:variational-inference}, one may wonder whether we could use variational inference instead of MCMC sampling. The answer is unfortunately no. The core idea of variational inference is to approximate a complex target distribution (the posterior distribution in \S\ref{sec:variational-inference}, and here the target distribution $p^*$) with a simpler distribution $q$ by minimizing
\begin{align}
    \mathrm{KL}(q \| p^*) = -\mathbb{E}_{x \sim q} \left[ \log \tilde{p}^*(x) \right] 
    + \underbrace{\log \int \tilde{p}^*(x) \mathrm{d}x}_{\text{const. w.r.t. }q}
    + \mathcal{H}(q).
\end{align}
If we focus on the first term of the KL divergence, we observe that we only care about the region of the observation space where $q$ is high. That is, the KL divergence only cares about the highly probable regions under $q$ and ignores any other regions that are highly probable under $p^*$ but not under $q$. In other words, samples we draw from $q$ after minimizing the KL divergence above would not be representative of $p^*$, because they will largely miss high probable regions under $p^*$.

This issue disappears as the complexity of $q$ increases and approaches that of $p^*$. This however comes with the very issue we want to solve; that is, we must sample from this equally complex $q$ in order to approximately compute and minimize the KL divergence above. Later in this chapter, we consider directly building a sampler so that an implicitly defined $q$ is both complex enough and approximately minimizes the KL divergence above.

\subsection{(Persistent) Contrastive Divergence}
\label{sec:pcd}

We need to sample from $p(x;\theta)$, to train an RBM. One way to produce a set of samples from $p(x;\theta)$ is to draw a set of $(x,z)$ samples from $p(x,z;\theta)$ and discard $z$ from each pair. In doing so, we want to use Gibbs sampling. Let us first try to write down the conditional probability of $z$ given $x$:
\begin{align}
    \log p(z | x; \theta) = 
    \sum_{j=1}^{|z|} 
    z_j 
    \left(\sum_{i=1}^{|x|}  w_{ij} x_i
    + c_j  \right)
    + \mathrm{const.}
\end{align}
This implies that the $z_1, \ldots, z_{|z|}$ are conditional independent given $x$, as 
\begin{align}
    p(z|x; \theta) = \prod_{j=1}^{|z|} 
    \frac{\exp\left( z_j \left(w_{\cdot j}^\top x + c_j\right)\right)}
    {
    \exp\left( 0\right) + 
    \exp\left( w_{\cdot j}^\top x + c_j \right)
    }.
\end{align}
We thus can look at each dimension of $z$ separately:
\begin{align}
    p(z_j=1 | x; \theta) = \frac{\exp(w_{\cdot j}^\top x + c_j)}{1+ \exp(w_{\cdot j}^\top x + c_j)} = \frac{1}{1+\exp(-w_{\cdot j}^\top x - c_j)} = \sigma(w_{\cdot j}^\top x + c_j),
\end{align}
where $\sigma$ is a sigmoid function we saw earlier: 
\begin{align}
    \sigma(a) = \frac{1}{1 + \exp(-a)}.
\end{align}

Sampling all $|z|$ dimensions is embarrassingly parallelizable, since they are conditionally independent. Let's say we have sampled a new $z$. We now need to sample a new $x$ given $z$. Following a similar derivation, we end up with
\begin{align}
    p(x|z; \theta) = \prod_{i=1}^{|x|} p(x_i | z; \theta),
\end{align}
where
\begin{align}
    p(x_i = 1 | z; \theta) = \sigma(w_{i\cdot}^\top z + b_i).
\end{align}
In other words, we can also sample all dimensions of $x$ in parallel as well. 

We can then alternate between sampling $x$ and $z$ repeatedly to collect a series of $(x,z)$ pairs that collectively constitute a set of samples drawn from $p(x,z; \theta)$. Of course, we want to probably throw away quite a few pairs from the early stage of sampling, as they have likely been collected before the Markov chain converged. Furthermore, in order to avoid the potentially slowly mixing rate of the Markov chain, we might want to use only every $k$-th sample. This strategy is often referred to as \textit{thinning}.

Of course, this does not really help us too much, since we must run a pretty long chain of Gibbs sampling in order to collect enough independent samples. If we run it too short, our stochastic gradient estimate will likely be incorrect, resulting in a disastrous outcome. 

Instead, it turned out that we can simply start the Gibbs sampling chain from a positive example, run it only a small number of steps (as few as just one) and use the resulting sample as the negative example. That is,
\begin{align}
    \nabla_{\theta} L^k(\theta; x) = 
        \underbrace{\nabla_\theta e(x, \theta)}_{=\text{positive}}
    -
    \underbrace{\frac{1}{S} \sum_{s=1}^S \nabla e(x'_s, \theta)}_{=\text{negative}},
\end{align}
where $x'_s$ is one of the $S$ samples drawn after running $k$ steps of Gibbs sampling starting from $x$. It is usual to set $S$ to $1$. In the limit of $k\to \infty$, this is exact, since the negative sample $x'$ would be from the stationary distribution which coincides with the true distribution $p(x; \theta)$. It is however not so with a finite $k$, and there is not even a guarantee that a larger $k$ leads to a better approximation, when $k$ is small. This strategy nevertheless results in a reasonably well trained RBM and is often called {\it contrastive divergence}.

It turned out that we can maintain the computational complexity with a minimal overhead in memory complexity by maintaining $S$ samples across multiple stochastic gradient steps while ensuring that learning converges to the exact solution asymptotically. We do so by running $S$ chains of Gibbs sampling in parallel to stochastic gradient descent. Between consecutive steps of SGD, we run $S$ chains of Gibbs sampling for $T \approx 1$ steps each to update the a set of $S$ samples that are more likely to have been drawn from the latest model. Then, we use these newly updated samples to compute the stochastic gradient estimate, to update the model parameters. 

As learning continues, the change to the model parameters slows down (since we are getting increasingly closer to a local minimum), and thereby Gibbs sampling chains in the background are increasingly getting closer to the stationary distribution of the final model. This makes it such that the early stage of learning is inexact but has a low variance (because we are not perturbing negative examples too much) but the later stage is exact since the model parameters change very slowly. This strategy is called \textit{persistent contrastive divergence}.

\section{Energy-based generative adversarial networks}

It is challenging to draw samples from a complex, high-dimensional distribution even with an advanced MCMC algorithm. Instead, we may want to consider training a neural network to draw samples from such a distribution. Any such neural network can be described as 
\begin{align}
    x = g(\epsilon; \theta_g),
\end{align}
where $\epsilon \sim p(\epsilon)$ and $p(\epsilon)$ is some easy-to-sample distribution of our choice. This sampler is parametrized by $\theta_g$. 

We can train this sampler by minimizing the following loss function:
\begin{align}
    L_{\mathrm{rkl}}(\theta_g) = \mathrm{KL}(p_g \| p_e) 
    &= -\mathbb{E}_{x \sim p_g}
    \left[
    \log p_e(x) - \log p_g(x)
    \right] \\
    &= -\underbrace{\mathbb{E}_{x\sim p_g}\left[\log p_e(x) \right]}_{=\text{(a)}}
    - \underbrace{\mathcal{H}(p_g)}_{=\text{(b)}},
\end{align}
where $p_g$ is the distribution underlying the sampler $g$ and $p_e$ is the distribution defined from the energy function $e$ using the Boltzmann formulation. We will consider two terms in this loss function separately.

The first term (a) is the negative expected energy of $x$ plus some constant:
\begin{align}
\label{eq:sampler-loss}
    \text{(a)} = \mathbb{E}_{x\sim p_g}\left[ \log p_e(x) \right] 
    &= \mathbb{E}\left[ -e(x) - \log \int \exp(-e(x')) \mathrm{d}x' \right]
    \\
    &= \mathbb{E}\left[ -e(x) \right] + \mathrm{const.}
\end{align}
Although we do not have $p_g$, we can draw samples from this distribution with $g$. We can thus compute the stochastic gradient of (a):
\begin{align}
    \nabla^{a}_{\theta_g} \approx 
    -\frac{1}{M} \sum_{m=1}^M \nabla_{\theta_g} e(g(\epsilon_m)), 
\end{align}
where $\epsilon_m \sim p(\epsilon)$. As long as $g$ is differentiable w.r.t $\theta_g$ and $e$ is differentiable w.r.t. the input, we can compute this stochastic gradient using backpropagation. By following the opposite direction to this stochastic gradient, we can effectively minimize the first term (a).

Unfortunately, (b) is less trivial to compute, since we do not have access to $p_g$. Instead of maximizing the entropy (see the negative sign in front of (b),) we can try to make $p_g$ closer to another distribution that potentially has a higher entropy. Assuming that $\mathcal{X}$ is a multi-dimensional real space, i.e. $\mathbb{R}^d$, the normal distribution is the maximum entropy distribution given a mean and a covariance matrix. We can thus draw many samples from $p_g$ using $g$, estimate the mean $\mu_g$ and covariance $\Sigma_g$ from these samples and then use the normal distribution with $\mu_g$ and $\alpha \Sigma_g$ as the mean and covariance, respectively, as the target distribution with a higher entropy than $p_g$, with $\alpha > 1$. 

When we have two sets of samples drawn from two distributions, we can use a kernelized maximum mean discrepancy (MMD) to measure the similarity between these two distributions. Unfortunately, it is definitely out of the scope of the course to discuss MMD and its kernelized estimator~\citep{gretton2012kernel}. Instead, we will trust that the following measures the discrepancy between two distributions when we have only two sets of samples:
\begin{align}
    \mathrm{MMD}^2(D, D') &= 
    \underbrace{\frac{1}{|D|(|D|-1)} \sum_{x \in D} \sum_{x' \in D': x' \neq x} k(x, x')}_{=\text{(a)}}
    \\
    &\quad+
    \underbrace{\frac{1}{|D'|(|D'|-1)} \sum_{x \in D'} \sum_{x' \in D': x' \neq x} k(x, x')}_{=\text{(b)}}
    \\
    &\quad-
    \underbrace{\frac{2}{|D||D'|} \sum_{x \in D} \sum_{x' \in D'} k(x, x')}_{=\text{(c)}},
\end{align}
where $k(\cdot, \cdot)$ is a kernel function. We will not discuss what kernel functions are, but you can think of the kernel function as some kind of a distance metric, such that any kernel function $k(a,b)$ satisfies two properties. First, it is symmetric:
\begin{align}
    k(a,b) = k(b,a).
\end{align}
Second, it is semi-positive definite:
\begin{align}
    x^\top K x \geq 0,\text{ for all } x \in \mathbb{R}^{n},
\end{align}
where $K$ is an $n\times n$ matrix with each entry $K_{ij} = k(v_i, v_j)$ for any set $\left\{ v_i \right\}_{i=1}^n$. For real vectors, one conventional choice is a Gaussian kernel defined as
\begin{align}
    k(a, b) = \exp\left( -\frac{1}{\sigma^2} \| a - b \|^2 \right).
\end{align}
Because the kernelized MMD above is differentiable w.r.t. the samples, as long as the kernel function was selected to be differentiable, we can compute the gradient of the MMD w.r.t. the parameters of the sampler $g$ and use it in place of the gradient of (b) from Eq.~\eqref{eq:sampler-loss}. 

Although we will not go into any technical detail behind this kernelized MMD, it is instructive to inspect it at an intuitive level. Let us start from the back. The third term (c) is intuitively correct, as it computes the average pair-wise distance between all possible pairs of samples from two distributions. If the average pair-wise distance is larger, the discrepancy between two underlying distributions must be high as well. 

Let's assume $|D|=|D'|$ (that is, we have the same number of samples from each distribution.) Then, the minimum this pair-wise distance can attain is determined by the average pair-wise distance within each set, since all these samples would be placed on top of the samples from the other distribution. Furthermore, when this happens, the first two terms, (a) and (b), would coincide with each other. Considering that the first two terms and the final term have opposite signs, they would cancel out each other, resulting in $0$, as desirable. In other words, (c) determines the overall discrepancy between two distributions, while (a) and (b) are there to take into account that the minimum discrepancy between two distributions is largely bounded from below by the intra-distribution dispersion. 

By minimizing the following loss, we can train a sampling network $g$ that transforms a sample from a simple distribution $p(\epsilon)$ into a sample from the target distribution defined from the energy function $e$:
\begin{align}
\label{eq:sampler-loss}
    J_g(\theta_g; e) = -\frac{1}{M} \sum_{m=1}^N e(g(\epsilon_m)) 
    -
    \lambda \underbrace{\mathrm{MMD}^2\left( 
    \left\{ s_n \right\}_{n=1}^N,
    \left\{ g(\epsilon_m) \right\}_{m=1}^M
    \right)}_{=R(\theta_g)},
\end{align}
where 
\begin{align}
    s_n \sim \mathcal{N}\left( 
    \mu = \frac{1}{M} \sum_{m=1}^M g(\epsilon_m),
    \alpha\Sigma
    \right)
\end{align}
with 
\begin{align}
    \Sigma = \frac{1}{M} \sum_{m=1}^M \left(g(\epsilon_m) - \mu\right) \left(g(\epsilon_m) - \mu\right)^\top.
\end{align}
It is important to treat $s_n$'s as constants rather than the functions of $\epsilon_m$'s. $\lambda > 0$ controls the balance between these two terms.

Now, we can use this sampler $g$ instead of using a costly MCMC sampler to draw samples from an energy function. In other words, we can compute the gradient for the energy function from Eq.~\eqref{eq:rbm-grad} by drawing samples from this sampler $g$:
\begin{align}
    \tilde{\nabla}_{\theta} = 
    \nabla_\theta e(x, \theta)
    -
    \frac{1}{M} \sum_{m=1}^M \nabla_\theta e(x_m, \theta), 
\end{align}
where $x_m = g(\epsilon_m; \theta_g)$ with $\epsilon_m \sim p(\epsilon)$. 

If you look at the first term from Eq.~\eqref{eq:sampler-loss} (the objective function to be maximized for training $g$) and the second term above, it is easy to see that they are identical. We can then put these two together into a single objective function and then see that we can train both the energy function and the sampler jointly by solving a minimax problem:
\begin{align}
    \min_{\theta} \max_{\theta_g} 
    \mathbb{E}_{x \sim D} \left[ e(x, \theta) \right]
    -
    \mathbb{E}_{\epsilon \sim p(\epsilon)} \left[ e(g(\epsilon; \theta_g), \theta)\right]
    - \lambda R(\theta_g).
\end{align}
In words, we try to adjust $\theta$ to ensure training instances are assigned lower energy values, while the samples drawn from $p_g$ are assigned higher energy values. Meanwhile, we ensure that the sampler $g$ draws samples that are assigned lower energy values and that the implicit distribution $p_g$ 's entropy is maximized. 

Because we are not reliant on Gibbs sampling, we can be much more relaxed about how to design an energy function, unlike with the RBM above. A natural choice is a deterministic autoencoder which is similar to the variational autoencoder from \S\ref{sec:vae} however without any noise in the middle. With the deterministic autoencoder, the energy function is defined as 
\begin{align}
    e(x; \theta) = \| F(G(x; \theta_G); \theta_F) - x \|^2,
\end{align}
where $\theta = \theta_G \cup \theta_F$. The energy value is lower if $x$ can be reconstructed better.

One can view this as the energy function $e$ and the sampler $g$ are playing an adversarial game. The energy function's job is to ensure that the sampler's samples are less likely than the true inputs, while the sampler's job is to ensure that the generated samples are as likely as true inputs according to the energy function. This approach was pioneered by \citet{goodfellow2014generative}, and this particular way to describe this approach using the energy function was explored soon after by \citet{zhao2016energy}. Once training is over, one can either use the sampler as is, or can use the sampler as the initialization for sampling from the trained energy function.

\section{Autoregressive models}

We have so far considered a family of generative models, called \textit{latent variable models}. Regardless of whether the probabilistic dependencies were described using directed or undirected edges, we used unobserved variables, or latent variables, in order to capture complex distributions. For each latent variable configuration, we define a relatively simple distribution over the observation. We call a distribution simple when this distribution has a small number of parameters and if we can build a differentiable neural net that maps the latent variable configuration to these parameters of the distribution. By marginalizing out these latent variables, we end up with a model that is able to capture a complex distribution. Then, is there any alternative?

Such a simple distribution is often inadequate to capture all variations of a full observation $X$ which almost always consists of simpler (lower-dimensional) constituents, i.e., $X = \left\{ x_1, \ldots, x_d \right\}$. Such a simple distribution is however often enough to capture the conditional distribution over an individual constituent which is often significantly lower-dimensional.  For instance, if $x_i$ is a categorical variable with $C$ categories, we can easily use softmax with $C$ parameters to capture this distribution. $X$ however can take $C^d$ many possible values, and this will not be easy to capture with simple softmax based parametrization. It is then tempting to imagine modeling these $d$ constituents of $X$ separately and combine them to build a model of $X$. 

Recall the chain rule of probabilities:
\begin{align}
    p(X) &= p(x_{\Pi(1)}) p(x_{\Pi(2)}|x_{\Pi(1)})p(x_{\Pi(3)} | x_{\Pi(1)}, x_{\Pi(2)}) \cdots \\
    &=\prod_{i=1}^d \underbrace{p(x_{\Pi(i)} | x_{\Pi(1)}, \ldots, x_{\Pi(i-1)})}_{=\text{(a)}}, 
\end{align}
where $\Pi$ is an arbitrary permutation of $\left(1, 2, \ldots, d\right)$. This chain rule states that the probability of any configuration of $X$ can be computed as the product of the probabilities of the $d$ constituents, appropriately conditioned on a subset of constituents. Without loss of generality, we assume $\Pi(i)=i$.

Our goal is to build a neural network that models (a) above and thereby model the joint probability function $p(X)$. There are two things to consider. First, we do not want to have $d$ separate neural networks to capture $d$ conditional probability distributions. We instead want to have a single neural network that is able to model the relationship between any pair of the target dimension $x_i$ and the context dimensions $x_{<i}=(x_1, \ldots, x_{i-1})$. This allows the predictor to benefit from patterns shared across these pairs. For instance, if $x_i$ was the $i$-th pixel in an image, we know that the pixel value of $x_i$ must be somewhat similar to $x_{i-1}$, regardless of $i$, due to the locality of pixel values. This knowledge should be more readily captured if a single predictor is used for all $i$.

Second, the number of parameters should not grow w.r.t. $d$, i.e., $|\theta| = o(d)$. It is in fact desirable to have $|\theta|=O(1)$, by having absolutely no dependency on $d$. This enables us to build an unsupervised model that can work on a variable-sized observation, which is critically important when dealing with variable-length sequences, such as natural language text and video. 

Combining these two considerations, we can now write this approach in the form of
\begin{align}
    x_i \sim G(F((x_1, x_2, \ldots, x_{i-1}); \theta), \epsilon),
\end{align}
where $\epsilon$ is noise. This reminds us of autoregressive modeling in signal processing,\footnote{
    A typical autoregressive model of order $p$ is signal processing is defined as 
    \begin{align}
        x_i = \sum_{k=1}^p \theta_k x_{i-k} + \epsilon_i.
    \end{align}
} and thus we refer to such an approach as \textit{autoregressive modeling}, as this is akin to a nonlinear autoregressive model with an unbounded context ($p\to\infty$). 

Two building blocks from \S\ref{sec:building-blocks} are particularly suitable for implementing $F$; a recurrent block and an attention block (with a positioning encoding.) In the case of a recurrent block, we do not need any modification, but can simply feed in the entire sequence $(x_0, x_1, x_2, \ldots, x_d)$ and read out $(p(x_1), p(x_2|x_1), \ldots, p(x_d|x_{<d}))$. More specifically, if we use gated recurrent units,
\begin{align}
    &h_i = F_{\mathrm{GRU}}([x_i, h_{i-1}]; \theta_r) \\
    &p(x_{i+1} | x_{\leq i}) = \frac
    {\exp(u_{x_{i+1}}^\top h_i + c_{x_{i+1}})}
    {\sum_{x \in C} \exp(u_x^\top h_i + c_x)},
\end{align}
where $h_0$ is a part of the parameters, and $x_0$ is a placeholder vector. We can then train this recurrent network to minimize the average log-loss:
\begin{align}
    \min_{\theta_r, U, c} 
    -\frac{1}{N} \sum_{n=1}^N
    \sum_{i=1}^{d_n}
    \log p(x_{i}^n | x_{<i}^n; \theta_r, U, c),
\end{align}
where we are being explicit about the possibility of variable-size observations by writing $d_n$. 

The attention block however requires one small modification. This modification is necessary, since we must ensure that $h_i$ is computed only using $(x_0, x_1, \ldots, x_i)$. This can be implemented by masking out the attention weights from Eq.~\eqref{eq:attention-weight} as
\begin{align}
\label{eq:masked-attention-weight}
    \alpha_i^j = \frac{\exp(q_i^\top k_j - m_{ij})}{\sum_{j'=1}^N \exp(q_i^\top k_{j'} - m_{ij'})},
\end{align}
where 
\begin{align}
    m_{ij} = 
    \begin{cases}
        0, &\text{if } j < i \\
        \infty, &\text{if } j \geq i
    \end{cases}
\end{align}
This would ensure that the output $\hat{v}_i$ from the attention block is not computed using any input vectors $(x_i, x_{i+1}, \ldots, x_{d})$. Some refer to this as \textit{causal masking} by borrowing from the concept of a causal system in signal processing. 

We must be careful when we are dealing with continuous $x_i$'s. We will discuss why this is the case, and how we can deal with it properly in \S\ref{sec:regression}, if time permits.

A major advantage of this autoregressive modeling approach is that we can compute the log-probability of any observation exactly. We simply need to compute the conditional log-probabilities and sum them to get the log-probability of the observation. This is unlike any latent variable approaches we have considered above. In the case of a variational autoencoder, we have to solve an intractable marginalization problem, and in the case of RBM's, we must compute the intractable log-partition function, or the log-normaliztaion constant. Furthermore, we can readily draw independent samples tractably with this autoregressive model, which is a great advantage for RBM's which require costly and challenging MCMC sampling.

This autoregressive modeling paradigm has become \textit{de facto} standard in building conversational agents in recent years since the successful demonstrations by \citet{brown2020language} and \citet{ouyang2022training}. To learn more about the fundamentals behind language modeling and related ideas, see this somewhat outdated lecture note~\citep{cho2015natural}. We do not go into any further detail, as these topics are out of the scope of this course.

\chapter{Further Topics}

\section{Reinforcement Learning}
\label{sec:reinforcement-learning}

\paragraph{Single-step reinforcement learning.}

We are in a situation where we must train a classifier but we are not given input-output pairs, but rather a black box that takes as input one of the outputs and returns a scalar reward, i.e., $R: \left\{1, \ldots, C \right\} \to \mathbb{R}$. This is truly a blackbox, unlike learning from \S\ref{sec:hyperopt} where learning was a blackbox due to its intractability. We want to train a classifier so that we maximize the reward by the blackbox on expectation:
\begin{align}
    \max_{\theta} \mathbb{E}_{x} \mathbb{E}_{y|x; \theta} \left[ R(y) \right].
\end{align}
The first question we often need to ask is whether we can compute the stochastic gradient of this objective w.r.t. the parameters $\theta$. Let us try that ourselves here:
\begin{align}
    \nabla_{\theta} \int p(x) \sum_{y=1}^C p(y|x; \theta) R(y) \mathrm{d}x =
    \int p(x) \underbrace{\nabla_{\theta} \sum_{y=1}^C p(y|x; \theta) R(y)}_{
    =\nabla \mathbb{E}_{y|x;\theta} R(y)
    } \mathrm{d}x.
\end{align}
We continue with $\nabla \mathbb{E}_{y|x;\theta} R(y)$ :
\begin{align}
    \nabla_{\theta} \sum_{y=1}^C p(y|x; \theta) R(y) =& 
    \sum_{y=1}^C \nabla_{\theta} p(y|x; \theta) R(y) \\
    =& 
    \sum_{y=1}^C p(y|x; \theta) R(y) \nabla \log p(y|x; \theta) \\
    =& \mathbb{E}_{y|x; \theta}[R(y) \nabla \log p(y|x; \theta)],
\end{align}
where we used the so-called \textit{log-derivative trick}.\footnote{
    \begin{align}
        f' = f \cdot (\log f)',
    \end{align}
    because
    \begin{align}
        (\log f)' = \frac{f'}{f}
    \end{align}
}

In other words, the stochastic gradient of the expected reward given an input $x$ is the weighted sum of the stochastic gradient of the log-probability assigned to each possible output, where the weights are the associated rewards and the outputs are drawn according to the classifier's output distribution. This intuitively makes sense. We want to follow the gradient direction that would encourage the classifier to put a higher probability on an output that is associated with a higher reward, more so than the other directions. 
Because it is often expensive (or even impossible) to run this blackbox, it is a usual practice to use a single sample dranw from $y|x; \theta$ to approximate this stochastic gradient:
\begin{align}
    \nabla_{\theta} \mathbb{E}_{y|x;\theta}\left[R(y)\right] 
    \approx
    R(\tilde{y}) \nabla_{\theta} \log p(\tilde{y}|x; \theta)
    = \hat{g},
\end{align}
where $\tilde{y} \sim y | x; \theta$.

Before declaring the victory, let us compute the variance of this stochastic gradient estimator:
\begin{align}
    \mathbb{V}[\hat{g}] = \mathbb{E}[\hat{g}^2] - \mathbb{E}[\hat{g}]^2.
\end{align}

Although we know that this is an unbiased estimator because we have derived it fully until we used single-sample Monte Carlos approximation (which is unbiased on its own), let's first compute $\mathbb{E}[\hat{g}]$:
\begin{align}
    \mathbb{E}[\hat{g}] &= \mathbb{E}_{y|x;\theta} \left[ R(y) \nabla_{\theta} \log p(y|x; \theta) \right] \\
&= \sum_y p(y|x;\theta) R(y) \nabla_{\theta} \log p(y|x; \theta) \\
&= \sum_y R(y) \nabla_{\theta} p(y|x; \theta) \\
&= \nabla_{\theta} \sum_y R(y) p(y|x; \theta) \\
&= \nabla_{\theta} \mathbb{E}_{y|x;\theta}\left[R(y)\right]
\end{align}
Then, we need to compute the first term of the variance above:
\begin{align}
    \mathbb{E}\left[\hat{g}^2\right] = 
    \mathbb{E}_{y|x;\theta} \left[ 
    R^2(y) \| \nabla_{\theta} \log p(y|x;\theta) \|^2
    \right]
\end{align}
Putting them together, we get
\begin{align}
\label{eq:reinforce-variance}
    \mathbb{V}\left[\hat{g}\right] = 
    \mathbb{E}\left[ R^2(y) \| \nabla \log p(y|x; \theta) \|^2\right]
    - \| \nabla_{\theta} \mathbb{E}[R(y)] \|^2.
\end{align}

Looking at the first term of the variance, we notice that there are two things that affect the variance greatly. The first factor is the magnitude of the reward. If the reward has a high magnitude, it results in an increased variance of the stochastic gradient estimate. This suggests that it is critical for us to \textit{control} the magnitude of the reward, although this is impossible if we are working with the truly black box $R$. The second factor is the norm of the gradient of the log-probability of the selected action w.r.t. the parameters $\theta$. In other words, the variance will be greater if the predictive probability computed by the model is sensitive to the change in the parameters. This suggests a very explicit way to regularize learning by minimizing this quantity directly, in order to stabilize learning. This technique is often referred to as \textit{gradient penalty}.

At this point, we begin to wonder if there is another stochastic estimator that is equally unbiased but potentially has a lower variance. Consider the following estimator, which is often referred to as a \textit{policy gradient} estimator:
\begin{align}
\label{eq:reinforce-baseline}
        \nabla_{\theta} \mathbb{E}_{y|x;\theta}\left[R(y)\right] 
    \approx
    (R(\tilde{y}) - b(x)) \nabla_{\theta} \log p(\tilde{y}|x; \theta),
\end{align}
where $b$ may be a function of $x$ but is independent of $y$. If we consider the expected value of the left-hand side, we notice that 
\begin{align}
    \mathbb{E}\left[ R(\tilde{y})\nabla_\theta \log p(y|x;\theta) \right] 
    -
    b(x) \underbrace{\mathbb{E} \left[\nabla_\theta \log p(y|x;\theta) \right]}_{=\text{(a)}}.
\end{align}
Let us dig deeper into (a) above:\footnote{
    I will omit $|x; \theta$ for the brevity without loss of generality.
}
\begin{align}
    \sum_{y=1}^C p(y) \nabla \log p(y) 
    =
    \sum_{y=1}^C \cancel{p(y)} \frac{1} {\cancel{p(y)}} \nabla p(y)
    =
    \nabla \underbrace{\sum_{y=1}^C p(y)}_{=1}
    =
    0.
\end{align}
In other words, the estimator in Eq.~\eqref{eq:reinforce-baseline} is an unbiased estimator. 

Although this estimator is identical to the original one in terms of the bias, this extra subtraction of $b(x)$ from $R(\tilde{y})$ has an important consequence for the variance. Let us consider the first term of the variance using the new estimator in Eq.~\eqref{eq:reinforce-variance}. We are particularly interested to find $b(x)$ that minimizes this term:
\begin{align}
&\nabla_b \frac{1}{2}
    \mathbb{E}_{y|x; \theta}
    \left[ 
    (R(y) - b)^2 
     \| s(y) \|^2
    \right] 
= -\mathbb{E}\left[ 
(R(y) - b) \|s(y)\|^2
\right]
= 0
\\
\iff& 
b \mathbb{E} \| s(y) \|^2 - \mathbb{E} \left[ R(y) \|s(y) \|^2\right] = 0 
\\
\iff&
b^* = \frac{\mathbb{E} \left[ R(y) \|s(y) \|^2\right]}{\mathbb{E} \| s(y) \|^2}.
\end{align}
where $s(y) = \nabla \log p(y|x;\theta)$. Unfortunately, this optimal baseline is intractable or impossible to compute, since it requires us to query the blackbox $R$ for each and every possible outcome for the input $x$. 

It is rather more informative to consider the upperbound to the first term of the variance. Let $c_{\max} = \max_{y=1, \ldots, C} \| s(y) \|^2 \ll \infty$, which we can encourage by the technique of gradient penalty. Then,
\begin{align}
    \mathbb{E}\left[(R(y) - b(x))^2 \| s(y) \|^2\right] 
    \leq  c_{\max} \mathbb{E}\left[(R(y) - b(x))^2\right]. 
\end{align}
The optimal baseline to minimize the upperbound on the right hand side is
\begin{align}
    &\nabla_b \frac{c_{\max}}{2} \mathbb{E}\left(R(y) - b\right)^2
    =
    -c_{\max} \left(\mathbb{E} R(y) - b \right) = 0 \\
    \iff&
    b^* = \mathbb{E}_{y|x;\theta} \left[ R(y) \right].
\end{align}
In other words, the optimal baseline is the expected reward we anticipated given the input $x$. 

Of course, this quantity is again intractable or impossible to compute exactly. We can however now fit a predictor of $b^*$ given $x$ using all the past observations of $(x, R(\tilde{y}))$, because each $R(\tilde{y})$ is a single-sample approximation to $\mathbb{E}_{y|x;\theta}\left[R(y)\right]$.\footnote{
    In particular, we should use a mean-squared error as the loss function when fitting a predictor to estimate the expected reward. This comes from the fact that the optimal solution to minimizing the mean-squared error corresponds to computing the average, as easily seen below:
    \begin{align}
        &
        \nabla_{\mu} \frac{1}{2N}\sum_{n=1}^N \left(\mu - x_n\right)^2 = 
        \frac{1}{N}\sum_{n=1}^N (\mu - x_n)
        =
        \mu - \frac{1}{N}\sum_{n=1}^N x_n = 0 \\
        \iff &
        \mu = \frac{1}{N} \sum_{n=1}^N x_n.
    \end{align}
} 
Because we update $\theta$ along the way, many of the past samples would not be valid under the current $\theta$. If we however assume that $\theta$ is updated slowly and that the predictor is adapted rapidly, asymptotically this is an exact procedure, just like persistent contrastive divergence from \S\ref{sec:pcd}. 

We then need to maintain two predictors. One predictor is often called a \textit{policy network} that maps the current input, or state, $x$ to the distribution over possible outputs, or actions. The other predictor is often referred to as a \textit{value network} that maps the current state $x$ to the expected reward. The latter is called a value network, because it predicts the value of the current state, regardless of the action to be taken by the policy. These networks are trained in parallel.

\paragraph{The case of noisy reward: an actor critic method.}

Let's imagine that the reward $R$ depends on both $x$ and $y$ and that it is random as well. That is, we observe only a noisy estimate of the reward at $x$ given the output choice $y$. We probably want then to maximize the expected, expected reward:
\begin{align}
    \max_{\theta} \mathbb{E}_{y|x;\theta} \mathbb{E}_{\epsilon} \left[ R(y, x; \epsilon)\right],
\end{align}
where we use $\epsilon$ to collectively refer to as any kind of uncertainty in the reward $R$. Then, we have to further approximate the policy gradient with a sample reward $\tilde{R}(y, x)$:
\begin{align}
    \nabla_{\theta} \mathbb{E}_{y|x;\theta}\mathbb{E}_{\epsilon}\left[R(y, x; \epsilon)\right] 
    &\approx
    (\mathbb{E}_\epsilon [R(\tilde{y}, x; \epsilon)] - b(x)) \nabla_{\theta} \log p(\tilde{y}|x; \theta) \\
    &\approx
    (\tilde{R}(\tilde{y}, x) - \hat{b}(x; \theta_b)) \nabla_{\theta} \log p(\tilde{y}|x; \theta),
\end{align}
where $\hat{b}(x)$ refers to a predicted baseline, e.g. the value of $x$. $\theta_b$ is the parameters of this value function.

Unfortunately, this estimator will have an extra variance due to noisy reward. Similarly to what we did with the baseline above, we can lower the variance by predicting the expected reward at $(x,\tilde{y})$ using a predictor trained on samples. That is,
\begin{align}
\label{eq:actor-update}
    \nabla_{\theta} \mathbb{E}_{y|x;\theta}\mathbb{E}_{\epsilon}\left[R(y, x; \epsilon)\right] 
    &\approx
    (\underbrace{\hat{R}(\tilde{y}, x; \theta_r) - \hat{b}(x; \theta_b)}_{=\text{(a)}}) \nabla_{\theta} \log p(\tilde{y}|x; \theta),
\end{align}
where $\hat{R}$ is the reward predictor, parametrized by $\theta_r$. Such a reward predictor is often referred to as a Q value of the state-action $(x,y)$ pair.\footnote{
    Although almost no paper explicitly mentions what `Q' stands for, it is widely acknowledged that it stands for quality.
}  This difference (a) between the Q value $\hat{R}$ and the value $\hat{b}$ is called an \textit{advantage}, since this is tells us about the advantage of choosing $\tilde{y}$ over other outputs/\textit{actions}. 

An interesting observation here is that if we have $\hat{R}(y,x;\theta_r)$ and if $C$, the number of all possible $y$ values, is small, we can replace the value network $\hat{b} $  with
\begin{align}
    \hat{b}(x) = \mathbb{E}_{y|x;\theta} \left[ \hat{R}(y, x; \theta_r) \right],
\end{align}
which may help reduce the variance from having to train two separator predictors. With a reasonable $C$, this can implemented quite efficiently by having $\hat{R}$ to output a $C$-dimensional real-valued vector, multiplying the output with the output from the $y$ predictor (which is often called a \textit{policy}) and sum these values. Sometimes we call this $\hat{R}$ a \textit{critic} and $p(y|x;\theta)$ an \textit{actor}. This approach is thus called an \textit{actor-critic algorithm}.

\paragraph{Multi-step reinforcement learning}

Let us assume there exist $C$-many $|\mathcal{X}|\times | \mathcal{X}|$ stochastic transition matrix $\Sigma(y)$ such that
\begin{align}
    \Sigma_{ij}(y) \geq 0
    \quad\text{and}\quad
    \sum_{i=1}^C \Sigma_{i \cdot}(y) = 1,
\end{align}
for $y \in \left\{1, 2, \ldots, C\right\}$. This transition matrix gives us the distribution over the next state given the current state $x_{t-1}$ and the selected action $y_t$, as 
\begin{align}
    q(x=k | x_{t-1}, y_t) = \Sigma_{x_{t-1}, k}(y_t),
\end{align}
where we assume $\mathcal{X}$ is a finite set, although it is easy to extend it to a continuous state space $\mathcal{X}$. 

In defining this transition operator $q$, we have made an important assumption called \textit{Markov assumption}. That is, at time $t-1$, where we will end up at time $t$ given my choice of $y_t$ is independent of the past states $(x_1, \ldots, x_{t-2})$ I have visited so far nor the action choices $(y_1, \ldots, y_{t-1})$ I have made so far. We further assume that a reward function $S^\star$ that is defined on each state and returns a scalar, i.e. $S^\star: \mathcal{X} \to \mathbb{R}$. Each time we transit from $x_{t-1}$ to $x_t$ due to $y_t$, we receive a reward $S^\star(x_t)$. 

Together with a policy $\pi(y | x; \theta)$, it defines a distribution over trajectories, or often called \textit{episodes}. We can then sample a (potentially infinitely long) sequence of tuples of previous state $x_{t-1}$, selected action $y_t$, next state $x_t$ and received reward $s_t=S^\star(x_t)$. Of course, these tuples are highly correlated with each other, since they are collected from a single trajectory defined by a shared set of distributions, the policy, transition and reward. We will however for now ignore this by saying that we are considering a particular time step $t$ from many independent trajectories. 

Let us use $n$ to refer to each of these trajectories. In order to apply the policy gradient, or actor-critic algorithm, from above, we must start with the Q network $\hat{Q}(x_{t-1}^n, y_t^n)$. This Q network approximates the expected quality of $(x_{t-1}^n, y_t^n)$. We define the expected quality by first defining the quality of $(x_{t-1}^n, y_t^n)$ from the $n$-th trajectory as
\begin{align}
\label{eq:instant-q}
    \tilde{Q}(x_{t-1}^n, y_t^n) = s_t^n + \sum_{t'=t+1}^{T_n} \gamma^{t'-t} s_{t'}^n, 
\end{align}
where $\gamma \in [0, 1]$ is a so-called discounting factor and $T_n$ is the length of the $n$-th trajectory.

This formulation tells us that the quality of any particular state-action pair is determined by the accumulated rewards from there on throughout the full trajectory. Because we assumed the Markov property, it is perfectly fine for us to ignore how we arrived at $(x_{t-1}, y_t)$. With $\gamma < 0$, we are specifying that we do not want to take into account what happens too far into the future. This is often a good strategy to facilitate learning in the case of finite-length episodes, i.e. $T_n < \infty$, and is necessary to define the quality to be finite with infinitely-long episodes, i.e. $T_n \to \infty$.\footnote{
    Unless $\gamma < 1$, the quality easily diverges, assuming $s_t > 0$ even when $|s_t| < \infty$. 
} 

This particular quality from the $n$-th trajectory can be thought of a sample from a random variable $Q(x_{t-1}^n, y_t^n)$ which is defined as
\begin{align}
    Q(x_{t-1}^n, y_t^n) = s_t^n +& \mathbb{E}_{q(x_t| x_{t-1}, y_{t}^n)} \left[
    \right. \\
    &
    \gamma \mathbb{E}_{\pi(y_{t+1}|x_t)q(x_{t+1} | x_t, y_{t+1})}\left[
    s^\star(x_{t+1}) + 
    \right. 
    \\
    &
    \left.
    \left.
    \gamma \mathbb{E}_{\pi(y_{t+2}|x_{t+1})q(x_{t+2} | x_{t+1}, y_{t+2})}\left[
    s^\star(x_{t+2}) + 
    \cdots
    \right]
    \right]
    \right].
\end{align}
In other words, the expected quality is the weighted sum of all future per-step rewards after marginalizing out all possible future trajectories according to the transition model and the policy. 

When we are working with finite-length trajectories, we can easily train the Q network to minimize the following quantity:
\begin{align}
    \min_{\theta_r} \frac{1}{N} \sum_{n=1}^N
    \sum_{t=2}^{L_n}
    \frac{1}{2} 
    \left(
    \hat{R}(x_{t-1}^n, y_t^n) - \tilde{Q}(x_{t-1}^n, y_t^n)
    \right)^2,
\end{align}
because $\hat{Q}$ is an unbiased sample drawn from the true distribution of the quality defined immediately above. 

Unfortunately this is not possible, if we are working with an infinitely-long episode. Such an infinitely-long episode is not common in the current-day setups, but it is something we aspire to working with in the future, where we would anticipate a learning based system to be deployed in real world situations and adapt itself on the fly. Of course, in this case, we must updaate the Q network also on-the-fly. It is unfortunately not possible to get even a single sample $\tilde{Q}$, since we never see the end of any episode. 

Let us re-arrange terms in Eq.~\eqref{eq:instant-q}:
\begin{align}
    \tilde{Q}(x_{t-1}^n, y_t^n) 
    =& s_t^n + \sum_{t'=t+1}^{T_n} \gamma^{t'-t} s_{t'}^n \\
    =& s_t^n + \gamma \left(
    \underbrace{s_{t+1}^n + \sum_{t'=2}^{T_n} \gamma^{t'-t} s_{t'}^n}_
    {=\tilde{Q}(x_{t}^n, y_{t+1}^n)}
    \right).
\end{align}
We see that the quality is recursively defined:
\begin{align}
    \tilde{Q}(x_{t-1}^n, y_t^n) = s_t^n + \gamma \tilde{Q}(x_{t}^n, y_{t+1}^n).
\end{align}
This allows us to write a loss function to train the Q network without waiting for the full episode to end (or never end) by considering the temporal difference at time $t$:
\begin{align}
    &\min_{\theta_r} 
    \frac{1}{N} \sum_{n=1}^N
    \left( 
    \hat{R}(x_{t-1}^n, y_t^n; \theta_r) 
    - \gamma \left( s_t^n
    + 
    \hat{R}(x_t^n, y_{t+1}^n; \tilde{\theta}_r)\right)
    \right)^2 
    \\
    \iff&
    \min_{\theta_r} 
    \frac{\gamma^2}{N} \sum_{n=1}^N
    \left( 
    \left(
    \frac{1}{\gamma} \hat{R}(x_{t-1}^n, y_t^n; \theta_r) - \hat{R}(x_t^n, y_{t+1}^n; \tilde{\theta}_r)
    \right)
    - s_t^n
    \right)^2,
\end{align}
where $\tilde{\theta}_r$ is a previous estimate of $\theta_r$. We bootstrap from some random Q function (or its estimate) and iteratively improve our estimate of the Q function by learning to predict the \textit{temporal difference}. Unsurprisingly, we refer to this kind of learning as the method of temporal difference~\citep{sutton1988learning}. 

It turned out that such temporal difference methods are effective even when we are dealing with finite-length episodes, when these episodes are long. It is however generally challenging to train the Q network with such a temporal difference method due to many factors. For instance, the objective function above effectively tells us that the objective function itself is a function of our previous estimate $\tilde{\theta}_r$, meaning that a minimum one finds now will not continue to be a minimum once you plug the new estimate $\hat{\theta}_r$ into $\tilde{\theta}_r$. Furthermore, it will take a long time for the quality estimate the capture the longer-term dependencies of the choice of a particular action $y_t^n$ at $x_{t-1}^n$ on many steps later, since the naive temporal difference only considers one step deviation at a time. There have been many improvements proposed since the initial work, but it is out of the scope of this course to cover those. 

With this Q network (or the critic network, as we learned to call it above,) we can rely on the policy gradient to update the policy (or the actor network) from Eq.~\eqref{eq:actor-update}. There are of course many different ways to improve the actor update, for instance by constraining the update to be somewhat limited. Again, these are more or less out of the scope of this course.

\section{Ensemble Methods}
\label{sec:bayes}

\paragraph{Bagging.}

As we discussed already multiple times throughout the course (see e.g. \S\ref{sec:error-rate},) we are often in a situation where we do not have just one predictor but have access to many different predictors. These predictors can be thought of as samples drawn from \textit{some} distribution over all possible predictors:
\begin{align}
    \tilde{\theta}_n \sim q(\theta).
\end{align}
We will discuss where such a distribution comes from later, but for now, we will assume it magically exists and that we can readily draw $N$ classifiers from this distribution $q$. 

We already considered the case of having $q$ earlier in \S\ref{sec:bias-variance} when we considered the following bias-variance decomposition from Eq.~\eqref{eq:bias-variance-decomposition}:
\begin{align}
    \mathbb{E}_{x, y, \theta}
    (y - \hat{y}(x, \theta))^2 
    \propto
    \mathbb{E}_{x}
    \left[ 
    \underbrace{\mathbb{E}_{y|x} \left[ (y - \mu_y )^2 \right]}_{=\text{(a)}}
    +
    \underbrace{\mathbb{E}_{\theta} \left[ (\hat{y}(x, \theta) - \hat{\mu}_y)^2 \right]}_{=\text{(b)}}
    + \underbrace{(\mu_y - \hat{\mu}_y)^2}_{=\text{(c)}}
    \right],
\end{align}
where 
\begin{align}
    &\mu_y = \mathbb{E}_{y|x} \left[y\right],
    \\
    &\hat{\mu}_y = \mathbb{E}_\theta \left[\hat{y}(x, \theta)\right].
\end{align}

This decomposition was done on the loss averaged over the predictors drawn from the posterior distribution $q$. We can instead consider the loss computed using the average prediction from the predictors drawn from the posterior distribution. That is, our prediction is
\begin{align}
    \hat{y}(x) = \mathbb{E}_{\theta} \left[ \hat{y}(x, \theta) \right]. 
\end{align}
Then, 
\begin{align}
    \mathbb{E}_{x,y}\left( y - \hat{y}(x) \right)^2 
    &\propto 
    \mathbb{E}_x \left[ \underbrace{\mathbb{E}_{y|x} \left( y -\mu_y \right)^2}_{=\text{(a')}}
    -2 \underbrace{\hat{y}(x)}_{=\hat{\mu}_y}\underbrace{\mathbb{E}_{y|x} \left[y\right]}_{=\mu_y} + \underbrace{\hat{y}^2(x)}_{=\hat{\mu}_y^2} \right] \\
    &=
    \mathbb{E}_x \left[ 
    \underbrace{\mathbb{E}_{y|x} \left( y -\mu_y \right)^2}_{=\text{(a')}}
    \underbrace{- \mu_y^2}_{=\text{(b')}}
    +\underbrace{\left( \hat{\mu}_y - \mu_y \right)^2}_{=\text{(c')}} 
    \right].
\end{align}
Now, let us consider the difference between these two loss function. Since (a) and (a') are equivalent and (c) and (c') are equivalent, we just need to consider (b) and (b'):
\begin{align}
    \mathbb{E}_\theta \left(\hat{y}(x;\theta) - \hat{\mu}_y \right)^2 + \mu_y^2 
    =
    \mathbb{E}_\theta \hat{y}^2(x;\theta) - 2 \hat{\mu}_y 
    \underbrace{\mathbb{E}_\theta \left[ \hat{y}(x;\theta)\right]}_{=\hat{\mu}_y}
    + \hat{\mu}^2_y + \hat{\mu}^2_y
    =
    \mathbb{E}_{\theta} \hat{y}^2(x; \theta) 
    \geq 0.
\end{align}
In other words, this tells us that the average loss over the predictors is always greater than or equal to the loss of the average prediction by the predictors. This motivates the idea of \textit{bagging} \citep{breiman1996bagging}. 

As long as we have $q$, or a sampler that draws predictors, or the corresponding parameters, from this distribution $q$, bagging tells us that it is never a bad idea to use many of those sampled predictors and average their predictions, rather than using any one of them solely, on average. It turned out there are many different ways that make our predictor $\theta$ random rather than deterministic. We have already covered most of them earlier in the course, but let us briefly go through them here once more.

In modern machine learning, a major source of randomness is the use of stochastic gradient descent on a non-convex loss function. The loss function is not convex w.r.t. the parameters, as we stack highly nonlinear blocks to build a deep neural network based predictor, and in doing so, we introduce a large degree of redundancies (or ambiguities). These ambiguities are more or less arbitrarily resolved by randomness in stochastic gradient descent. For instance, our choice of the initial values of the parameters affect a subspace over which stochastic gradient descent can explore and find a local minimum. In addition to initialization, there are other types of randomness in stochastic gradient descent, that is, how we construct minibatches by selecting random subsets of the training set. Furthermore, quite a few building blocks are inherently stochastic. Recall the variational autoencoder from \S\ref{sec:vae}, where we injected noise for processing each and every instance during training. In other words, we can think of the resulting solution by running stochastic gradient descent as a sample drawn from some distribution implicitly defined by this process of learning. 

Of course, another major source of randomness is the choice of the training set. As we have discussed earlier in \S\ref{sec:error-rate}, we can imitate the randomness in data collection even when we have a single set of data points drawn from the underlying distribution by the process of bootstrap resampling. Instead of using the training set as it is, we can resample it to match its original size however by resampling training examples with replacement. Each time, we use a different resampled training set, we end up with a somewhat different solution which can be considered a sample drawn from the distribution again implicitly defined by the process of training set construction. 

In summary, we should embrace stochasticity inherent in learning and data collection in order to produce a set of distinct predictors and average their predictions for each input. On average, this will give us a low-loss predictor, thanks for the theory of bagging, above. 

\paragraph{Bayesian machine learning.}

Our discussion so far has progressed assuming that we are given this distribution $q(\theta)$. When $q(\theta)$ is given, bagging tells us that we want to use the average prediction from many sampled predictors from $q$ to build a lower-loss predictor on average. This however does not tell me anything about how we can create this distribution ourselves, or what this distribution $q$ is. 

It turned out that we can rely on probability to guide us in designing as well as understanding this distribution $q(\theta)$. This will resemble much what we have done in \S\ref{sec:probabilistic-ml}, and if you did not have much trouble following that section, you would not find it any confusing. Let us try to derive this $q$ distribution by first treating the loss value on the training set of a single predictor $\theta$ as the energy function:
\begin{align}
    e(\theta; D) = \sum_{x \in D} L(x; \theta),
\end{align}
with a very generic loss function $L$. 

We can interpret this energy function just like any other energy function we have defined and used throughout the semester. We want the predictor parametrized by $\theta$ to be assigned a low energy value when it is good. The goodness of the predictor is defined as how low the loss function this predictor attains on the training set $D$. 

We can now turn this energy function into the probability function using the Boltzmann formulation, as we have done over and over by now:
\begin{align}
    q(\theta| D, \beta) &= 
    \frac{\exp\left( -\beta \sum_{x \in D} L(x; \theta) \right)}
    {\int_{\Theta} \exp\left(-\beta \sum_{x \in D} L(x; \theta') \right) \mathrm{d}\theta'}
    \\
    &=
    \prod_{x\in D} 
    \frac{\exp\left(-\beta L(x; \theta)\right)}
    {\int_{\Theta} \exp\left(-\beta \sum_{x \in D} L(x; \theta') \right) \mathrm{d}\theta'} 
    \\
    &=\prod_{x \in D} 
    \frac{\exp\left(-\beta L(x; \theta)\right)}
    {\int_{\mathcal{X}} \exp\left(-\beta L(x'; \theta)\right) \mathrm{d}x'}
    \frac{\int_{\mathcal{X}} \exp\left(-\beta L(x'; \theta)\right) \mathrm{d}x'}
    {\int_{\Theta} \exp\left(-\beta \sum_{x \in D} L(x; \theta') \right) \mathrm{d}\theta'} 
    \\
    &=\prod_{x \in D} p(x | \theta, \beta) 
    \frac{\int_{\mathcal{X}} \exp\left(-\beta L(x'; \theta)\right) \mathrm{d}x'}
    {\int_{\Theta} \int_{\mathcal{X}} \exp\left(-\beta L(x'; \theta')\right) \mathrm{d}x' \mathrm{d}\theta'}
    \frac{\int_{\Theta} \int_{\mathcal{X}} \exp\left(-\beta L(x'; \theta')\right) \mathrm{d}x' \mathrm{d}\theta'}
    {\int_{\Theta} \exp\left(-\beta \sum_{x \in D} L(x; \theta') \right) \mathrm{d}\theta'} 
    \\
    &=\prod_{x \in D} p(x | \theta, \beta) \frac{p(\theta)}{\prod_{x' \in D} p(x' | \beta)}.
\end{align}
This is precisely the posterior distribution over $\theta$, where we consider $\theta$ to be a random variable. It states that our belief (probability) of a particular parameter configuration $\theta$ is proportional to the product of the likelihood $p(D|\theta, \beta) = \prod_{x\in D} p(x | \theta, \beta)$ and the prior belief of $\theta$. 

With this our updated (that is, posterior) belief over $\theta$, we probably want to marginalize $\theta$ out when we make a prediction on a new instance $x' \notin D$:
\begin{align}
    \label{eq:predictive-distribution}
    p(x'| D, \beta) = 
    \int_{\Theta} p(x'| \theta, \beta) q(\theta| D, \beta) \mathrm{d}\theta.
\end{align}
This formulation tells us that we should sample many predictors according to $q(\theta | D, \beta)$ and average their predictions, just like bagging above:
\begin{align}
    p(x' | D, \beta) \approx \frac{1}{M} \sum_{m=1}^M p(x' | \theta_m, \beta), 
\end{align}
where $\theta_m \sim q(\theta | D, \beta)$. In other words, if we follow the Bayes' rule and think of the loss function as an energy function of the parameter $\theta$ given an individual instance, we arrive at the conclusion that we should draw predictors from the posterior distribution $q(\theta | D, \beta)$. This is a great property, since we now have a good guideline on what we should do, although the inclusion of $\beta$ here was quite intentional, as it says that we still need some kind of hyperparameter search even in so-called \textit{Bayesian machine learning}. 

Let us now connect this (log-)posterior distribution with what we have learned so far by writing it as
\begin{align}
    \log p(\theta | D, \beta) 
    =&
    \sum_{x \in D} \log p(x | \theta, \beta) + \log p(\theta) - \log Z(D, \beta) 
    \\
    =& 
    -\beta \sum_{x \in D} L(x; \theta) + \log p(\theta) - \log Z'(D, \beta),
\end{align}
where we collect all terms that are constant w.r.t. $\beta$ into $\log Z'$. 

By setting $\beta = \frac{\alpha}{|D|}$, we end up with 
\begin{align}
    -\log p(\theta | D) = 
    \underbrace{\frac{\alpha}{|D|} \sum_{x \in D} L(x; \theta)
    - \log p(\theta)}_{=-\log p^*(\theta | D, \alpha)}
    + \mathrm{const.}
\end{align}
If we minimize this, that is, if we maximize the log-posterior, this is precisely what we have already been doing all along. We look for the parameter configuration $\theta$ that minimizes the average loss but use the regularizer to ensure that we end up with a parameter that generalizes. The balance between these two are determined by the constant $\alpha$. 

Since we can exactly compute the unnormalized posterior probability, we can think of using an advanced sampling technique, based on Markov Chain Monte Carlo methods, from \S\ref{sec:mcmc}~\citep{neal1996bayesian}. Unfortunately, this is often computationally too costly, because we must evaluate the loss over the entire training set $D$ each time we evaluate the acceptance probability. After all, the whole reason why we introduced stochastic gradient descent earlier was precisely because it was too costly to evaluate the loss over the whole training set. 

Fortunately, or obviously in retrospect, researchers have realized that stochastic gradient descent, with some adjustments or sometimes without much of adjustment, draws samples from this particular posterior distribution~\citep[see, e.g.,][]{welling2011bayesian}.  A general idea behind these recent algorithms, or findings, is that if we do not try to reduce the effect of noise, i.e. (b) in Eq.~\eqref{eq:decent-lamma}, stochastic gradient descent will tend toward a local minimum but will not tend to stay at the local minimum and jump out toward another local minimum. These local minima correspond to modes of the posterior distribution. By collecting all the parameter configurations visited by stochastic gradient descent, or some subset of them via thinning, we call parameter samples approximately according to the posterior distribution. 

This view of stochastic gradient descent as a posterior sampler tells us one more alternative to create a set of predictors for bagging. That is, we simply run stochastic gradient descent, without annealing the learning rate toward zero or while explicitly adding extra noise, and collect every once a while a predictor, to form a set of predictors for bagging. This approach explains why it has been successful to build a bag of deep neural networks to build an ensemble classifier~\citep{krizhevsky2012imagenet}, because those were approximate samples from the posterior distribution. 

\paragraph{Gradient Boosting.}

Consider a regression problem in which the target is $y \in \mathbb{R}^{d}$, and the energy function is defined as
\begin{align}
\label{eq:l2-energy}
    e([x;y], \theta) = \frac{1}{2} \| y - f(x; \theta) \|^2.
\end{align}

Let us imagine that we already have a trained predictor $f(x; \theta)$ which is not perfect. We want to fit another predictor $g(x;\theta')$ in order to ensure that we can make better prediction on $x$. We can approach this by first defining an aggregate predictor as
\begin{align}
    h(x; \{\theta, \theta'\}) = f(x; \theta) + \alpha g(x; \theta),
\end{align}
where $\alpha > 0$. We can then write the energy function that includes $h$ as
\begin{align}
\label{eq:one-step-boosting}
    e'([x;y], \{\theta, \theta'\}) &= \| y - h(x; \{\theta, \theta'\}) \|^2 
    \\
    &= \| \underbrace{(y - f(x;\theta))}_{\mathrm{(a)}} - \alpha g(x; \theta') \|^2.
\end{align}
We can minimize this energy function w.r.t. $\alpha$ and $\theta'$ , which results in $g$ that complements the existing predictor $f$ to minimize any remaining error by $f$. This idea is often referred as {\it boosting}~\citep{Schapire1990Strength}, as it boosts the representational power of weak predictors by combining two weak predictors, here $f$ and $g$, to form a stronger predictor. This procedure can be repeated by considering $h$ as $f$ and introducing yet another weak predictor $g$ into the mix, until the point at which a satisfyingly low level of the loss is achieved. Although we have derived it in the context of a single example $(x,y)$, it should be readily extended to multiple example pairs. 

By carefully inspecting (a), we realize that this term is the negative gradient of Eq.~\eqref{eq:l2-energy} w.r.t. $f(x; \theta)$:
\begin{align}
    \frac{\partial e}{\partial f(x;\theta)} = -(y - f(x; \theta)).
\end{align}
Instead of $e$, which was equivalent to the loss, because it was formulated using L2 distance, we can use a more generic loss $l(\theta; [x,y])$. We can then further rewrite $e'$ as
\begin{align}
    e'([x;y], \{\theta, \theta'\}) \propto 
    \| -\nabla_{\hat{y}} l(\theta; [x,y]) - \alpha g(x; \theta') \|^2,
\end{align}
where $\hat{y} = f(x; \theta)$ . By minimizing $e'$ w.r.t. $\theta'$ and $\alpha$, we effectively let $g$ capture the (scaled) negative gradient of the loss w.r.t. $\hat{y}$. 

As we have learned repeatedly over the course so far, the gradient is only meaningful in some small neighbourhood. In other words, taking the full step in the direction of the negative gradient may not necessarily decrease the overall loss, and we must scale the gradient accordingly. We thus search for the right step size by solving
\begin{align}
    \min_{\gamma \geq 0} l(\{\theta, \theta'\}; [x,y]),
\end{align}
where the loss $l$ is computed by comparing $y$ and 
\begin{align}
    \hat{y} = f(x; \theta) + \gamma g(x; \theta').
\end{align}
This procedure resembles the process of gradient descent from \S\ref{sec:sgd}, and is thereby referred to as {\it gradient boosting}~\citep{friedman2001gbm}. 

Boosting does not specify how to estimate $\beta$ and $g$ (or equivalently $\theta'$) at each iteration, and it is up to the practitioner to decide which (weak) learner $g$ they use and which loss function $l$ they choose. Popular choices include decision trees and kernel-based support vector machines. In this sense, this is not a learning algorithm but more a meta-heuristics.

\section{Meta-Learning}

In the previous section \S\ref{sec:bayes}, we learned that it is a good idea to average the predictions from multiple models if we have a distribution $q(\theta)$ over the models (or predictors) rather than a single predictor. We then learned that Bayesian machine learning tells us that this distribution should be conditioned on the training set, resulting $q(\theta | D)$, and that we can obtain this posterior distribution following the Bayes' rule:
\begin{align}
    q(\theta |D) \propto p(\theta) \prod_{x \in D} p(x | \theta).
\end{align}
It is a fair question at this point whether we must follow this particular formulation based on the Bayes' rule. Perhaps there is a better way to map the training set $D$ to the posterior distribution over $\theta$. 

Let us assume that we have not one but multiple training set $\left\{ D^1, D^2, \ldots, D^M \right\}$, corresponding to the $M$ prediction tasks. For each training set, we can define a so-called $K$-fold cross-validation loss as
\begin{align}
    L_{K\mathrm{CV}}(\phi; D^m) =  
    -\frac{1}{K}
    \sum_{k=1}^K 
    \sum_{x \in D^m_{\sigma_k(1):\sigma_k(\lceil\frac{1}{K} |D^m|\rceil)}}
    \log \int_{\Omega} p(x | \theta) q\left(\theta | D^m_{\sigma_k(\lceil\frac{1}{K}|D^m|\rceil+1):\sigma_k(|D^m|)}; \phi \right) \mathrm{d}\theta,
\end{align}
where $\sigma_k$ is the $k$-th permutation of the indices from $1$ to $|D^m|$. To compute this loss, we often partition the data $D^m$ into $K$ partitions. For each partition, we use the rest of the partitions to train a predictor (or a set of predictors) and use this predictor to compute the loss. We average these $K$ loss values and use it as a proxy to the generalization loss~\citep{kohavi1995study}. 

In this particular case above, this cross-validation loss is a function $\phi$ which parameterizes the posterior distribution $q$ over the parameters $\theta$. This parametrization effectively turns the posterior inference problem in Bayesian machine learning into building a predictor that maps a set of training data points into a distribution over the parameters, where this predictor is parametrized using $\theta$. In other words, we train a predictor that solves the posterior inference problem by solving
\begin{align}
    \label{eq:meta-objective}
    \min_{\phi} \frac{1}{M} \sum_{m=1}^M L_{K\mathrm{CV}}(\phi; D^m).
\end{align}
In this case, we would call $\left\{ D^1, \ldots, D^M\right\}$ a \textit{meta-training set}. 

Just like what we have seen earlier in \S\ref{sec:probabilistic-ml}, this $K$-fold cross-validation loss is not easy to compute nor to minimize. Instead, we can use the same technique from variational inference from earlier to minimize the upperbound to $L_{K\mathrm{CV}}$:
\begin{align}
    L_{K\mathrm{CV}}(\phi; D^m) \leq 
    -\frac{1}{K} \sum_{k=1}^K
    \sum_{x \in D^m_{\sigma_k(1):\sigma_k(\lceil\frac{1}{K} |D^m|\rceil)}}
    \frac{1}{B}
    \sum_{b=1}^B 
    \log p(x | \theta^b),
\end{align}
where $\theta^b \sim q\left(\theta | D^m_{\sigma_k(\lceil\frac{1}{K}|D^m|\rceil+1):\sigma_k(|D^m|)}; \phi \right)$. Since $\theta$ is often continuous, we can for instance compute the gradient of $L_{K\mathrm{CV}}$ w.r.t. $\phi$ with the reparametrization trick as long as $q$ is differentiable w.r.t. $\phi$. 

This is interesting, since we can be flexible about how we parametrize $q$, and this $q$ is directly optimized to result in a distribution over $\theta$ or a set of $\theta$'s under which the predictive loss is minimal. In other words, $q$ is a learning algorithm, and we are training a learning algorithm by minimizing the \textit{meta-objective function} in Eq.~\eqref{eq:meta-objective} w.r.t. $q$. 

For instance, we can define $q$ implicitly by drawing a sample of the parameters $\theta$ from $q$ using just a few steps $N$ of stochastic gradient descent, as opposed to running it until convergence as from \S\ref{sec:sgd}. In doing so, we can consider the initialization $\theta_0$ of the parameters as $\phi$. By minimizing the meta-objective function w.r.t. $\theta_0$, we are looking for the initialization of the parameters that are optimal with $N$ SGD steps. If the new training set after such \textit{meta-learning} is similar to the meta-training sets, we would expect that $N$ SGD steps would be enough if not optimal to obtain the best predictor. This approach was originally proposed by \citep{finn2017model} and called model-agnostic meta-learning.

Of course, we can completely forego of any iterative optimization when designing $q$ and build a predictor that directly maps a set of training data points $D$ to the prediction on a new observation $x'$. In doing so, it is important to realize that this predictor cannot simply take as input $D$ but needs to model noise in learning itself. This naturally calls for including latent variables $z$ into this predictor, just like how we did earlier with generative models in \S\ref{sec:probabilistic-ml}. In this case, the posterior distribution $q(\theta)$ is implicit, and we directly predict the predictive probability by
\begin{align}
    p(x | D; \phi) = \int_{\mathcal{Z}} p(x | z; \phi_x) p_z(z | D; \phi_z)\mathrm{d}z,
\end{align}
where $p_z$ is the prior over $z$ and we marginalize out $z$. This approach is often referred as \textit{neural processes}~\citep{garnelo2018neural}. Because this marginalization is often intractable, it is a common practice to approach it from variational inference and learning which we learned already in \S\ref{sec:vae}.

Overall, these approaches are referred as \textit{meta-learning}, since such a procedure results in a predictor that knows how to learn to solve a problem given a set of new examples. Meta-learning can then be used to solve not only learning problems but also any kind of set-to-set problems, such as causal discovery and statistical inference problems. This is an exciting and active area of research.

\section{Regression: Mixture Density Networks}
\label{sec:regression}

Let $e([x,y], \theta)$ be the energy function where $y$ is not categorical with a small number of categories.\footnote{
    A categorical variable takes a value out of a small number of predefined possible values, just like classification. 
} 
Without loss of generality, let $y\in\mathbb{R}^d$. We can turn this into a probability density function by
\begin{align}
    p(y | x; \theta) = 
    \frac{\exp\left(-e([x,y]; \theta)\right)}
    {\int_{\mathbb{R}^d} \exp\left(-e([x, y']; \theta)\right)\mathrm{d}y'}.
\end{align}
Unlike the classification problem we saw in Eq.~\eqref{eq:cross-entropy}, it is extremely challenging to compute the normalization constant in this case with a non-categorical $y$, in general. In fact, this problem is identical to undirected graphical models, such as restricted Boltzmann machines from \S\ref{sec:rbm}, which requires costly MCMC sampling~\citep{boulanger2012modeling}.   

It is thus natural to consider the parametrization of the energy function so that the normalization constant is automatically $1$. We have already considered one particular approach under this paradigm earlier in \S\ref{sec:vae}. With a latent variable $z$ (an unobserved variable), we can make it readily normalized:
\begin{align}
    p(y|x; \theta) = \int_{\mathcal{Z}} p(z) 
    \frac{\exp(-e([x,y], z, \theta))}
    {\int_{\mathbb{R}^d} \exp(-e([x,y'], z, \theta) \mathrm{d}y'}
    \mathrm{d}z.
\end{align}
If we choose the following parametrization of the energy function, we know how to compute the normalization constant exactly, because we end up with the Gaussian distribution over $y$ given $x$ and $z$:
\begin{align}
    e([x, y], z, \theta) = \frac{1}{2} \| y - \mu(x, z; \theta) \|^2.
\end{align}
Unfortunately, this approach is not trivial either, as we must marginalize out the latent variable $z$. This marginalization problem is not easily solvable in general, and we often need to resort to an approximate approach, such as variational inference~\citep{chung2015recurrent}.

Because $y$ is often lower-dimensional than $x$ , there is a tractable alternative to these two intractable approaches. This approach constrains the latent variable approach above so that $|\mathcal{Z}| \ll \infty$ , that is, $z$ can take one of only a few possible values, i.e., $\mathcal{Z} = \left\{1, 2, \ldots, K \right\}$. In that case, we can solve the marginalization problem exactly and arrive at 
\begin{align}
    p(y|x;\theta) = \sum_{z=1}^K \frac{1}{K} \mathcal{N}\left(y; \mu_z(x), \sigma_z^2(x)I\right),
\end{align}
assuming 
\begin{align}
    e([x,y], z, \theta) =
    \frac{1}{2 \sigma_z^2(F(x;\theta_F); \theta_\sigma)} \left\|y - \mu_z(F(x;\theta_F); \theta_\mu)\right\|^2.
\end{align}
This is exactly a mixture of Gaussians however conditioned on $x$. $F(x; \theta_F)$ is a feature extractor of the input $x$, and this extractor is shared between $\mu$ and $\sigma^2$. We further assumed that the prior over the mixture components was uniform, i.e. $p(z) = \frac{1}{K}$. This is not strictly necessary, but simply makes learning easier, as we remove any extra parameters for computing the prior from the input $x$.

As long as $\mu_z$ and $\sigma_z^2$ are differentiable w.r.t. $\theta_\mu$ , $\theta_\sigma$ and $\theta_F$ (collectively, comprising $\theta$,) we can train this predictor all together without having to rely on some approximate marginalization by
\begin{align}
    \min_{\theta_F, \theta_\mu, \theta_\sigma}
    -\frac{1}{N}
    \sum_{n=1}^N
    \log \sum_{z=1}^K \frac{1}{K} \mathcal{N}(y^n; \mu_z(x^n), \sigma_z^2(x^n) I).
\end{align}
Such a predictor is called a mixture density network and outdates all the other approaches above~\citep{bishop1994mixture}. 

A main special case of this mixture density network is when there is only one mixture component, i.e. $K=1$. In that case, this reduces to a more familiar linear regression with the mean squared error loss function, assuming the constant variance, i.e., $\sigma_z^2(x) = c$. Although this is a usual approach and also what we did earlier when we derived backpropagation in \S\ref{sec:backprop}, this approach of a single mixture component has a major disadvantage is that there can only be a single mode in the predictive distribution. This is particularly problematic when the underlying true distribution has multiple local modes, as learning with the criterion above would make this predicted distribution to be dispersed in order to cover all those multiple modes of the true distribution, resulting in an unnecessarily uncertain prediction with the probability mass concentrated on a region of the output space that is relevant to where true modes are. By increasing $K$ beyond 1, we increase the chance of capturing the inherent uncertainty in regression.

Although training can be done exactly, this does not imply that we can make prediction readily with the mixture density network. Unless $K=1$, there is no analytical solution to
\begin{align}
    \hat{y}(x) = \arg\max_{y \in \mathbb{R}^d} 
    \log \sum_{z=1}^K \mathcal{N}(y; \mu_z(x), \sigma^2_z(x) I) - \log K.
\end{align}
We can solve this problem by gradient descent which will find one of at most $K$ modes of this complex distribution or find a saddle point. 

It however is unsatisfactory to return a single point estimate of the solution, when we trained our predictor to capture the full distribution over the output space. Rather, it may be desirable to return a set of possible values of the outcome $y$ that are within a credible region, following the procedure from \S\ref{sec:error-rate}. This is particularly desirable, as we can readily draw as many independent samples from the mixture of Gaussians. Once the samples $\left\{y_1, \ldots, y_M\right\} $ are drawn, we score each sample with the mixture density network, which is again trivial, resulting in $\left\{ p_1, \ldots, p_M \right\}$. We can then fit a cumulative density function on these scores and pick only those that are above a predefined threshold. These selected outputs can be considered a credible set of outputs for $x$. 



\section{Causality}

A major limitation of all methods in this lecture note, perhaps except for reinforcement learning in \S\ref{sec:reinforcement-learning}, is that they all rely almost entirely on association, or correlation. These algorithms all look for which patterns appear together with which other patterns frequently within a given dataset. 

Already in \S\ref{sec:backprop}, this was apparent. For instance, recall the following update rule for a linear block in Eq.~\eqref{eq:hebbian-rule}:
\begin{align}
    \frac{\partial}{\partial u_{ij}} = x_i h_j - x_i \hat{h}_j,
\end{align}
where we assume there was no nonlinearity, i.e. $h'_j=1$. The first term decreases the value of $u_{ij}$ toward the origin $0$ if $x_i$ and the old, undesired value of the $j$-th hidden neuron had the same sign.\footnote{
    We are following the opposite of the gradient direction.
} The second term on the other hand increases the value of $u_{ij}$ away from the origin if $x_i$ and the new, desirable $\hat{h}_j$ have the same sign. In other words, $u_{ij}$, one of the many parameters of this predictor, encodes how correlated the $i$-th dimension of the observation and the $j$-th dimension of the hidden variable are with each other. 

This is perfectly fine, if the goal is to capture such correlations and use them to impute missing values, such as outputs associated with test-time observations. This is not enough however if we want to infer the causal relationship among variables, because as we often say casually, ``correlation does not imply causation.''\footnote{
    When we say this, we are referring to \textit{dependence} by \textit{correlation}, but unless it is technically confusing, I will interchangeably use correlation and dependence in this section. 
}

Let us dig slightly deeper into this statement and consider a few cases where correlation exists but causation does not. The first case is when there exists an unobserved confounder, where the confounder $z$ is defined to affect both the input $x$ and the outcome $y$, such that

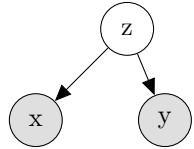
\begin{center}
\begin{tikzpicture}
  \node[obs] (x) {x};
  \node[obs, right=of x] (y) {y};
  \node[latent, above right=of x] (z) {z};

  \path (x) -- (y) -- (z);

  \draw [->] (z) -- (x);
  \draw [->] (z) -- (y);
\end{tikzpicture}
\end{center}

Both $x$ and $y$ are caused by this unobserved confounder $z$ in this diagram, and we can write down the marginal distribution over $(x,y)$ as 
\begin{align}
    p(x,y) = \int p(x|z)p(y|z)p(z) \mathrm{d}z.
\end{align}
It is relatively straightforward to see that this would not be factorized into the product of $p(x)$ and $p(z)$, i.e.
\begin{align}
    \int p(x|z)p(y|z)p(z)\mathrm{d}z \neq p(x) p(z),
\end{align}
unless 
\begin{align}
    \int p(x|z)p(y|z)p(z)\mathrm{d}z = p(x) \int p(y|z)p(z) \mathrm{d}z,
\end{align}
which would imply that there is no edge going from $z$ to $x$ in the first place. 

That we cannot factor $p(x,y)$ into the product of the marginals of $x$ and $y$ implies that $x$ and $y$ are dependent on each other. Equivalently, we can say that $x$ and $y$ are correlated with each other (potentially nonlinearly.) They are however unrelated to each other causally, since intervening on $x$ would not cause any change in $y$ and vice versa. 

An example of this case of an unobserved confounder can be found in driving. If one is not aware of how driving works and only looks at the dashboard of a car,\footnote{
    Imagine you are collecting data from the car to build a self-driving model.
} it is easy to see that the turn indicator and the steering wheel angle are highly correlated with each other, which may result in an incorrect causal conclusion that the turn indicator causes the steering wheel to turn, or vice versa. This is missing a big confounder that is a driver and their intention to turn the car.

The second case is what we often referred to as \textit{confirmation bias}. Consider the following causal model:

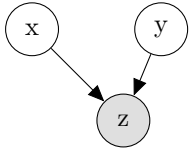
\begin{center}
\begin{tikzpicture}
  \node[latent] (x) {x};
  \node[latent, right=of x] (y) {y};
  \node[obs, below right=of x] (z) {z};

  \path (x) -- (y) -- (z);

  \draw [<-] (z) -- (x);
  \draw [<-] (z) -- (y);
\end{tikzpicture}
\end{center}

In this case, $x$ and $y$ are independent of each other \textit{a priori}. It is clear that they are not causally related to each other, since manually setting one of these to a particular value should not change the value taken by the other variable. It is however interesting to observe that these two variables, $x$ and $y$, are suddenly dependent on each other, once we observe $z$. That is, under the posterior distribution, $x$ and $y$ are not independent:
\begin{align}
    p(x, y | z) = 
    \frac{p(x)p(y)p(z|x,y)}{\int p(x')p(y')p(z|x',y')\mathrm{d}x'\mathrm{d}y'}.
\end{align}
Because of $p(z|x,y)$, we cannot factor $p(x,y|z)$ into the product of two terms, each of which depends only on either $x$ or $y$. If we could, that would imply that $z$ is caused by either one of $x$ or $y$ (or neither.) The input and outcome are correlated in this case, because we only selectively consider a subset of $(x,y)$ pairs that are associated with a particular value of $z$. This is thus also called a \textit{selection bias}. 

Let us consider an example, where $x$ corresponds to a burglary and $y$ to an earthquake. $z$ is a house alarm. The house alarm goes off ($z=1$) when either there is burglary ($x=1$) or there is an earthquake ($y=1$). It is pretty safe for now to assume that the chances of burglary and earthquake are pretty much independent of each other. If you however hear that your alarm went off, that is, if you condition on $z=1$, burglary and earthquake are not independent anymore, since I would be able to \textit{explain away} the chance of burglary if I felt earthquake myself. That is, what's the chance that earthquake and burglary happened together and triggered the alarm. Although there is no causal relationship between the earthquake and burglary, they are now correlated with each other negatively because we are conditioned on alarm going off.

These cases emphasize the difference between association (correlation) and causality. In order to capture causal relationships among variables and use them to \textit{control} the underlying system, we must use an extra set of assumptions and tools to rule out non-causal associations, or so-called spurious correlations. Once we are equipped with such tools, we can make machine learning more robust in more realistic scenarios, for instance where the distribution from which observations are drawn shifts between training and test times. This is a fascinating topic in machine learning and more broadly artificial intelligence, but is out of the scope of this course. I suggest you check out my lecture note ``A Brief Introduction to Causal Inference in Machine Learning''~\citep{cho2024brief} and then move on to more in-depth materials on causal inference, causal discovery and causal representation learning.

\bibliography{main}
\bibliographystyle{abbrvnat}

\end{document}